\pdfoutput=1
\documentclass[a4paper,fleqn]{cas-dc}

\usepackage[numbers]{natbib}
\usepackage{amsmath,amsfonts}
\usepackage{algorithm}
\usepackage{algorithmic}
\usepackage{array}
\usepackage{booktabs}
\usepackage{multirow}
\usepackage{subfig}
\usepackage{textcomp}
\usepackage{url}
\usepackage{verbatim}

\graphicspath{{figs/}}

\begin{document}
\let\WriteBookmarks\relax
\def\floatpagepagefraction{1}
\def\textpagefraction{.001}

\shorttitle{ARB4WM}
\shortauthors{Zhang et al.}

\title [mode = title]{ARB4WM: An Adversarial Robustness Benchmark for World Models in Continuous Control}

\author[1,4]{Junjian Zhang}
\ead{zhangjunjian24@nudt.edu.cn}

\author[2]{Hao Tan}
\ead{23b951016@stu.hit.edu.cn}

\author[3]{Ruonan Li}
\ead{lirn@pcl.ac.cn}

\author[1,4]{Dong Zhu}
\ead{zhud@nudt.edu.cn}

\author[1,4]{Aiping Li}
\cormark[1]
\ead{liaiping@nudt.edu.cn}

\author[2]{Zhaoquan Gu}
\cormark[1]
\ead{guzhaoquan@hit.edu.cn}

\affiliation[1]{organization={College of Computer Science, National University of Defense Technology},
                city={Changsha},
                postcode={410073},
                country={China}}

\affiliation[2]{organization={College of Computer Science and Technology, Harbin Institute of Technology},
                city={Shenzhen},
                postcode={518055},
                country={China}}

\affiliation[3]{organization={Department of New Networks, Peng Cheng Laboratory},
                city={Shenzhen},
                postcode={518055},
                country={China}}

\affiliation[4]{organization={National Key Laboratory of Advanced Communication Networks},
                city={Shijiazhuang},
                postcode={050081},
                state={Hebei},
                country={China}}

\cortext[1]{Correspondence: liaiping@nudt.edu.cn (A.L.); guzhaoquan@hit.edu.cn (Z.G.)}

\begin{abstract}
World models are widely used in robotic and continuous-control systems due to their ability to learn latent dynamics for planning and decision-making. As these systems are increasingly deployed in safety-critical settings, understanding their robustness under adversarial conditions has become essential. However, existing evaluations lack a unified benchmark for testing adversarial threats across the policy, value, and latent-dynamics levels of world-model agents.
To fill this gap, we present ARB4WM, a unified evaluation framework for adversarial robustness of world-model agents under visual perturbations. ARB4WM defines five white-box loss objectives across these three levels and studies their effects when combined with single-step or multi-step perturbation strategies and temporal attack modes, including full-frame, half-sequence, and sparse-frame exposure.
Specifically, we evaluate four Dreamer-style agents across 20 tasks from MetaWorld and the DeepMind Control Suite under different loss objectives, perturbation strategies, and temporal attack modes.
Results show that attacks targeting value estimation, latent representations, and RSSM dynamics can be as damaging as direct policy disruption, and that early or frequent perturbations are especially harmful, while input-level defenses provide limited recovery under adaptive attacks.
These findings suggest that robustness evaluation for world models should cover multiple component-oriented attack objectives and temporal exposure protocols rather than relying solely on action-space robustness.
The source code is available at \url{https://github.com/zaoanguai/ARB4WM}.
\end{abstract}

% \sep robotic manipulation
\maketitle

\section{Introduction}
World models are becoming an important component of intelligent continuous-control systems.
Instead of mapping raw observations directly to actions, these agents encode visual inputs into compact latent states, learn recurrent latent dynamics, and optimize behavior through imagined trajectories.
This structure is attractive for industrial robotics, autonomous inspection, intelligent manufacturing, and other engineering systems in which controllers must reason from high-dimensional perception while interacting with physical processes.
Early recurrent world models~\cite{ha2018world}, PlaNet~\cite{hafner2019planet}, and Dreamer-style latent imagination systems \cite{hafner2020dreamer,hafner2020mastering,hafner2025mastering} have enabled strong performance on pixel-based control and robotic manipulation tasks.
Recent variants \cite{oord2018representation,deng2022dreamerpro,morihira2026r2,micheli2023iris,hansen2024tdmpc2} further improve representation learning, planning, or model scaling through contrastive objectives, prototypical representations, redundancy reduction, transformer dynamics, and model-predictive control.
As these models move from benchmark control toward safety-sensitive engineering applications, their reliability under corrupted or adversarial visual observations becomes a practical safety-assessment problem.
This concern is amplified by world-model-like systems \cite{lecun2022path,gu2024advancing,wang2024drivedreamer,zheng2024occworld} being increasingly used in robotics, autonomous driving, and general embodied intelligence.
In a closed-loop industrial controller, a small visual perturbation is not merely an image-level error.
It may shift the estimated pose of a manipulator, destabilize a balancing controller, corrupt a latent belief state, or cause the agent to evaluate an unsafe trajectory as acceptable.
Therefore, adversarial threats against world-model agents are not merely perception-level errors. They can propagate through latent dynamics, imagined rollouts, and decision-making, ultimately causing control degradation or internal decision failure.

Adversarial examples \cite{szegedy2013intriguing,yuan2019adversarial} provide a concrete and widely studied manifestation of such adversarial threats.
In reinforcement learning, the problem is more difficult because a perturbed observation can change the current action, future state distribution, and accumulated return.
Prior work \cite{behzadan2017robustness,huang2017adversarial,kos2017delving,lin2017tactics,gleave2020adversarial,pinto2017robust,zhang2020robust} has shown that deep policies can be vulnerable to adversarial observations, strategically timed attacks, adversarial policies, and perturbed state observations.
However, most existing attacks and evaluations are designed for model-free policies or fixed observation-space attacks.
They usually treat the victim agent as a policy network and evaluate a limited set of attack objectives.
This is insufficient for pixel-based world models.
In Dreamer-style agents, an adversarial observation can affect not only the immediate action, but also the recurrent belief state, RSSM posterior and prior, latent representation, imagined value prediction, and future rollout dynamics.
Therefore, a policy-level evaluation may miss important failure modes that occur inside the latent world-model pipeline.

Recent work on world-model safety and adversarial evaluation \cite{zeng2024world,huang2024safedreamer,ye2024robustmbrl,sun2024latentdynamic,zhang2026hallucination,guo2026wmattack} further highlights this concern.
These studies show that world-model agents can fail under adversarial observations, latent-space perturbations, and attacks on learned dynamics, and that robustness estimates depend strongly on the attack protocol.
However, we still lack a testing framework that compares multiple Dreamer-family agents across a broad task set while separating failures in policy distributions, value prediction, latent representation stability, and RSSM dynamics consistency.

The central problem is therefore to diagnose where adversarial visual perturbations disrupt the Dreamer-style latent decision-making pipeline, how these vulnerabilities differ across Dreamer-family architectures, and how recurrent latent states respond to continuous or temporally sparse attacks.
To address this problem, we present ARB4WM, a repeatable robustness testing framework for world-model agents.
ARB4WM evaluates Dreamer, R2-Dreamer, Dreamer-InfoNCE, and Dreamer-Pro on MetaWorld and the DeepMind Control Suite (DMC) \cite{yu2020metaworld,tassa2018dmcontrol} under a unified attack interface.
Given a victim world model, task suite, perturbation objective, temporal exposure mode, and defense setting, the framework reports clean-normalized area under curve (nAUC), defense recovery, adaptive-defense gap, and policy saliency diagnostics to connect closed-loop return degradation with internal failure modes.

In summary, we make the following contributions.
\begin{itemize}
    \item We build a safety-oriented robustness testing framework for world-model agents in industrial continuous control, covering four Dreamer-family agents and 20 simulated robotic manipulation and continuous-control tasks from MetaWorld and DMC.
    \item We design five white-box test objectives that target different components of the world-model pipeline, including policy distributions, value estimation, latent representations, and RSSM dynamics consistency.
    \item We introduce clean-normalized robustness metrics, adaptive-defense recovery analysis, and saliency diagnostics that support safety assessment by linking closed-loop performance loss to latent dynamics, value prediction, and policy evidence.
    \item We evaluate both single-step and multi-step iterative perturbation optimization, together with full-frame, half-sequence, and sparse temporal exposure protocols, to study how adversarial effects propagate through recurrent latent states.
\end{itemize}

\section{Related Work}

In this section, we review the literature most relevant to ARB4WM, including world-model control, adversarial robustness in reinforcement learning, and recent safety evaluations of world-model agents.

% \subsection{Engineering Informatics for Safe Intelligent Control}

% Engineering informatics studies how computational representations, data-driven models, and decision-support systems can improve engineering design, operation, and risk management.
% For intelligent continuous-control systems, this perspective is different from evaluating an agent only by its clean-task return.
% The engineering question is whether the learned perception-control pipeline remains dependable when observations are corrupted, whether failures persist after the disturbance disappears, and whether the diagnostic information is sufficient for model selection or pre-deployment testing.
% World-model controllers introduce a particularly informative case because their decisions are mediated by latent representations, recurrent dynamics, value estimates, and imagined trajectories.
% A robustness test for such systems should therefore expose not only action-level failure, but also the internal computational components that make the controller unsafe under visual perturbations.

\subsection{World Models for Visual Control}

World models learn compact internal dynamics that can be used for planning, imagination, and policy optimization.
Recent neural world-model approaches \cite{ha2018world,hafner2019planet,hafner2020dreamer} showed that latent dynamics can support policy search, planning, and behavior learning from pixels.
DreamerV2 and DreamerV3 \cite{hafner2020mastering,hafner2025mastering} scaled recurrent state-space models to Atari, continuous-control, and diverse control domains.
Other model-based agents \cite{micheli2023iris,hansen2024tdmpc2,deng2022dreamerpro,oord2018representation,morihira2026r2} improve sequence modeling, model-predictive control, or representation learning through transformer world models, TD-MPC-style objectives, reconstruction-free prototypical learning, contrastive predictive coding, and redundancy-reduced latent states.
World models for autonomous driving, humanoid locomotion, embodied interaction, and general simulation \cite{gu2024advancing,wang2024drivedreamer,zheng2024occworld,yang2024simulators} are also increasingly studied beyond standard control benchmarks.
These studies mainly emphasize return, sample efficiency, or representation quality.
They do not evaluate whether the learned latent dynamics remain reliable when the visual input is adversarially perturbed across a controlled set of internal attack targets.

\subsection{Adversarial Robustness in Reinforcement Learning}

Adversarial-example studies \cite{szegedy2013intriguing,FGSM,PGD} first showed in supervised learning that small bounded perturbations can cause incorrect predictions.
Later attack and benchmark work \cite{papernot2017practical,chen2020hopskipjumpattack,brown2017adversarial,eykholt2018robust,croce2020autoattack,andriushchenko2020square,croce2021robustbench} broadened the evaluation toolkit with black-box, decision-based, patch, physical-world, and standardized benchmark attacks.
Defense studies \cite{stutz2019disentangling,cohen2019certified} also showed that apparent robustness can depend strongly on the evaluation protocol, motivating careful use of adaptive attacks and standardized metrics.
In reinforcement learning, the problem is more complex because perturbations can change actions, future states, and accumulated returns.
Early deep-RL attack work \cite{behzadan2017robustness,huang2017adversarial,kos2017delving} showed that neural network policies are vulnerable to adversarial observation perturbations.
Strategically timed attacks \cite{lin2017tactics} perturb only selected frames while still degrading long-horizon behavior, and adversarial policies \cite{gleave2020adversarial} show that the opponent or environment interaction can be used as an attack surface.
Robustness-oriented methods \cite{pinto2017robust,zhang2020robust} have also been proposed, including robust adversarial reinforcement learning with adversarial dynamics disturbances and state-adversarial MDP formulations for observation perturbations.
Safety-critical domains such as autonomous driving and robotic vision-language-action control \cite{cao2019adversarial,jones2025adversarial} further show that perception attacks can create downstream control risk.
Most of these works focus on model-free policies or general observation attacks.
They reveal that reinforcement learning agents are vulnerable, but they do not explain which internal components of a world model are responsible for the failure.

\subsection{Adversarial Evaluation of World Models}

World-model agents introduce new robustness challenges because observations are first mapped into latent states and then propagated through recurrent dynamics.
Safety and robustness studies for world models \cite{as2024safe,huang2024safedreamer,ye2024robustmbrl,zollicoffer2025surprise} have begun to examine safe exploration, robust model-based RL, adversarial corruption, and surprise-based detection.
Diagnostic world-model evaluations \cite{upadhyay2026worldbench} analyze the reliability of learned world models through physics disambiguation and related diagnostic tasks.
Hallucination-driven policy failure \cite{zhang2026hallucination} is closely related to our work because it directly studies adversarial attacks against world-model agents.
It verifies the adversarial risk of early world-model controllers through white-box attacks and explores spatial and temporal vulnerability through latent-space and temporally correlated perturbations.
However, its empirical scope is limited to the original World Models architecture and Dreamer, uses a smaller set of control tasks, and concentrates on policy-level and RSSM-level attack targets.
WMAttack \cite{guo2026wmattack} is also related because it targets adversarial evaluation of world-model agents and shows that automated search can find stronger attack configurations than manually selected baselines.
It formulates evaluation as a finite-budget search over attack families, perturbation budgets, optimization steps, restarts, schedules, and allocation rules, then uses retrieval and feedback-conditioned refinement to improve search efficiency.

% \subsection{Summary and Positioning}

The existing literature establishes three relevant foundations: world models provide effective latent dynamics for visual control, adversarial reinforcement-learning studies show that closed-loop policies can be vulnerable to observation perturbations, and recent world-model attack studies demonstrate that world-model agents can fail under adversarial visual inputs.
However, prior work either focuses on nominal control performance, attacks model-free policies, studies a limited set of world-model agents and attack targets, or searches for strong attack configurations without isolating which internal component fails.
ARB4WM is positioned as a complementary robustness benchmark.
It compares four Dreamer-family agents on 20 MetaWorld and DMC tasks, separates five component-level objectives covering policy distributions, policy entropy, value prediction, latent-state drift, and RSSM dynamics consistency, and evaluates temporal exposure, input defense, adaptive-defense gaps, and saliency diagnostics within one repeatable testing workflow.

\section{Preliminaries}

In this section, we define the Dreamer-style world-model pipeline and the adversarial threat model used throughout the benchmark.

\subsection{World-Model Reinforcement Learning}
\label{sec:prelim_world_model_rl}

World-model reinforcement learning first maps high-dimensional observations into compact latent states and then uses learned latent dynamics for prediction, imagination, and control.
At timestep $t$, the agent receives an observation $o_t$, executes an action $a_t$, and receives a reward $r_t$.
The observation is encoded as
\begin{equation}
\label{eq_pre_encoder}
e_t = \mathrm{Enc}(o_t).
\end{equation}
The world model maintains a latent state $s_t$ and learns a transition model $p(s_t \mid s_{t-1}, a_{t-1})$, as well as an observation-conditioned posterior $q(s_t \mid s_{t-1}, a_{t-1}, o_t)$.
The latent state is then used for policy prediction, value estimation, and imagined rollouts.

Among world-model agents, the Dreamer family \cite{hafner2020dreamer,hafner2020mastering,hafner2025mastering} is a representative RSSM-based framework.
Dreamer-style agents decompose the latent state into a stochastic state $z_t$ and a deterministic recurrent state $h_t$, and use their concatenation as the feature for downstream prediction:
\begin{equation}
\label{eq_feature}
f_t = [z_t, h_t],
\end{equation}
where $[\cdot,\cdot]$ denotes concatenation.
The actor predicts the action distribution $\pi(a_t\mid f_t)$, the critic estimates $V(f_t)$, and the dynamics model supports latent imagination $(z_{t+1}, h_{t+1}) \sim p(\cdot \mid z_t, h_t, a_t)$.
This shared encoder--RSSM--actor-critic pipeline is the main reason why adversarial visual perturbations can affect not only the current action, but also recurrent belief states, value estimates, and imagined future trajectories.

This work evaluates four agents that share this general structure but differ in representation learning or optimization objectives: Dreamer \cite{hafner2020dreamer}, R2-Dreamer \cite{morihira2026r2}, Dreamer-Pro \cite{deng2022dreamerpro}, and Dreamer-InfoNCE \cite{oord2018representation}.
We do not rely on these variants having identical training losses.
Instead, ARB4WM uses their shared inference components to define comparable attack surfaces for policy distributions, value prediction, latent-state drift, and RSSM dynamics consistency.

\subsection{Adversarial Threat Model}
\label{adv}

We consider a white-box adversarial setting in which the attacker has full access to model parameters, gradients, and intermediate latent representations of the target agent.
Given an input observation $o_t$, the attacker aims to construct an adversarial perturbation $\delta_t$ that maximally degrades the agent's decision-making performance while remaining visually imperceptible.
The perturbation is constrained by $\|\delta_t\|_\infty \le \epsilon$, where $\epsilon$ controls the maximum perturbation magnitude, and the perturbed adversarial observation is defined as $\tilde o_t = o_t + \delta_t$.

Under the white-box setting, the attacker first defines an attack objective function $\mathcal L$, which can target policy distributions, value estimation, latent dynamics, or other internal components of the Dreamer-family agent.
The adversarial perturbation is then generated by maximizing the attack objective with respect to the input observation:
\begin{equation}
\label{eq_attack_obj}
\delta_t^*
=
\arg\max_{\|\delta_t\|_\infty \le \epsilon}
\mathcal L(o_t + \delta_t).
\end{equation}

To solve the above optimization problem, we consider single-step and multi-step iterative gradient-based attack methods.
The single-step variant follows the Fast Gradient Sign Method (FGSM) \cite{FGSM}, while the multi-step variant follows Projected Gradient Descent (PGD) \cite{PGD}.

\section{Methodology}

In this section, we present the ARB4WM testing framework, including its attack objectives, optimization procedures, temporal exposure protocols, data-level defenses, and adaptive attack setting.

\begin{figure*}[pos=tbp]
\centering
\includegraphics[width=\linewidth]{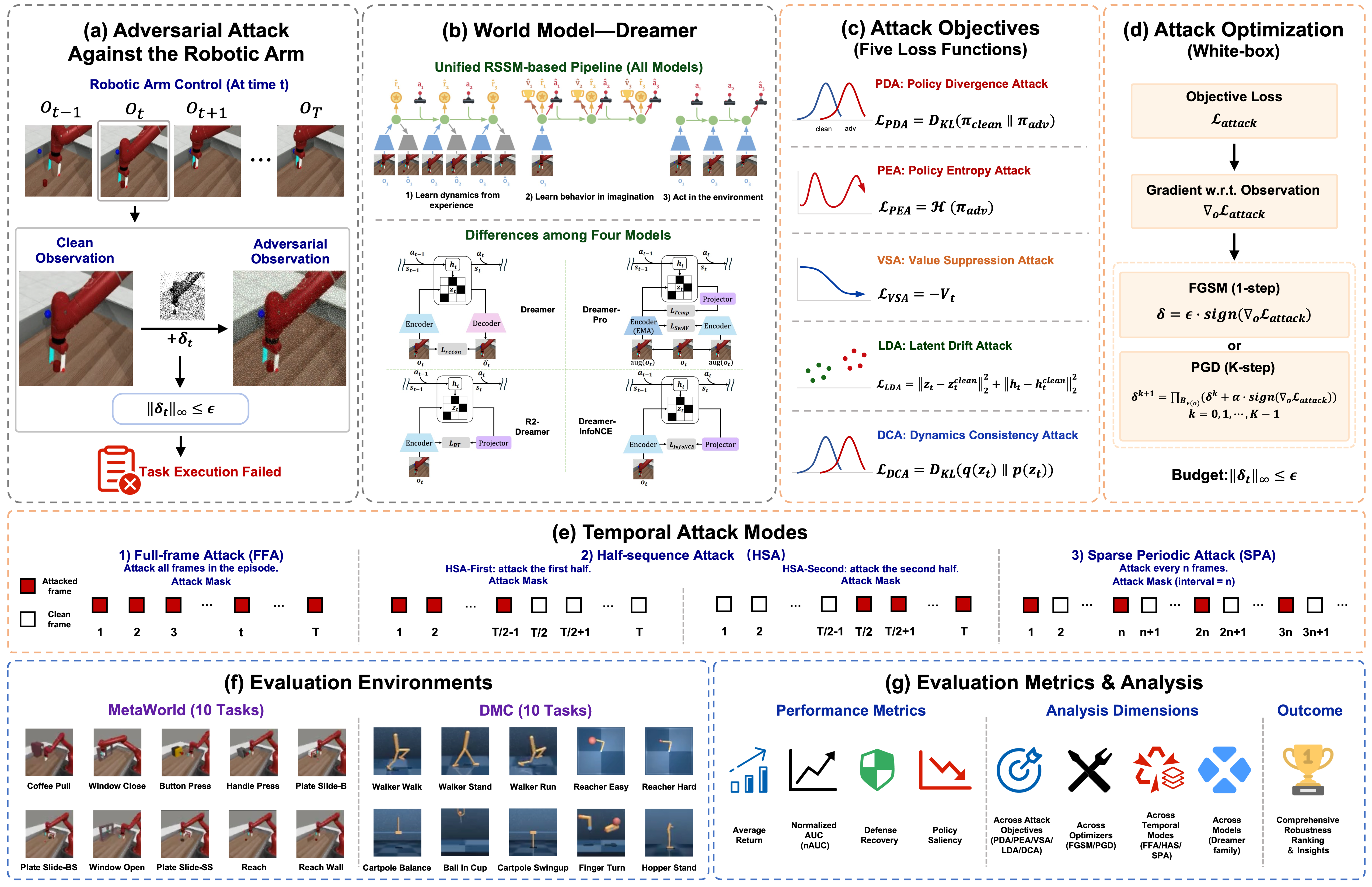}
\caption{
Overview of ARB4WM.
Panel (a) illustrates a motivating adversarial-attack example for robotic-arm visual control, where a bounded visual perturbation can change the controller's action and trajectory.
Panel (b) summarizes the common Dreamer-family world-model pipeline and the main differences among Dreamer, R2-Dreamer, Dreamer-InfoNCE, and Dreamer-Pro.
Panel (c) defines the five component-level attack objectives used in ARB4WM: policy divergence, policy entropy, value suppression, latent drift, and dynamics consistency.
Panel (d) shows the two perturbation optimizers, single-step and multi-step iterative attacks.
Panel (e) shows the three temporal exposure modes: full-frame, half-sequence, and sparse-frame attacks.
Panel (f) lists the 10 MetaWorld and 10 DMC tasks used for evaluation.
Panel (g) summarizes the reported evaluation and analysis outputs, including clean performance, nAUC, defense recovery, adaptive-defense gap, robustness ranking, and policy saliency diagnostics.
}
\label{fig:arb4wm_framework}
\end{figure*}

\subsection{ARB4WM Overview}

To evaluate the safety of world-model controllers under adversarial visual observations, we establish a unified robustness testing framework for the Dreamer family, including Dreamer \cite{hafner2020dreamer}, R2-Dreamer \cite{morihira2026r2}, Dreamer-Pro \cite{deng2022dreamerpro}, and Dreamer-InfoNCE \cite{oord2018representation}.
Dreamer and its optimized variants are representative world-model agents because they combine pixel encoders, RSSM latent dynamics, imagined rollouts, and actor-critic decision making within a common latent-control pipeline.
Fig.~\ref{fig:arb4wm_framework}(a) illustrates the motivating visual-control attack scenario, and Fig.~\ref{fig:arb4wm_framework}(b) summarizes the shared Dreamer-family pipeline used to define comparable attack surfaces.

Unlike conventional adversarial reinforcement learning studies that focus on a single attack objective or a single policy architecture, ARB4WM evaluates robustness from multiple perspectives of the latent world-model pipeline.
Specifically, we design five test objectives targeting different components of the Dreamer architecture, including policy distributions, value estimation, latent representations, and RSSM dynamics consistency.
In addition, we evaluate multiple temporal exposure protocols to analyze both continuous and sparse adversarial perturbation scenarios.
From an engineering-testing perspective, each run is defined by a task set, a victim controller, an attack objective, an optimizer, a perturbation budget, a temporal exposure mode, and an optional input defense.
The output is a structured safety report containing clean performance, attacked performance curves, clean-normalized robustness summaries, recovery scores, adaptive-defense gaps, and saliency diagnostics.

The framework consists of five major testing dimensions:

\begin{itemize}
    \item \textbf{Test Objectives:}
    policy, value, latent, and dynamics-consistency attacks;

    \item \textbf{Attack Optimizers:}
    single-step and multi-step iterative adversarial perturbation generation;

    \item \textbf{Temporal Exposure Modes:}
    full-frame exposure, half-sequence exposure, and sparse periodic exposure;

    \item \textbf{Data-Level Defense and Adaptive Testing:}
    input preprocessing defenses such as compression, smoothing, and noise injection, together with defense-aware attacks that optimize perturbations through or around the defended inference pipeline.
\end{itemize}

This framework enables a comprehensive robustness evaluation of world-model reinforcement learning agents under diverse adversarial conditions while preserving a system-level view of closed-loop safety.

\subsection{Safety Testing Workflow}

ARB4WM follows a four-stage workflow.
The first stage defines the operational test bed by selecting continuous-control tasks, visual observation settings, victim world-model agents, and clean reference trajectories.
The second stage configures adversarial exposure by selecting the component-level objective, perturbation optimizer, budget, and attack timing.
The third stage runs closed-loop evaluation and records task return or success rate together with intermediate world-model quantities, including policy outputs, value estimates, stochastic latent states, deterministic recurrent states, and RSSM posterior-prior statistics.
The fourth stage summarizes the results through robustness metrics and diagnostic visualizations.

This workflow is intended to support pre-deployment assessment rather than post-hoc attack demonstration.
For industrial continuous control, a testing framework should answer whether a controller fails under persistent corruption, whether it can recover after early perturbations stop, whether simple input transformations provide genuine protection under adaptive attacks, and whether the observed failure is associated with policy, value, latent-state, or dynamics-consistency disruption.
The following sections describe how ARB4WM exposes these failure modes through a unified attack interface and a fixed taxonomy of test objectives.

\subsection{Unified Attack Interface for Dreamer-family Agents}

Although Dreamer \cite{hafner2020dreamer}, R2-Dreamer \cite{morihira2026r2}, Dreamer-Pro \cite{deng2022dreamerpro}, and Dreamer-InfoNCE \cite{oord2018representation} differ in representation learning objectives and policy optimization strategies, they share a common RSSM-based latent imagination architecture, as summarized in Section~\ref{sec:prelim_world_model_rl}.
This shared structure enables adversarial attacks to access and perturb multiple internal components of the world-model pipeline in a unified manner.

At each timestep, the agent maintains latent representations composed of stochastic latent states, deterministic recurrent hidden states, latent features, policy outputs, value predictions, and imagined future trajectories.
These intermediate representations provide multiple attack surfaces beyond conventional observation-space attacks.

To support unified adversarial evaluation across different Dreamer-family agents, we extract the policy distribution $\pi(a_t|f_t)$, value prediction $V(f_t)$, stochastic latent state $z_t$, deterministic recurrent hidden state $h_t$, RSSM posterior and prior distributions $q(z_t)$ and $p(z_t)$, and imagined latent trajectories with future value estimates $(\hat z_{t+k}, \hat h_{t+k}, \hat V_{t+k})$ during inference and imagination rollout.

Based on these internal representations, adversarial objectives can target different stages of the latent world-model pipeline, including policy-space perturbation, value-space disruption, latent representation manipulation, and transition dynamics inconsistency.

This unified interface enables consistent adversarial evaluation across all considered Dreamer-family agents while preserving architecture-specific learning characteristics.

\subsection{Attack Objective Taxonomy}

ARB4WM defines five white-box attack objectives, each corresponding to a different failure mode in the latent decision-making pipeline.
The first three objectives target policy or value quantities that are common to actor-critic reinforcement learning agents, while the last two target latent-state and dynamics-consistency mechanisms that are specific to Dreamer-style world models.
These component-level objectives correspond to the attack targets summarized in Fig.~\ref{fig:arb4wm_framework}(c).

\subsubsection{Policy Divergence Attack (PDA)}

PDA is motivated by prior adversarial attacks on deep reinforcement-learning policies \cite{huang2017adversarial,kos2017delving,lin2017tactics}, which show that small observation changes can induce harmful action changes.
For continuous-control agents, however, an action sample alone is too narrow a target because the policy represents a full Gaussian decision distribution.
PDA therefore attacks the distributional stability of the policy by forcing the adversarial policy to deviate from the clean decision distribution:
\begin{equation}
\label{eq_pda}
\mathcal L_{\mathrm{PDA}}
=
D_{\mathrm{KL}}
\left(
\pi_{\mathrm{clean}}
\;\|\;
\pi_{\mathrm{adv}}
\right),
\end{equation}
where $\pi_{\mathrm{clean}}$ and $\pi_{\mathrm{adv}}$ denote the policy distributions generated from clean and adversarial observations, respectively, and $D_{\mathrm{KL}}(\cdot\|\cdot)$ denotes the Kullback--Leibler divergence.
For continuous control policies modeled as Gaussian distributions, the KL divergence is computed using the predicted action means and standard deviations.
By maximizing policy divergence, the attack can shift the action mean away from the clean trajectory or alter the predicted standard deviation, leading to unstable action generation and degraded control performance.
Compared with earlier policy-output attacks, PDA is used here as a controlled benchmark objective that can be applied to both model-free actor-critic agents and Dreamer-family world models while keeping the clean policy as the reference distribution.

\subsubsection{Policy Entropy Attack (PEA)}

PEA targets a different policy-level failure mode.
Entropy in actor-critic reinforcement learning \cite{andrychowicz2021matters} is a standard quantity because it controls the stochasticity of action selection and exploration.
During evaluation, excessive entropy can make a continuous-control policy indecisive even if its mean action is not explicitly pushed toward a chosen wrong direction.
PEA therefore increases the uncertainty of the adversarial policy:
\begin{equation}
\label{eq_pea}
\mathcal L_{\mathrm{PEA}}
=
\mathcal H(\pi_{\mathrm{adv}}),
\end{equation}
where $\mathcal H(\cdot)$ denotes the entropy operator and $\pi_{\mathrm{adv}}$ denotes the adversarial policy distribution.
For Gaussian policies, the entropy is computed from the predicted action variances.
In continuous control, small increases in action stochasticity can accumulate into unstable torques, missed contacts, or inaccurate reaching motions, especially when the policy must maintain precise posture or object interaction.
Unlike PDA, PEA does not require the adversarial policy to move away from the clean policy in a particular distributional direction.
It evaluates whether the policy head preserves confident action selection under adversarial perturbations, and it is not tied to Dreamer-specific latent dynamics.

\subsubsection{Value Suppression Attack (VSA)}

VSA is motivated by the role of critic estimates in actor-critic methods \cite{andrychowicz2021matters} and by robust RL formulations in which value estimates are affected by adversarially perturbed observations \cite{zhang2020robust}.
In Dreamer-family agents, this vulnerability is amplified because value prediction is also used inside latent imagination to guide long-horizon behavior learning.
VSA therefore attacks the critic by minimizing the predicted state value:
\begin{equation}
\label{eq_vsa}
\mathcal L_{\mathrm{VSA}}
=
-
V(\tilde f_t),
\end{equation}
where $V(\cdot)$ denotes the critic value function and $\tilde f_t$ denotes the adversarial feature used by the critic.
If adversarial observations drive the critic to underestimate promising states, the agent may avoid useful actions or assign low value to imagined rollouts that would otherwise support successful control.
Compared with policy-only attacks, VSA exposes a value-oriented attack surface: the perturbation does not need to directly maximize action error if it can corrupt the critic signal that shapes action selection and imagined trajectory evaluation.

The second group targets components that are specific to Dreamer-family world models.
Unlike PPO-style agents, Dreamer agents maintain recurrent latent states and learn RSSM transition dynamics.
An adversarial observation can therefore affect future behavior by corrupting the internal belief state, even if the immediate policy output is not directly attacked.
LDA and DCA are designed to evaluate these world-model-specific vulnerabilities.

\subsubsection{Latent Drift Attack (LDA)}

LDA is motivated by the dependence of Dreamer-family agents on stable latent belief states.
The stochastic latent state $z_t$ captures uncertainty-aware environment representations, while the deterministic hidden state $h_t$ carries recurrent temporal information.
If an adversarial observation moves either component away from its clean counterpart, the resulting belief error can propagate into policy prediction, value estimation, and imagination rollout.
LDA therefore perturbs latent representations by maximizing the discrepancy between clean and adversarial latent states:
\begin{equation}
\label{eq_lda}
\mathcal L_{\mathrm{LDA}}
=
\|z_t-z_t^{\mathrm{clean}}\|_2^2
+
\|h_t-h_t^{\mathrm{clean}}\|_2^2,
\end{equation}
where $z_t$ and $h_t$ denote the adversarial stochastic latent state and deterministic recurrent hidden state, respectively, $z_t^{\mathrm{clean}}$ and $h_t^{\mathrm{clean}}$ denote their corresponding clean latent representations, and $\|\cdot\|_2^2$ denotes the squared $\ell_2$ distance.
Because the deterministic state $h_t$ carries recurrent memory, latent drift can persist beyond the attacked frame and contaminate subsequent belief updates even when later observations are less perturbed.
LDA evaluates the robustness of latent representation learning under adversarial perturbations.

\subsubsection{Dynamics Consistency Attack (DCA)}

DCA is motivated by the posterior-prior consistency that underlies RSSM learning in Dreamer-style agents.
The posterior distribution incorporates the current observation, while the prior distribution predicts the latent state from past dynamics and actions.
Stable latent imagination requires these two distributions to remain compatible.
DCA therefore attacks the consistency between RSSM posterior and prior latent distributions:
\begin{equation}
\label{eq_dca}
\mathcal L_{\mathrm{DCA}}
=
D_{\mathrm{KL}}
\left(
q(z_t)
\;\|\;
p(z_t)
\right),
\end{equation}
where $q(z_t)$ denotes the RSSM posterior latent distribution conditioned on current observations and $p(z_t)$ denotes the RSSM prior transition distribution predicted from previous latent states and actions.
By maximizing this discrepancy, the attack degrades the fidelity of latent imagination rollouts and destabilizes future trajectory prediction.
When the posterior inferred from the current observation becomes inconsistent with the prior predicted from past dynamics, the agent receives conflicting latent information, which can weaken transition consistency and impair planning through imagination.
DCA evaluates the robustness of RSSM dynamics modeling against adversarial perturbations.

These five objectives separate two levels of adversarial failure.
PDA, PEA, and VSA evaluate whether adversarial observations can disrupt policy distributions or critic estimates without relying on world-model-specific variables, so they are also meaningful for model-free actor-critic agents.
LDA and DCA evaluate vulnerabilities introduced by Dreamer-style latent modeling: the recurrent belief state can drift away from the clean trajectory, and the posterior inferred from observations can become inconsistent with the prior predicted by learned dynamics.
This taxonomy allows ARB4WM to compare general actor-critic attack surfaces with failure modes that arise from latent world-model learning.

\subsection{Adversarial Optimization}

Following the white-box adversarial threat model defined in Section~\ref{adv}, we optimize adversarial perturbations using gradient-based attack methods under the $\ell_\infty$ perturbation constraint.
ARB4WM uses two optimization regimes because they stress different aspects of robustness.
A single-step update measures whether the local input gradient is already sufficient to disrupt the agent, while a multi-step iterative update tests whether repeated projected ascent can find stronger perturbation directions under the same budget.
This distinction is useful for world models because a perturbation can affect both the immediate policy output and the recurrent belief state that is carried forward.
Fig.~\ref{fig:arb4wm_framework}(d) shows these two perturbation optimizers within the benchmark workflow.

The single-step optimizer follows FGSM \cite{FGSM} and generates the perturbation with one gradient ascent step:
\begin{equation}
\label{eq_fgsm}
\delta
=
\epsilon \cdot
\mathrm{sign}
\left(
\nabla_o \mathcal L_{\mathrm{attack}}
\right).
\end{equation}

This optimizer provides an efficient local-sensitivity baseline.
The multi-step optimizer follows PGD \cite{PGD} and iteratively updates the perturbed observation through projected gradient ascent:
\begin{equation}
\label{eq_pgd}
o^{k+1}
=
\Pi_{\mathcal B_\epsilon(o)}
\left(
o^k
+
\alpha \cdot
\mathrm{sign}
\left(
\nabla_o \mathcal L_{\mathrm{attack}}
\right)
\right),
\end{equation}
where $\Pi_{\mathcal B_\epsilon(o)}$ denotes projection onto the $\ell_\infty$ perturbation region.
Compared with the single-step variant, the multi-step optimizer can refine the perturbation after projection and therefore evaluates robustness under more challenging attack conditions.

\subsection{Temporal Attack Protocols}

In practical deployment scenarios, continuous adversarial attacks on every observation frame may be unrealistic due to computational and resource constraints.
Therefore, in addition to conventional full-frame attacks, we further evaluate robustness under partial and sparse temporal attack settings.
These exposure modes are summarized in Fig.~\ref{fig:arb4wm_framework}(e).

\subsubsection{Full-frame Attacks}

Full-frame attacks perturb every observation frame throughout the entire episode.
This setting evaluates the fundamental robustness limit of world-model agents under continuous adversarial perturbations and represents the strongest adversarial scenario in our benchmark.
Because every recurrent update receives an adversarially modified observation, full-frame attacks can continuously corrupt both immediate action selection and the latent state carried into future timesteps.
This protocol therefore measures the upper-bound attack pressure that a fixed perturbation budget can impose during closed-loop evaluation.

\subsubsection{Half-sequence Attacks}

Half-sequence attacks perturb only part of the episode trajectory.
We consider two variants:
\begin{itemize}
    \item attacking only the first half of frames;
    \item attacking only the second half of frames.
\end{itemize}

The first-half attack evaluates the long-term destructive effect of adversarial perturbations introduced during early-stage trajectory execution.
Since Dreamer-family agents rely on recurrent latent memory and imagination rollout, early perturbations may propagate through future latent dynamics.
If the agent fails to recover after the attack stops, this suggests that early latent corruption has been stored in the recurrent belief state and continues to affect subsequent decisions.

The second-half attack evaluates the vulnerability of the agent during stable policy execution after latent states have already been formed.
This setting is complementary to the first-half attack: it tests whether a policy that has already entered a reasonable trajectory can still be destabilized by late-stage adversarial observations, for example during precise reaching, balancing, or object interaction.

Furthermore, by stopping adversarial perturbations midway through the episode, half-sequence attacks additionally evaluate the recovery capability of world-model agents after adversarial interference disappears.

\subsubsection{Sparse Periodic Attacks}

Sparse periodic attacks perturb observations every $n$ frames:
\begin{equation}
\label{eq_sparse_attack}
\mathcal T
=
\{t \mid t \bmod n = 0\}.
\end{equation}

This setting simulates realistic sparse adversaries with limited attack frequency.
Sparse attacks evaluate include:
\begin{itemize}
    \item the minimum perturbation frequency required to destabilize the agent;
    \item the temporal robustness of latent imagination dynamics;
    \item the resilience of recurrent latent states against intermittent adversarial interference.
\end{itemize}

Compared with continuous attacks, sparse attacks provide additional insights into how adversarial perturbations accumulate and propagate within world-model reinforcement learning systems.
They are also closer to constrained attack scenarios where the adversary cannot perturb every frame due to limited computation, communication bandwidth, or sensing opportunities.
If sparse perturbations still cause large return degradation, the result indicates that world-model agents may amplify low-frequency observation corruption through recurrent state updates.

\subsection{Unified Attack Evaluation Procedure}

Algorithm~\ref{alg:arb4wm_attack} summarizes the unified evaluation procedure implemented in ARB4WM.
The procedure exposes the main experimental choices used in our benchmark: the attack objective, temporal attack mode, perturbation optimizer, optional input defense, and whether the attack is adaptive to the defense.
When defense is enabled, the defended observation is passed to the agent before action selection.
For adaptive attacks, gradients are computed through the defended input using BPDA/EOT-style approximation so that the attacker optimizes against the actual defended inference pipeline.

\begin{algorithm}[t]
\caption{ARB4WM Unified Adversarial Evaluation}
\label{alg:arb4wm_attack}
\begin{algorithmic}[1]
\REQUIRE Agent $M$, environment $E$, objective $a$, temporal mode $q$, optimizer $o$, budget $\epsilon$, steps $K$, defense $d$, adaptive flag $\eta$
\STATE Initialize recurrent state $s_0$ and previous action $u_0$
\FOR{each timestep $t$ until episode termination}
    \STATE Receive clean observation $x_t$ from $E$
    \IF{$q$ selects timestep $t$}
        \STATE Set $x_t^{0}=x_t$
        \FOR{$k=0$ to $K-1$}
            \STATE Choose loss $\mathcal L_a$ from \{PDA, PEA, VSA, LDA, DCA\}
            \IF{$d$ is enabled and $\eta$ is true}
                \STATE Compute $\mathcal L_a$ on defended input $d(x_t^k)$ using adaptive gradients
            \ELSE
                \STATE Compute $\mathcal L_a$ on current adversarial input $x_t^k$
            \ENDIF
            \STATE Update perturbation with optimizer $o$ (single-step if $K=1$, multi-step if $K>1$)
            \STATE Project $x_t^{k+1}$ to $\|x_t^{k+1}-x_t\|_\infty \le \epsilon$
        \ENDFOR
        \STATE Set attacked observation $\tilde{x}_t=x_t^K$
    \ELSE
        \STATE Set attacked observation $\tilde{x}_t=x_t$
    \ENDIF
    \IF{$d$ is enabled}
        \STATE Set inference observation $\hat{x}_t=d(\tilde{x}_t)$
    \ELSE
        \STATE Set inference observation $\hat{x}_t=\tilde{x}_t$
    \ENDIF
    \STATE Agent selects action $u_t=M(\hat{x}_t,s_t)$ and updates recurrent state
    \STATE Step environment with $u_t$ and accumulate return
\ENDFOR
\RETURN episode return and robustness metrics
\end{algorithmic}
\end{algorithm}

\section{Experimental Setup}

In this section, we describe the benchmark tasks, victim agents, attack protocol, and evaluation metrics used to quantify robustness and recovery.

\subsection{Evaluation Environments}

We evaluate adversarial robustness on two pixel-based benchmarks that serve as simulated proxies for industrial visual control: MetaWorld \cite{yu2020metaworld} and the DeepMind Control Suite (DMC) \cite{tassa2018dmcontrol}.
MetaWorld contains goal-conditioned robotic manipulation tasks, where performance is measured by task success rate.
DMC contains MuJoCo-based continuous-control tasks with normalized return scores, where higher scores indicate better control performance.
Using both benchmarks allows us to test whether adversarial vulnerabilities appear consistently across control regimes that are relevant to engineered automation systems, including reaching, object interaction, goal-conditioned manipulation, balance control, and locomotion.
Fig.~\ref{fig:arb4wm_framework}(f) summarizes the selected environments, and Fig.~\ref{fig:arb4wm_framework}(g) summarizes the evaluation metrics and analysis dimensions used in ARB4WM.

For MetaWorld, we select 10 manipulation tasks:
\textit{coffee-pull}, \textit{reach}, \textit{reach-wall}, \textit{window-close}, \textit{window-open}, \textit{button-press}, \textit{handle-press}, \textit{plate-slide-back}, \textit{plate-slide-back-side}, and \textit{plate-slide-side}.
These tasks cover reaching, pushing, sliding, pressing, and object manipulation under different geometric constraints.
MetaWorld places more emphasis on goal-conditioned visual manipulation and success-based evaluation, which makes it particularly relevant to industrial robotic-control settings.

For DMC, we select 10 tasks:
\textit{reacher easy}, \textit{reacher hard}, \textit{walker walk}, \textit{walker run}, \textit{walker stand}, \textit{ball-in-cup catch}, \textit{cartpole balance}, \textit{cartpole swingup}, \textit{finger turn easy}, and \textit{hopper stand}.
These tasks cover multiple control regimes, including sparse target reaching, balance control, locomotion, swing-up control, catching, and contact-rich stabilization.
This diversity is important for safety testing because adversarial perturbations can affect agents differently depending on whether the task requires precise visual localization, stable posture control, or long-horizon motion coordination.
Although these environments are simulated rather than physical industrial plants, they provide repeatable visual closed-loop tasks for isolating safety-relevant failure modes before deployment in higher-fidelity digital twins or real equipment.

\subsection{Victim Agents and Attack Protocol}

We compare four Dreamer-family agents: Dreamer \cite{hafner2020dreamer}, R2-Dreamer \cite{morihira2026r2}, Dreamer-InfoNCE \cite{oord2018representation}, and Dreamer-Pro \cite{deng2022dreamerpro}.
All agents are evaluated from visual observations under the same task set, perturbation budgets, and evaluation episodes.
The attack is applied only during evaluation and does not modify the environment dynamics, reward function, or model parameters.
We consider one random perturbation baseline and five white-box attack objectives: PDA, PEA, DCA, VSA, and LDA.
For each objective, adversarial perturbations are generated using both single-step and multi-step iterative optimizers.

For comparability, all victim agents follow the training protocol and hyperparameter setting used by R2-Dreamer \cite{morihira2026r2}.
The four agents are trained with the same training budget and use the 12M model-size configuration.
All reported attack results are averaged over 20 independent evaluation runs with random seeds from 20 to 39.
For single-step attacks, the step size is set to the perturbation budget, $\alpha=\epsilon$.
For multi-step attacks, we use 10 iterations and set the step size to $\alpha=\epsilon/10$.

Table~\ref{tab:clean_perf} reports the average clean performance of the four victim models before adversarial perturbation.
The DMC column reports normalized environment score, while the MetaWorld column reports average success rate.
These clean results provide context for interpreting robustness, since models with different nominal performance may also have different robustness profiles.

\begin{table}[t]
\centering
\caption{Average clean performance of victim models on the selected MetaWorld and DMC tasks.}
\label{tab:clean_perf}
\footnotesize
\setlength{\tabcolsep}{5pt}
\renewcommand{\arraystretch}{0.9}
\resizebox{\columnwidth}{!}{
\begin{tabular}{lcc}
\toprule
\textbf{Model} & \textbf{DMC Score} & \textbf{MetaWorld Success} \\
\midrule
Dreamer~\cite{hafner2020dreamer}         & 943.22 & 1.00 \\
R2-Dreamer~\cite{morihira2026r2}      & 928.52 & 0.98 \\
Dreamer-InfoNCE~\cite{oord2018representation} & 903.34 & 1.00 \\
Dreamer-Pro~\cite{deng2022dreamerpro}     & 859.03 & 0.97 \\
\bottomrule
\end{tabular}
}
\end{table}

\subsection{Evaluation Metrics}
Our evaluation covers five complementary aspects of robustness in world models: (i) task performance under perturbation, (ii) robustness aggregation across attack budgets, (iii) defense effectiveness and recovery, (iv) adaptive robustness gap, and (v) policy saliency behavior. Specifically, we evaluate clean performance, normalized area-under-curve robustness, defense recovery, adaptive attack gap, and saliency-based attention alignment.

% ---------------- Clean Performance ----------------
\subsubsection{Clean Performance}
Let $\mathcal T$ denote a benchmark task set and let $R_{m,\tau}(\epsilon)$ denote the evaluation return of model $m$ on task $\tau$ under perturbation budget $\epsilon$.
For DMC, $R_{m,\tau}$ is the normalized environment score.
For MetaWorld, $R_{m,\tau}$ is the success rate.
The average clean performance is computed as
\begin{equation}
\label{eq_clean_perf}
\mathrm{Clean}(m)
=
\frac{1}{|\mathcal T|}
\sum_{\tau \in \mathcal T}
R_{m,\tau}(0).
\end{equation}

% ---------------- Robustness AUC ----------------
\subsubsection{Robustness AUC}

To summarize robustness across perturbation magnitudes, we compute the area under the return curve over a set of perturbation budgets $\mathcal E=\{\epsilon_0,\epsilon_1,\ldots,\epsilon_K\}$.
For a given model $m$, task $\tau$, and attack objective $a$, the AUC is approximated by the trapezoidal rule:
\begin{equation}
\label{eq_auc}
\mathrm{AUC}
=
\sum_{k=0}^{K-1}
\frac{
R_{m,\tau,a}(\epsilon_k)
+
R_{m,\tau,a}(\epsilon_{k+1})
}{2}
(\epsilon_{k+1}-\epsilon_k).
\end{equation}

Because different models can have different clean performance, we use clean-normalized AUC (nAUC) to measure robust task performance against a fixed benchmark scale:
\begin{equation}
\label{eq_nauc}
\mathrm{nAUC}
=
\frac{
\mathrm{AUC}
}{
(\epsilon_K-\epsilon_0) C_{\mathcal B}
},
\end{equation}
where $C_{\mathcal B}$ is the benchmark normalization constant.
We use $C_{\mathcal B}=1000$ for DMC normalized scores and $C_{\mathcal B}=1$ for MetaWorld success rates.
The reported mean nAUC averages Eq.~\eqref{eq_nauc} over tasks and over single-step and multi-step iterative results for the same attack objective:
\begin{equation}
\label{eq_mean_nauc}
\overline{\mathrm{nAUC}}
=
\frac{1}{|\mathcal T||\mathcal O|}
\sum_{\tau \in \mathcal T}
\sum_{o \in \mathcal O}
\mathrm{nAUC}(\tau,o),
\end{equation}
where $\mathcal O=\{\mathrm{Single}, \mathrm{Iterative}\}$ denotes the set of attack optimizers.
Higher nAUC indicates stronger robust task performance under adversarial perturbations.

% ---------------- Defense Recovery ----------------
\subsubsection{Defense Recovery}
For defense evaluation, we further measure how much of the attack-induced performance loss is recovered by an input defense.
Let $R^{\mathrm{def}}_{m,\tau,a,d,o}(\epsilon)$ denote the return of model $m$ on task $\tau$ under attack objective $a$, defense $d$, optimizer $o$, and perturbation budget $\epsilon$.
Let $R^{\mathrm{atk}}_{m,\tau,a,o}(\epsilon)$ denote the corresponding attacked return without defense, and let $R^{\mathrm{clean}}_{m,\tau}$ denote the clean return.
The defense recovery score is defined as
\begin{equation}
\label{eq_defense_recovery}
\mathrm{Rec}(\epsilon)
=
\frac{
R^{\mathrm{def}}_{m,\tau,a,d,o}(\epsilon)
-
R^{\mathrm{atk}}_{m,\tau,a,o}(\epsilon)
}{
R^{\mathrm{clean}}_{m,\tau}
-
R^{\mathrm{atk}}_{m,\tau,a,o}(\epsilon)
}.
\end{equation}
Thus, $\mathrm{Rec}=0$ means no improvement over the undefended attack, $\mathrm{Rec}=1$ means clean performance is restored, and negative values mean that preprocessing further degrades performance.
Mean recovery is obtained by averaging Eq.~\eqref{eq_defense_recovery} over perturbation budgets, attack objectives, optimizers, tasks, and models as appropriate for each figure.

% ---------------- Adaptive Robustness Gap ----------------
\subsubsection{Adaptive Robustness Gap}
To quantify whether a defense remains effective against an adaptive attacker, we compute the adaptive gap:
\begin{equation}
\label{eq_adaptive_gap}
\mathrm{Gap}
=
\overline{\mathrm{Rec}}^{\mathrm{non\text{-}adaptive}}_{m,\tau,d}
-
\overline{\mathrm{Rec}}^{\mathrm{adaptive}}_{m,\tau,d},
\end{equation}
where the overline denotes averaging over attack objectives, perturbation budgets, and attack optimizers.
Larger values indicate that accounting for the defense more strongly reduces the apparent defense benefit.

% ---------------- Saliency Metrics ----------------
\subsubsection{Policy Saliency Metrics}
For the policy-saliency analysis, let $S(i,j)$ denote the nonnegative occlusion saliency score at pixel location $(i,j)$ after normalizing the saliency map so that $\sum_{i,j}S(i,j)=1$.
Let $\Omega$ denote the target region and let $c_\Omega$ denote the target center.
The Target Ratio measures how much saliency mass lies inside the target region:
\begin{equation}
\label{eq_target_ratio}
\mathrm{TR}
=
\sum_{(i,j)\in\Omega}
S(i,j).
\end{equation}
The Center Distance measures the Euclidean distance between the saliency centroid and the target center:
\begin{equation}
\label{eq_center_distance}
\mathrm{CD}
=
\left\|
\sum_{i,j} S(i,j) [i,j]^\top
-
c_\Omega
\right\|_2.
\end{equation}
The saliency entropy measures spatial dispersion:
\begin{equation}
\label{eq_saliency_entropy}
\mathrm{Ent}
=
-
\sum_{i,j}
S(i,j)\log(S(i,j)+\eta),
\end{equation}
where $\eta$ is a small numerical constant.
Higher Target Ratio indicates stronger target-centered evidence; lower Center Distance and Entropy indicate more concentrated saliency around the target.

\section{Experiments and Analysis}

In this section, we analyze ARB4WM results across attack objectives, temporal exposure modes, data-level defenses, transferability tests, saliency diagnostics, and the testing-system interface.

\subsection{Robustness Across Benchmarks and Attack Objectives}

\begin{figure*}[pos=tbp]
\centering

\subfloat[Dreamer]{
    \includegraphics[width=0.24\textwidth]{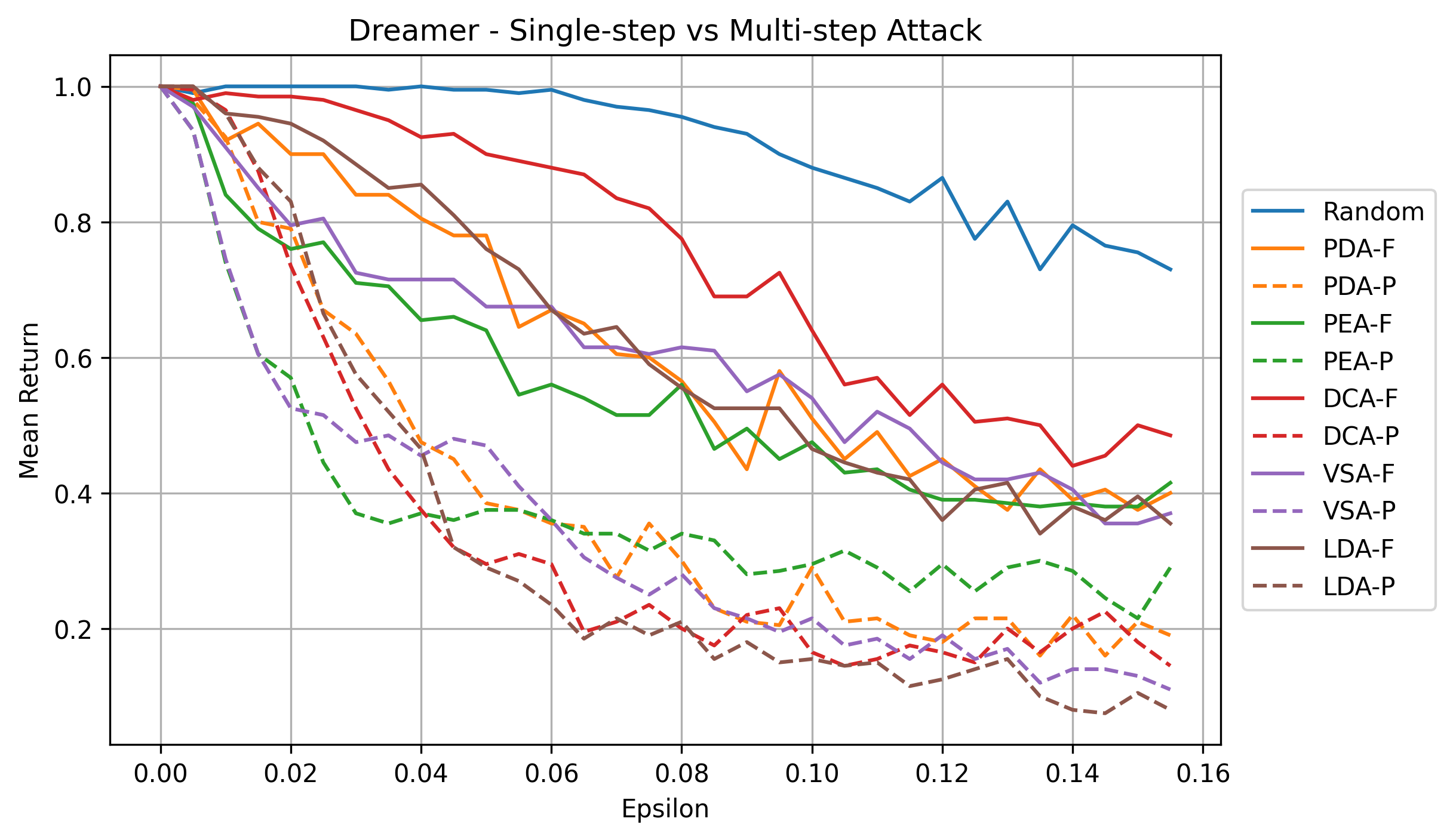}
}
\subfloat[R2-Dreamer]{
    \includegraphics[width=0.24\textwidth]{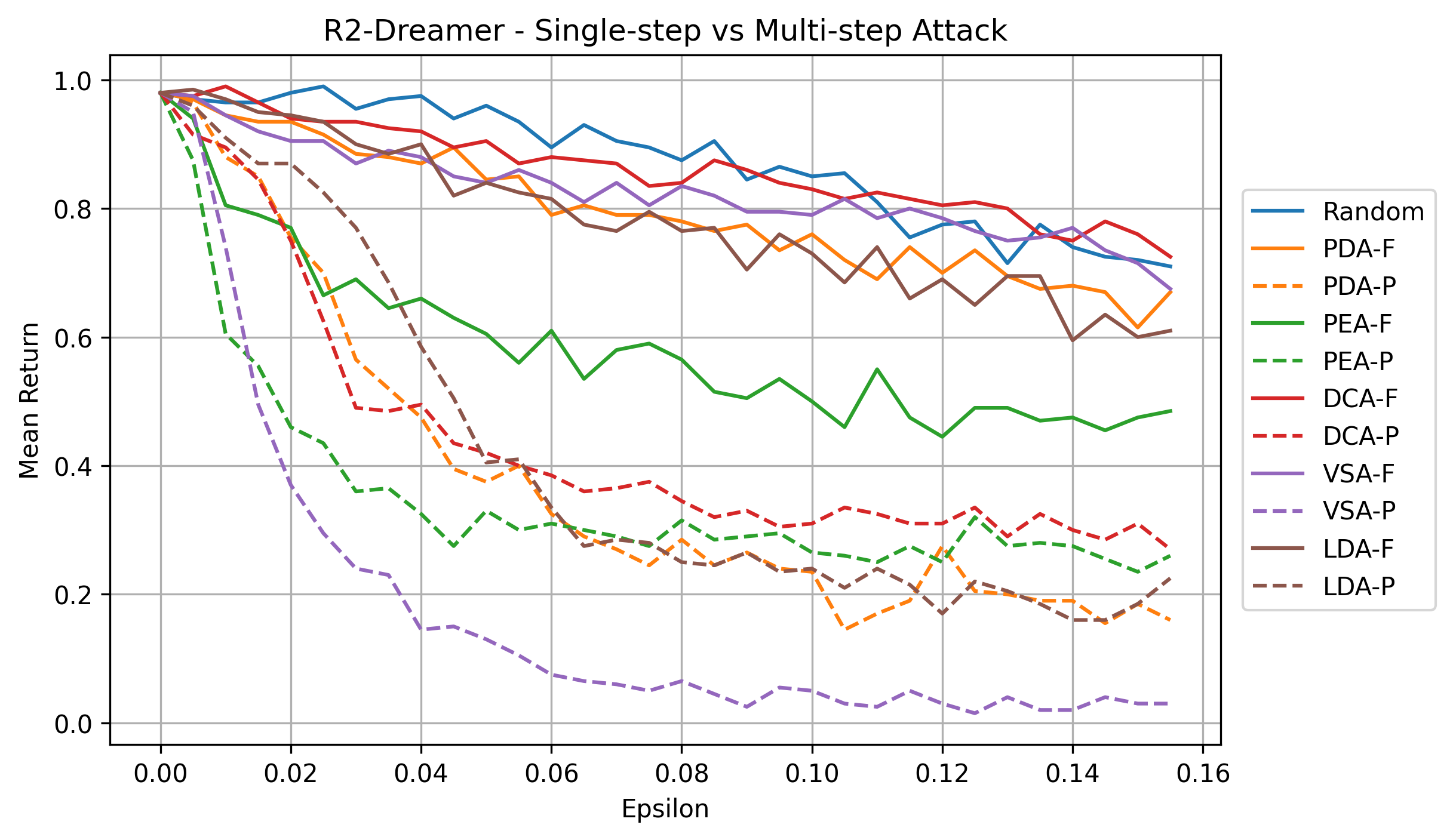}
}
\subfloat[Dreamer-InfoNCE]{
    \includegraphics[width=0.24\textwidth]{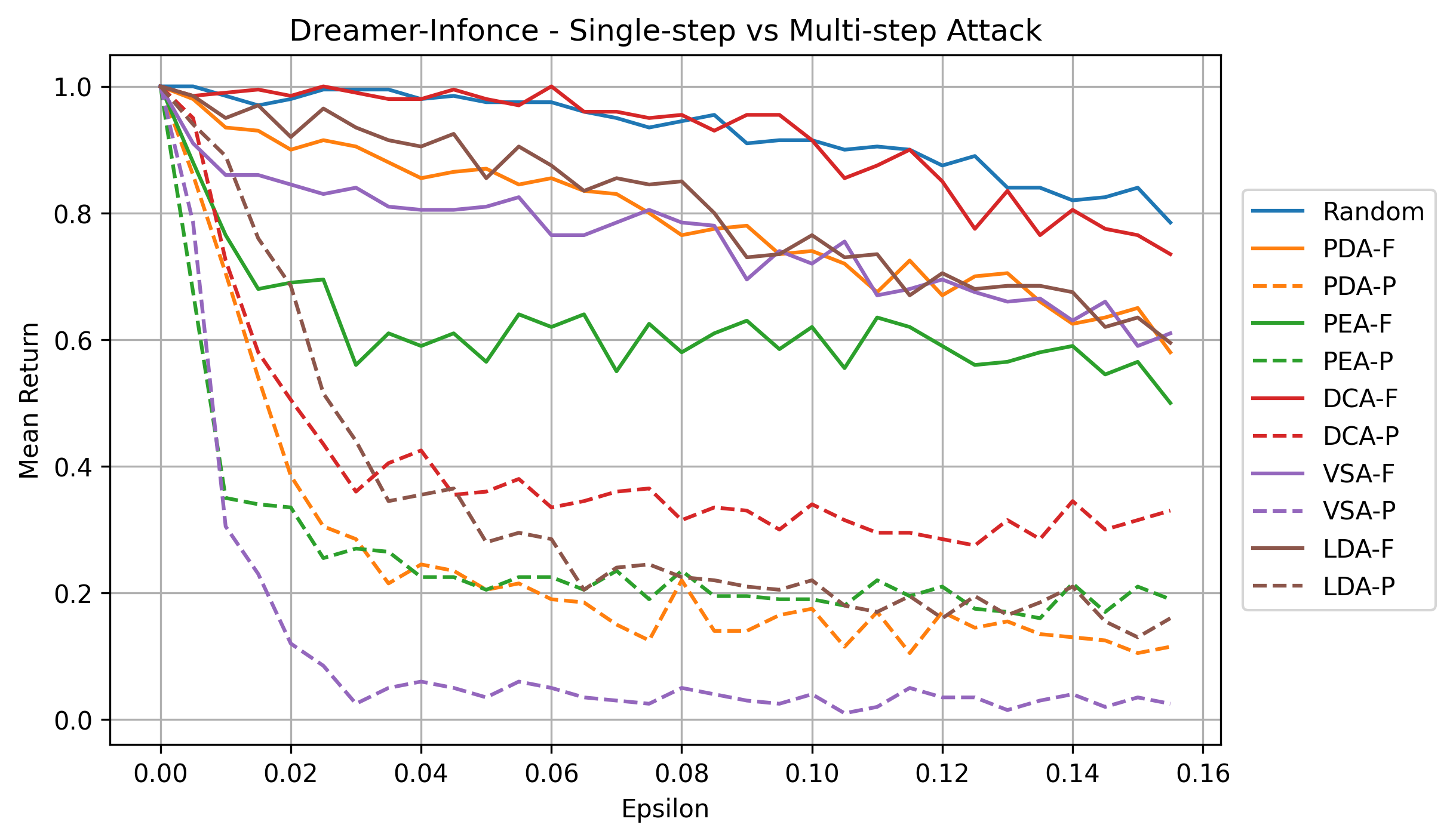}
}
\subfloat[Dreamer-Pro]{
    \includegraphics[width=0.24\textwidth]{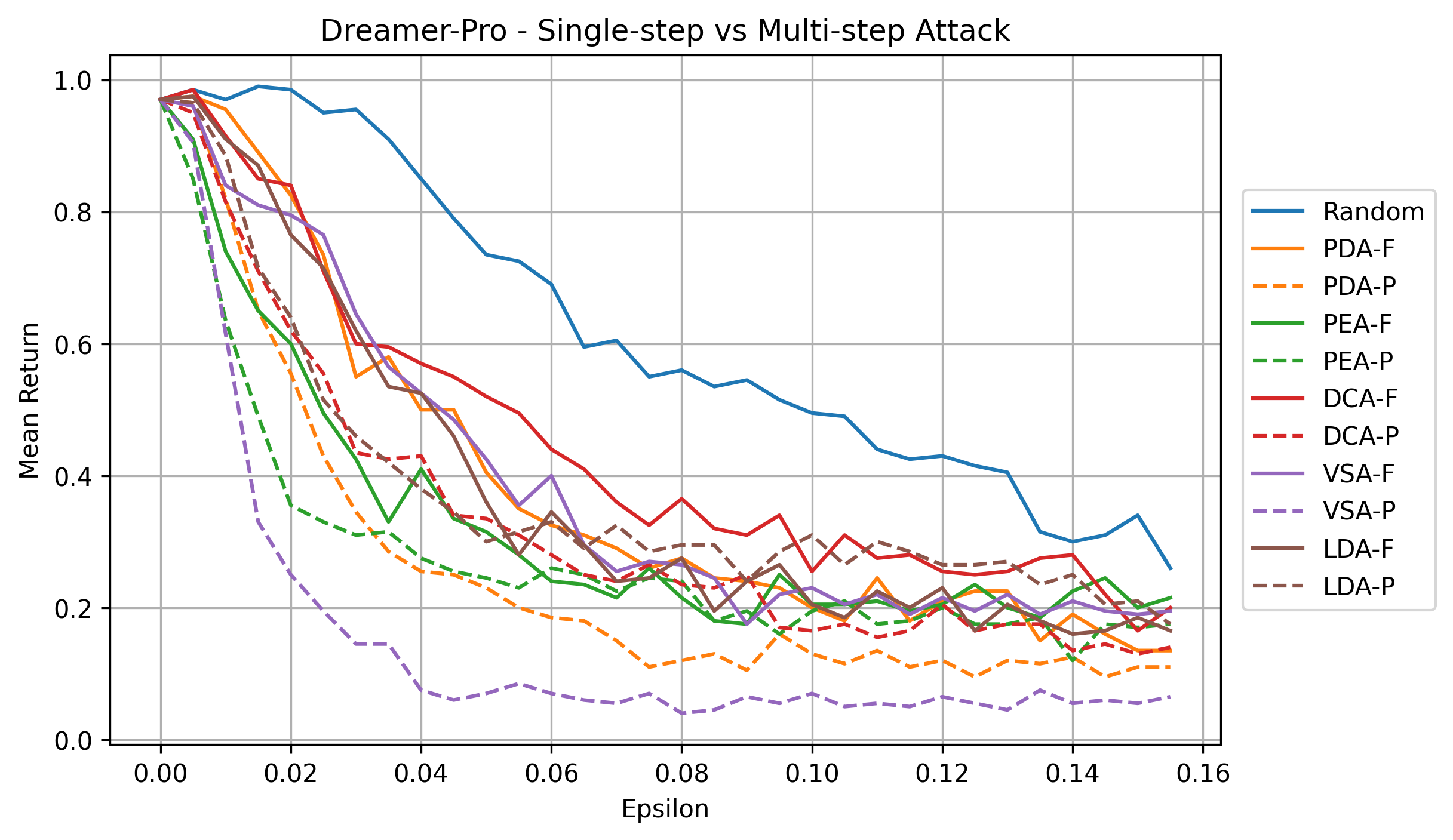}
}

\caption{
Success-rate curves under different perturbation magnitudes $\epsilon$ on 10 MetaWorld tasks.
Each subplot corresponds to one Dreamer-family agent and compares five attack objectives under single-step and multi-step iterative optimization strategies.
The x-axis denotes the perturbation budget $\epsilon$, while the y-axis denotes the average task success rate.
Lower curves indicate stronger adversarial attack effectiveness and weaker robustness.
}
\label{fig:metaworld_attack_curves}

\end{figure*}

\begin{figure*}[pos=tbp]
\centering

\subfloat[Dreamer]{
    \includegraphics[width=0.24\textwidth]{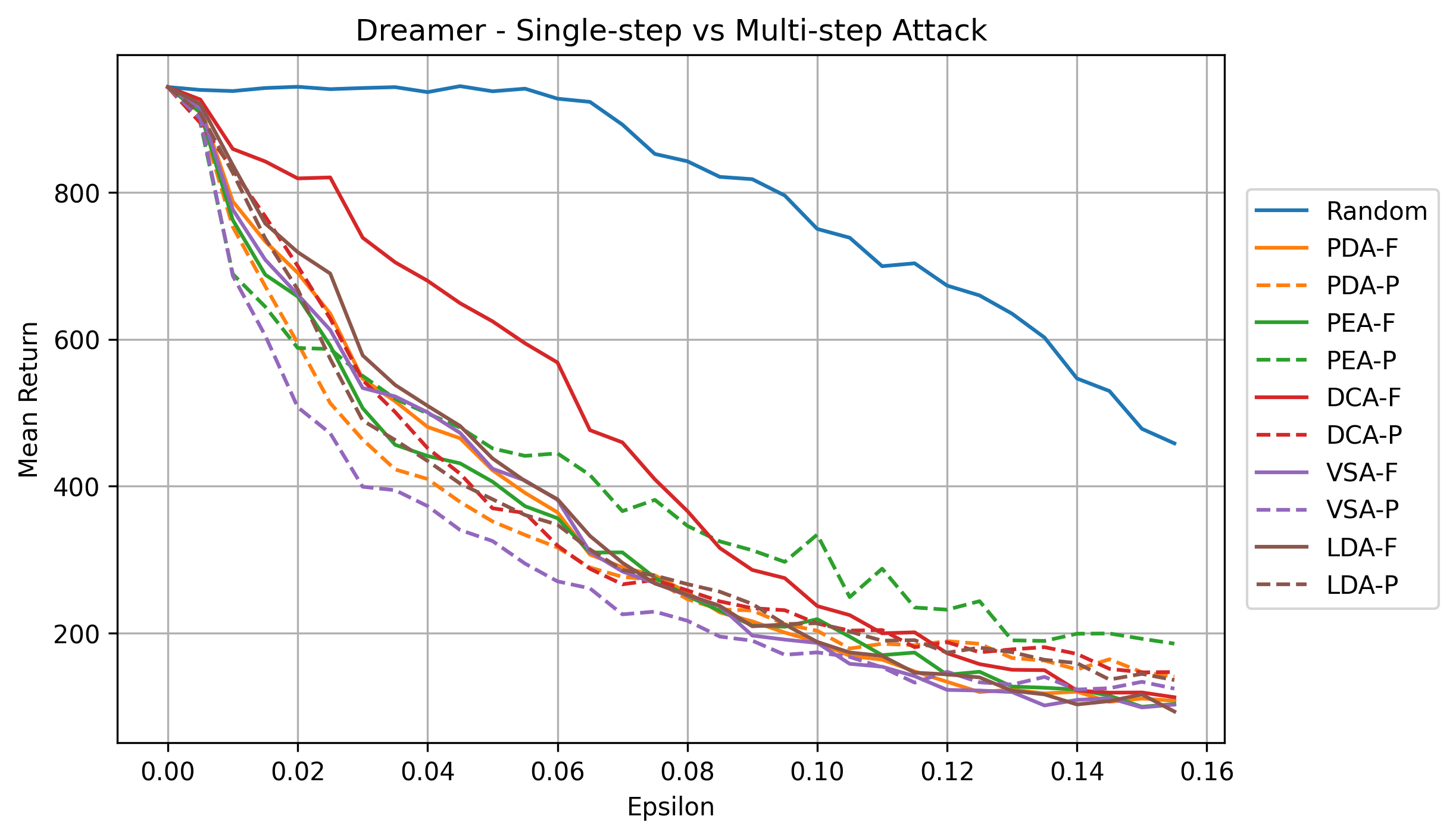}
}
\subfloat[R2-Dreamer]{
    \includegraphics[width=0.24\textwidth]{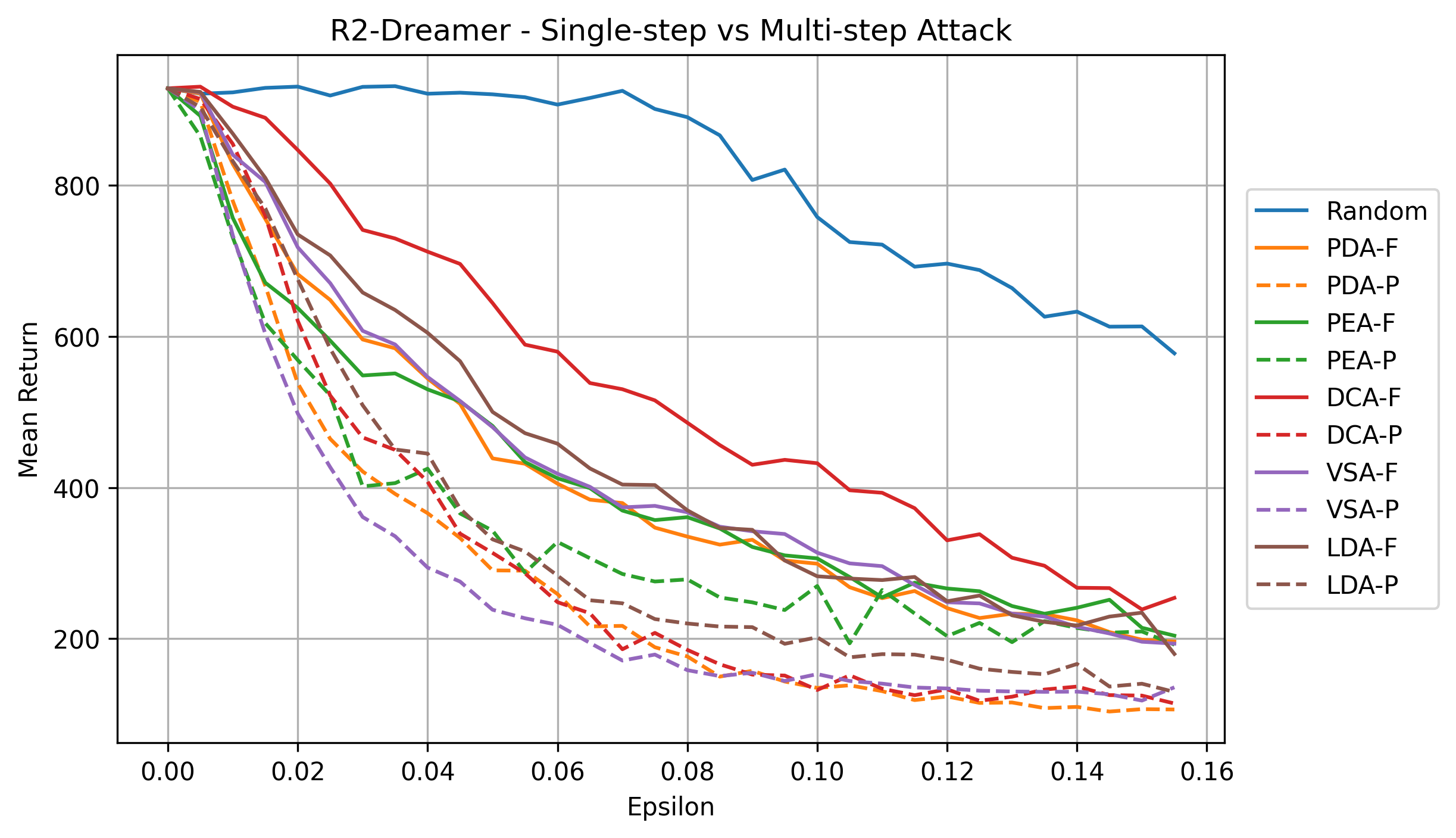}
}
\subfloat[Dreamer-InfoNCE]{
    \includegraphics[width=0.24\textwidth]{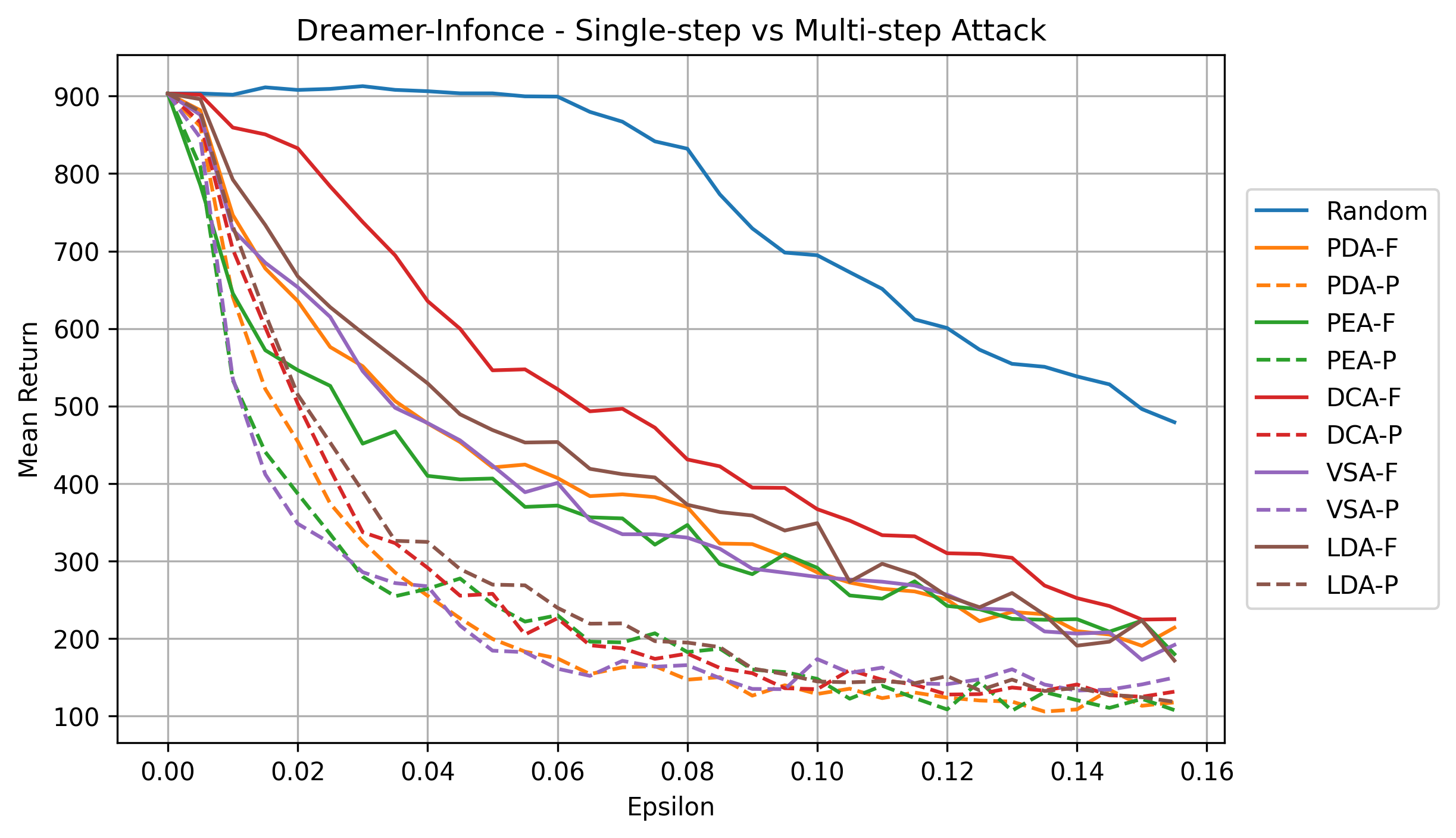}
}
\subfloat[Dreamer-Pro]{
    \includegraphics[width=0.24\textwidth]{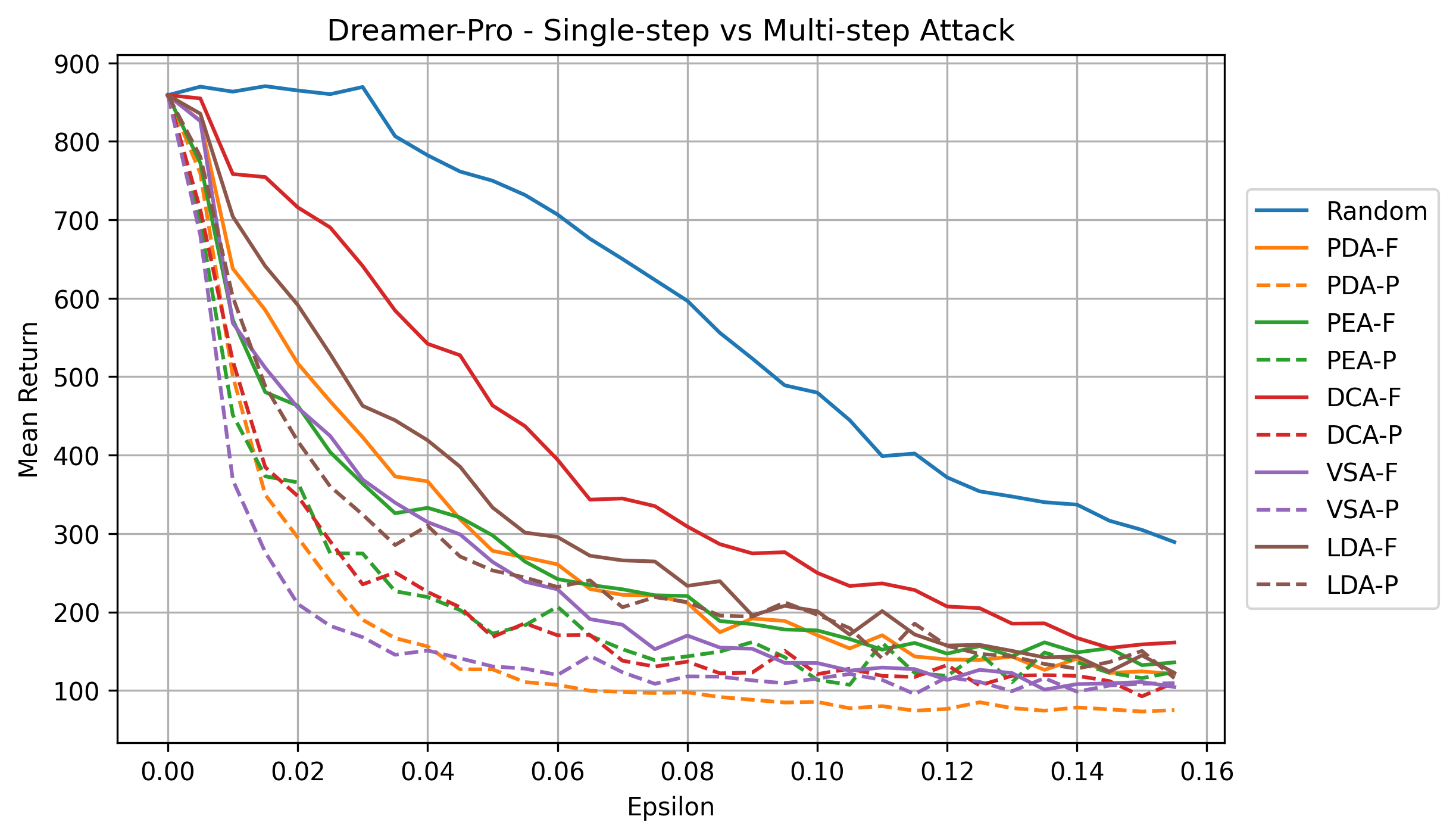}
}

\caption{
Normalized return curves under different perturbation magnitudes $\epsilon$ on 10 DMC tasks.
Each subplot corresponds to one Dreamer-family agent and compares five attack objectives under single-step and multi-step iterative optimization strategies.
The x-axis denotes the perturbation budget $\epsilon$, while the y-axis denotes the normalized environment return.
Lower curves indicate stronger adversarial attack effectiveness and weaker robustness.
}
\label{fig:dmc_attack_curves}

\end{figure*}

\begin{table}[t]
\centering
\caption{Mean clean-normalized Area Under Curve (nAUC) results on MetaWorld and DMC under different attack objectives. 
Higher nAUC indicates stronger robust task performance under attack.}
\label{tab:main_nauc}

\scriptsize
\setlength{\tabcolsep}{2pt}
\renewcommand{\arraystretch}{0.92}

\resizebox{\columnwidth}{!}{
\begin{tabular}{llccccccc}
\toprule
\textbf{Benchmark} & \textbf{Model} & \textbf{Random} & \textbf{PDA} & \textbf{PEA} & \textbf{DCA} & \textbf{VSA} & \textbf{LDA} & \textbf{Avg.} \\
\midrule
\multirow{4}{*}{MetaWorld}
& Dreamer~\cite{hafner2020dreamer}          & 0.916 & 0.509 & 0.469 & 0.548 & 0.476 & 0.478 & 0.566 \\
& R2-Dreamer~\cite{morihira2026r2}       & 0.873 & 0.587 & 0.468 & 0.651 & 0.497 & 0.597 & 0.612 \\
& Dreamer-InfoNCE~\cite{oord2018representation}  & 0.930 & 0.519 & 0.436 & 0.656 & 0.427 & 0.571 & 0.590 \\
& Dreamer-Pro~\cite{deng2022dreamerpro}      & 0.626 & 0.323 & 0.302 & 0.389 & 0.268 & 0.379 & 0.381 \\
\midrule
\multirow{4}{*}{DMC}
& Dreamer~\cite{hafner2020dreamer}          & 0.804 & 0.340 & 0.369 & 0.402 & 0.319 & 0.355 & 0.431 \\
& R2-Dreamer~\cite{morihira2026r2}       & 0.819 & 0.351 & 0.381 & 0.421 & 0.350 & 0.391 & 0.452 \\
& Dreamer-InfoNCE~\cite{oord2018representation}  & 0.763 & 0.319 & 0.304 & 0.382 & 0.313 & 0.354 & 0.406 \\
& Dreamer-Pro~\cite{deng2022dreamerpro}      & 0.597 & 0.225 & 0.243 & 0.301 & 0.210 & 0.292 & 0.311 \\
\bottomrule
\end{tabular}
}
\end{table}

\begin{table}[t]
\centering
\caption{Number of model--task pairs for which each attack objective achieves the lowest mean nAUC. The count is computed over four models and 20 tasks. Lower nAUC indicates stronger attack performance.}
\label{tab:strongest_attack_counts}
\begin{tabular}{lcc}
\toprule
\textbf{Attack objective} & \textbf{Strongest count} & \textbf{Share} \\
\midrule
PDA & 12 / 80 & 15.0\% \\
PEA & 28 / 80 & 35.0\% \\
DCA & 4 / 80 & 5.0\% \\
VSA & 32 / 80 & 40.0\% \\
LDA & 4 / 80 & 5.0\% \\
\bottomrule
\end{tabular}
\end{table}

Figs.~\ref{fig:metaworld_attack_curves} and~\ref{fig:dmc_attack_curves} reveal a consistent robustness gap between random visual noise and optimized adversarial perturbations.
Although the perturbation budgets are small, the gradient-based attacks reduce returns much more strongly than the random baseline across both benchmarks.
This gap indicates that the failure mode is not simply sensitivity to image corruption.
Instead, the attacks exploit structured directions in observation space that are aligned with the agent's policy, value, latent representation, or dynamics computation.
For world models, this is especially important because the same perturbed observation can affect the current action and also update the recurrent belief state used by later decisions.

The model ranking in Table~\ref{tab:main_nauc} suggests that representation design has a measurable effect on adversarial robustness.
R2-Dreamer \cite{morihira2026r2} achieves the strongest average nAUC on both MetaWorld and DMC, while Dreamer-Pro \cite{deng2022dreamerpro} is consistently the weakest model despite being designed to improve representation learning through prototypical structure.
This contrast suggests that nominal representation quality and adversarial stability are not equivalent objectives.
An agent can learn features that are effective for clean control while still relying on visual directions that are easy to manipulate adversarially.
Dreamer-InfoNCE \cite{oord2018representation} performs competitively on MetaWorld but is weaker on DMC, indicating that the robustness effect of contrastive representation learning is task-dependent rather than uniformly beneficial.

The attack-objective trends also show why a single policy-level attack is insufficient for evaluating world models.
VSA is highly damaging in several settings, which indicates that corrupting value estimates can destabilize behavior even when the attack is applied only to observations.
This is consistent with Dreamer-style control, where values are used to evaluate imagined trajectories and guide actor learning.
Table~\ref{tab:strongest_attack_counts} provides a task-level count of the strongest attack objective across the 80 model--task pairs.
For each pair, we average the nAUC of each objective over the single-step and multi-step optimizers and select the objective with the lowest mean nAUC.
VSA is the strongest objective in 32 cases, and PEA is strongest in 28 cases, indicating that value suppression and policy-entropy disruption are the most frequent high-impact objectives in this benchmark.
DCA and LDA remain competitive attack directions because they target RSSM consistency and latent-state stability rather than the final action distribution alone.
These results support the central design choice of ARB4WM: robustness evaluation for world-model agents should cover the latent decision-making pipeline, including policy outputs, critic estimates, latent representations, and dynamics consistency.

\subsection{Temporal Vulnerability}

\begin{table}[t]
\centering
\caption{Mean clean-normalized Area Under Curve (nAUC) results on four DMC tasks under different temporal attack modes.
The full-frame row uses the main robustness experiments, restricted to the perturbation budgets shared with the partial-frame experiments, $\epsilon \in \{0, 0.04, 0.08, 0.12\}$.
Higher nAUC indicates stronger robust task performance under attack.}
\label{tab:temporal_attack_modes}

\scriptsize
\setlength{\tabcolsep}{2pt}
\renewcommand{\arraystretch}{0.92}

\resizebox{\columnwidth}{!}{
\begin{tabular}{llccccccc}
\toprule
\textbf{Attack Mode} & \textbf{Model} & \textbf{Random} & \textbf{PDA} & \textbf{PEA} & \textbf{DCA} & \textbf{VSA} & \textbf{LDA} & \textbf{Avg.} \\
\midrule
\multirow{4}{*}{Full-frame}
& Dreamer~\cite{hafner2020dreamer}          & 0.911 & 0.521 & 0.516 & 0.585 & 0.470 & 0.534 & 0.589 \\
& R2-Dreamer~\cite{morihira2026r2}       & 0.912 & 0.487 & 0.500 & 0.607 & 0.442 & 0.540 & 0.581 \\
& Dreamer-InfoNCE~\cite{oord2018representation}  & 0.872 & 0.438 & 0.385 & 0.513 & 0.399 & 0.464 & 0.512 \\
& Dreamer-Pro~\cite{deng2022dreamerpro}      & 0.864 & 0.395 & 0.376 & 0.505 & 0.355 & 0.478 & 0.496 \\
\midrule
\multirow{4}{*}{First-half}
& Dreamer~\cite{hafner2020dreamer}          & 0.917 & 0.710 & 0.712 & 0.748 & 0.668 & 0.710 & 0.744 \\
& R2-Dreamer~\cite{morihira2026r2}       & 0.935 & 0.692 & 0.707 & 0.763 & 0.678 & 0.720 & 0.749 \\
& Dreamer-InfoNCE~\cite{oord2018representation}  & 0.903 & 0.629 & 0.609 & 0.686 & 0.616 & 0.646 & 0.681 \\
& Dreamer-Pro~\cite{deng2022dreamerpro}      & 0.905 & 0.631 & 0.630 & 0.703 & 0.610 & 0.671 & 0.692 \\
\midrule
\multirow{4}{*}{Second-half}
& Dreamer~\cite{hafner2020dreamer}          & 0.942 & 0.766 & 0.810 & 0.802 & 0.735 & 0.770 & 0.804 \\
& R2-Dreamer~\cite{morihira2026r2}       & 0.936 & 0.749 & 0.820 & 0.826 & 0.737 & 0.776 & 0.808 \\
& Dreamer-InfoNCE~\cite{oord2018representation}  & 0.926 & 0.755 & 0.791 & 0.803 & 0.703 & 0.772 & 0.792 \\
& Dreamer-Pro~\cite{deng2022dreamerpro}      & 0.919 & 0.703 & 0.781 & 0.783 & 0.684 & 0.769 & 0.773 \\
\midrule
\multirow{4}{*}{Sparse ($n=2$)}
& Dreamer~\cite{hafner2020dreamer}          & 0.921 & 0.743 & 0.745 & 0.803 & 0.680 & 0.754 & 0.774 \\
& R2-Dreamer~\cite{morihira2026r2}       & 0.922 & 0.719 & 0.738 & 0.804 & 0.678 & 0.762 & 0.771 \\
& Dreamer-InfoNCE~\cite{oord2018representation}  & 0.913 & 0.711 & 0.670 & 0.773 & 0.616 & 0.698 & 0.730 \\
& Dreamer-Pro~\cite{deng2022dreamerpro}      & 0.915 & 0.685 & 0.627 & 0.757 & 0.583 & 0.776 & 0.724 \\
\midrule
\multirow{4}{*}{Sparse ($n=4$)}
& Dreamer~\cite{hafner2020dreamer}          & 0.966 & 0.908 & 0.931 & 0.950 & 0.845 & 0.916 & 0.920 \\
& R2-Dreamer~\cite{morihira2026r2}       & 0.954 & 0.933 & 0.936 & 0.959 & 0.890 & 0.953 & 0.937 \\
& Dreamer-InfoNCE~\cite{oord2018representation}  & 0.953 & 0.893 & 0.877 & 0.925 & 0.886 & 0.917 & 0.909 \\
& Dreamer-Pro~\cite{deng2022dreamerpro}      & 0.946 & 0.899 & 0.887 & 0.926 & 0.906 & 0.929 & 0.916 \\
\midrule
\multirow{4}{*}{Sparse ($n=6$)}
& Dreamer~\cite{hafner2020dreamer}          & 0.966 & 0.949 & 0.947 & 0.963 & 0.891 & 0.952 & 0.945 \\
& R2-Dreamer~\cite{morihira2026r2}       & 0.962 & 0.950 & 0.949 & 0.964 & 0.940 & 0.964 & 0.955 \\
& Dreamer-InfoNCE~\cite{oord2018representation}  & 0.954 & 0.926 & 0.905 & 0.942 & 0.934 & 0.933 & 0.932 \\
& Dreamer-Pro~\cite{deng2022dreamerpro}      & 0.955 & 0.941 & 0.923 & 0.946 & 0.944 & 0.945 & 0.942 \\
\bottomrule
\end{tabular}
}
\end{table}

Table~\ref{tab:temporal_attack_modes} isolates a failure mode that is specific to recurrent world-model agents: adversarial observations can contaminate the latent state over time.
Full-frame attacks are the most damaging setting because every belief update is computed from a perturbed visual input.
This means that the attacker does not need to rely only on immediate action errors; it can also maintain pressure on the RSSM posterior, deterministic recurrent state, value prediction, and subsequent policy inputs throughout the episode.
The lower nAUC under full-frame attacks therefore reflects a compounding process rather than a collection of independent single-step mistakes.

The difference between first-half and second-half attacks further supports this recurrent-state interpretation.
Perturbing the first half of the episode is generally more damaging than perturbing the second half, even though both settings attack the same number of frames.
Early perturbations occur before the agent has stabilized its internal belief and trajectory, so errors injected into the latent state can influence later decisions after the attack has stopped.
By contrast, second-half attacks affect a shorter remaining horizon and begin after the agent has already accumulated a cleaner history.
This asymmetry shows why robustness for world models should be evaluated temporally, not only frame by frame.

Sparse attacks reveal a frequency threshold for maintaining adversarial influence.
When every other frame is attacked ($n=2$), the degradation remains substantial, suggesting that the recurrent state does not fully recover between adjacent clean frames.
However, as the interval increases to $n=4$ and $n=6$, the nAUC approaches the random baseline.
This indicates that clean observations can partially re-anchor the latent belief state when they arrive frequently enough.
The temporal results therefore refine the main robustness findings: attack strength depends not only on the objective and perturbation budget, but also on whether perturbations are frequent enough and early enough to propagate through the RSSM memory.

\begin{figure}[pos=t]
\centering
\includegraphics[width=0.49\textwidth]{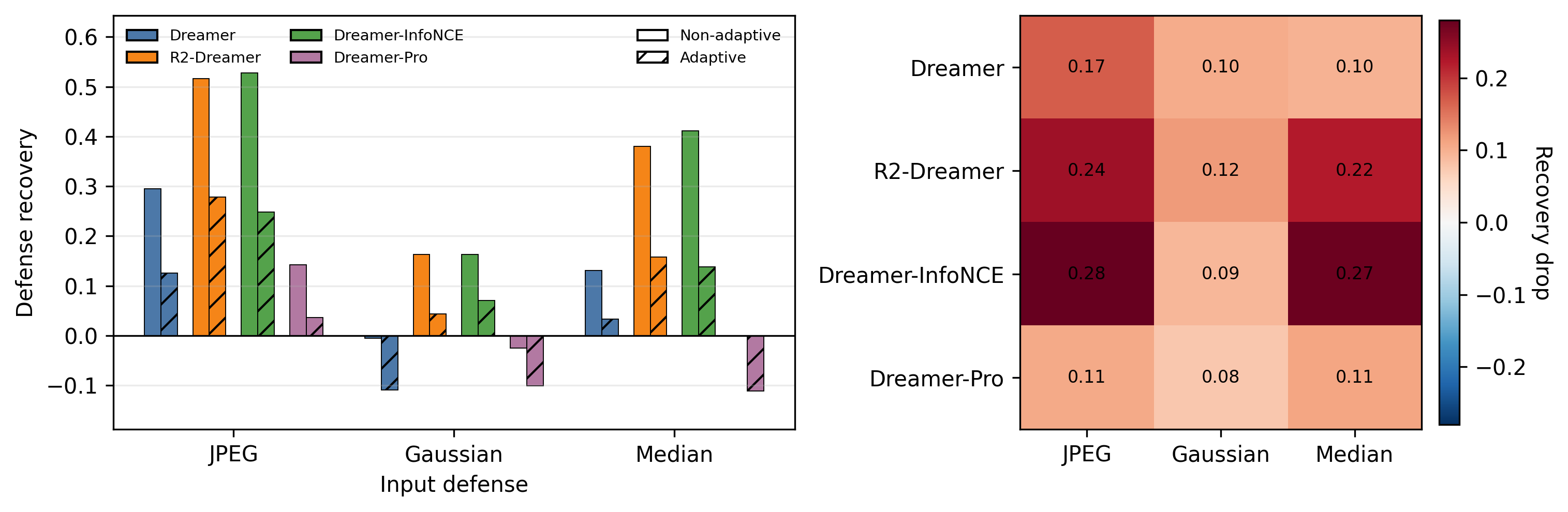}
\caption{
Defense recovery under non-adaptive and adaptive attacks, averaged over eight representative tasks from MetaWorld and DMC.
Panel (a) reports defense recovery, and panel (b) reports the adaptive gap.
Higher recovery indicates stronger mitigation, while a larger adaptive gap indicates weaker protection under defense-aware attacks.
}
\label{fig:defense_recovery_average}
\end{figure}

\begin{figure}[pos=t]
\centering
\includegraphics[width=0.49\textwidth]{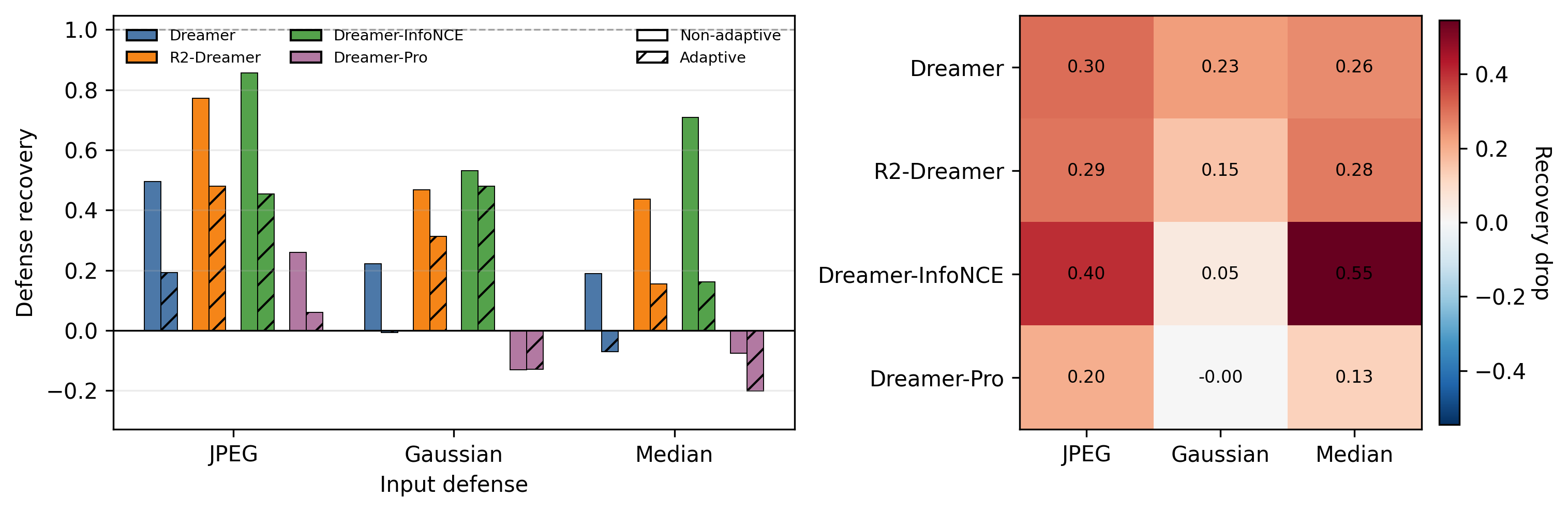}
\caption{
Task-level defense recovery on Plate Slide Back, the task with the highest average recovery among the eight evaluated tasks.
}
\label{fig:defense_recovery_plate_slide_back}
\end{figure}

\begin{figure}[pos=th]
\centering
\includegraphics[width=0.49\textwidth]{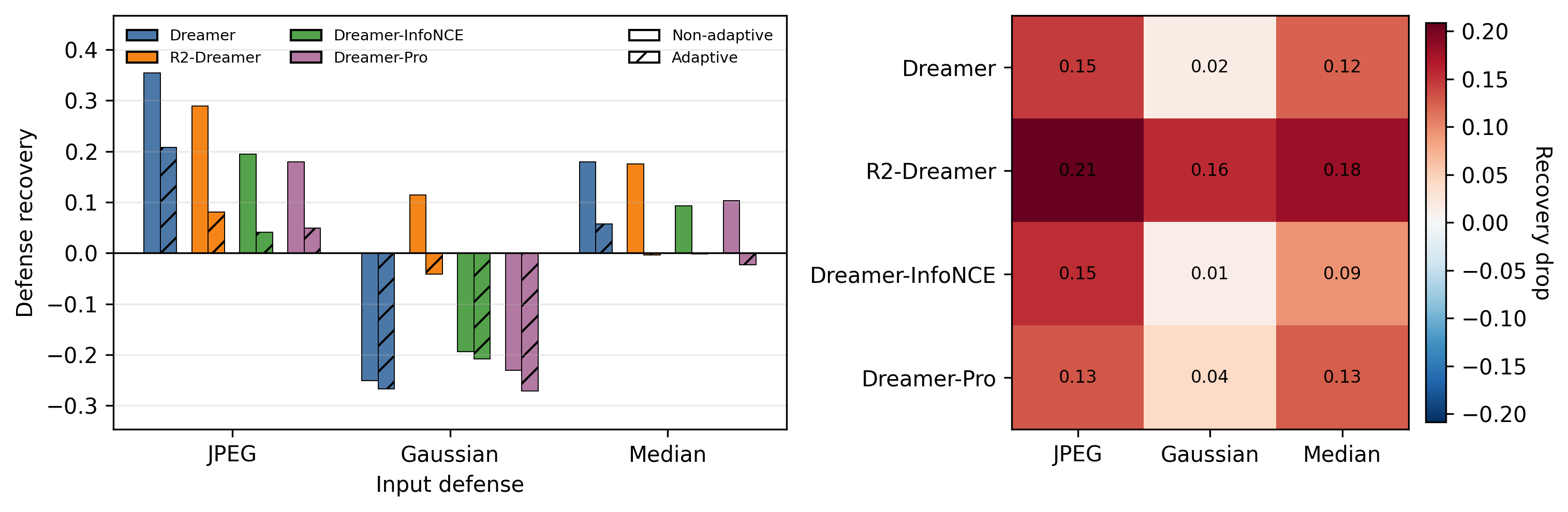}
\caption{
Task-level defense recovery on Cartpole Balance, the task with the lowest average recovery among the eight evaluated tasks.
}
\label{fig:defense_recovery_cartpole_balance}
\end{figure}

\subsection{Robustness under Data-Level Defenses}
Fig.~\ref{fig:defense_recovery_average} shows that simple input transformations can recover part of the attack-induced performance loss, but the recovery is incomplete and highly sensitive to the attacker's knowledge.
Under non-adaptive attacks, JPEG compression and median filtering obtain the largest $\mathrm{Rec}$ values because they can remove high-frequency perturbation patterns while preserving much of the low-frequency visual structure needed for control.
Gaussian noise is weaker because it does not selectively remove adversarial directions and can also corrupt task-relevant pixels.
These results indicate that data-level preprocessing can be useful when the attack is generated against the undefended observation pipeline.

The adaptive setting gives a stricter test.
Once the attacker differentiates through, or approximates gradients through, the preprocessing step, the perturbation can be optimized to survive or exploit the transformation.
The drop from non-adaptive $\mathrm{Rec}$ to adaptive $\mathrm{Rec}$ is summarized by $\mathrm{Gap}$ in Fig.~\ref{fig:defense_recovery_average}(b).
JPEG compression and median filtering show larger $\mathrm{Gap}$ values than Gaussian noise because they impose structured, deterministic transformations that an adaptive attacker can model.
Gaussian noise has a smaller $\mathrm{Gap}$, but this does not imply strong defense: its absolute recovery is weak and sometimes negative.
Thus, none of the tested input defenses eliminates the underlying adversarial vulnerability; they only partially change the perturbation distribution seen by the model.

\begin{figure*}[t]
\centering
\subfloat[Source-target transfer matrix]{
    \includegraphics[width=0.32\textwidth]{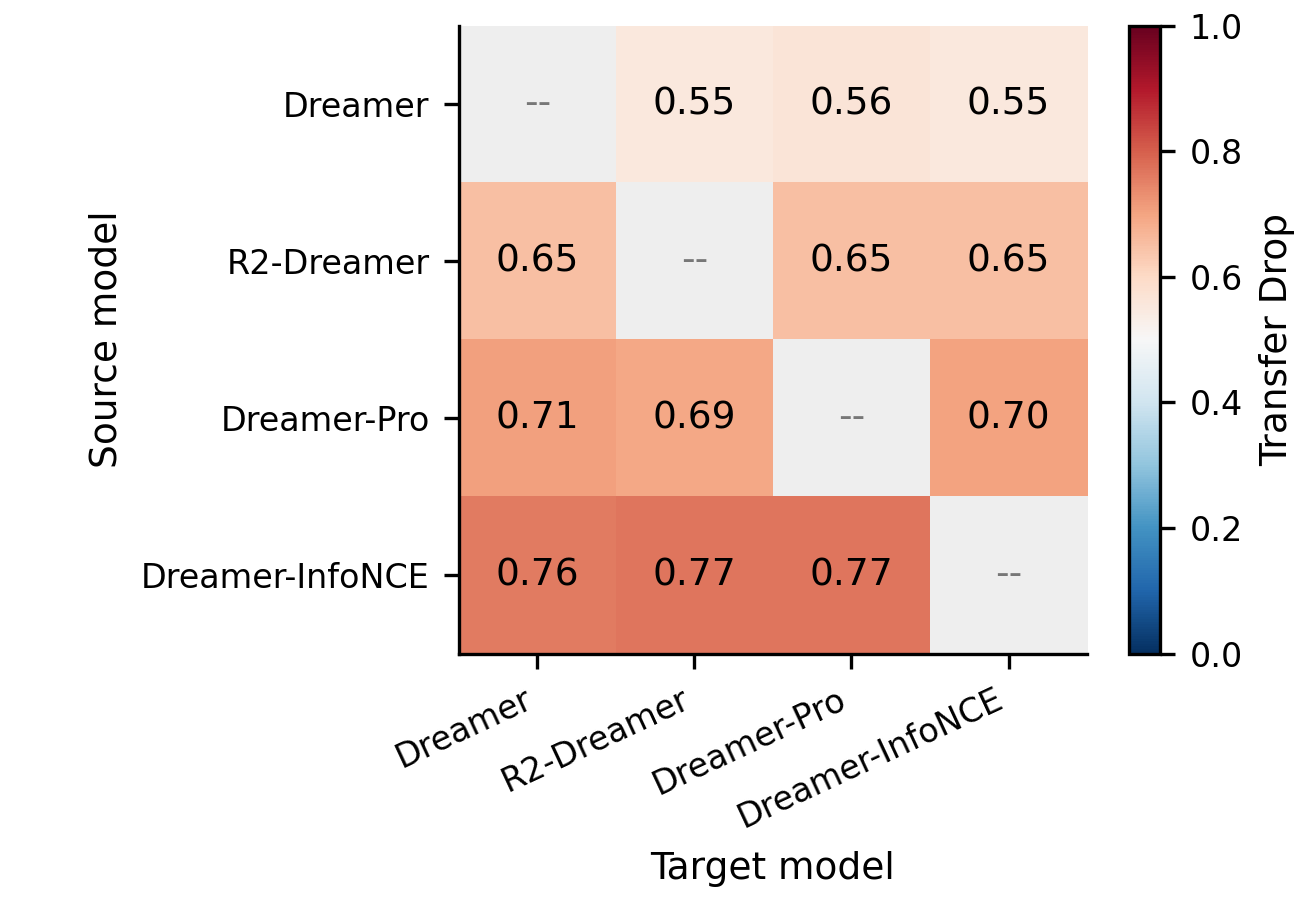}
}
\subfloat[Objective-wise transfer strength]{
    \includegraphics[width=0.32\textwidth]{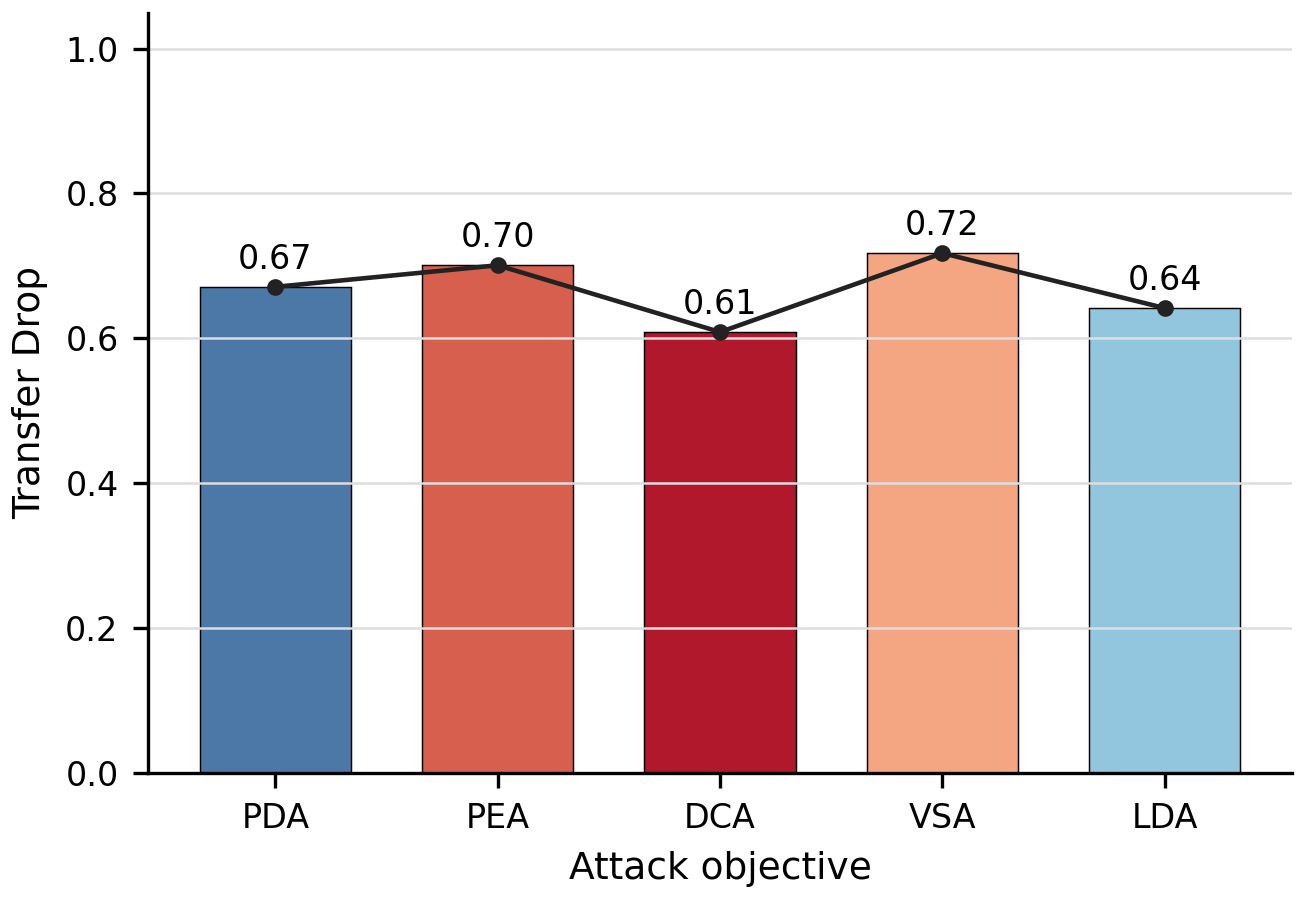}
}
\subfloat[Objective transfer by target model]{
    \includegraphics[width=0.32\textwidth]{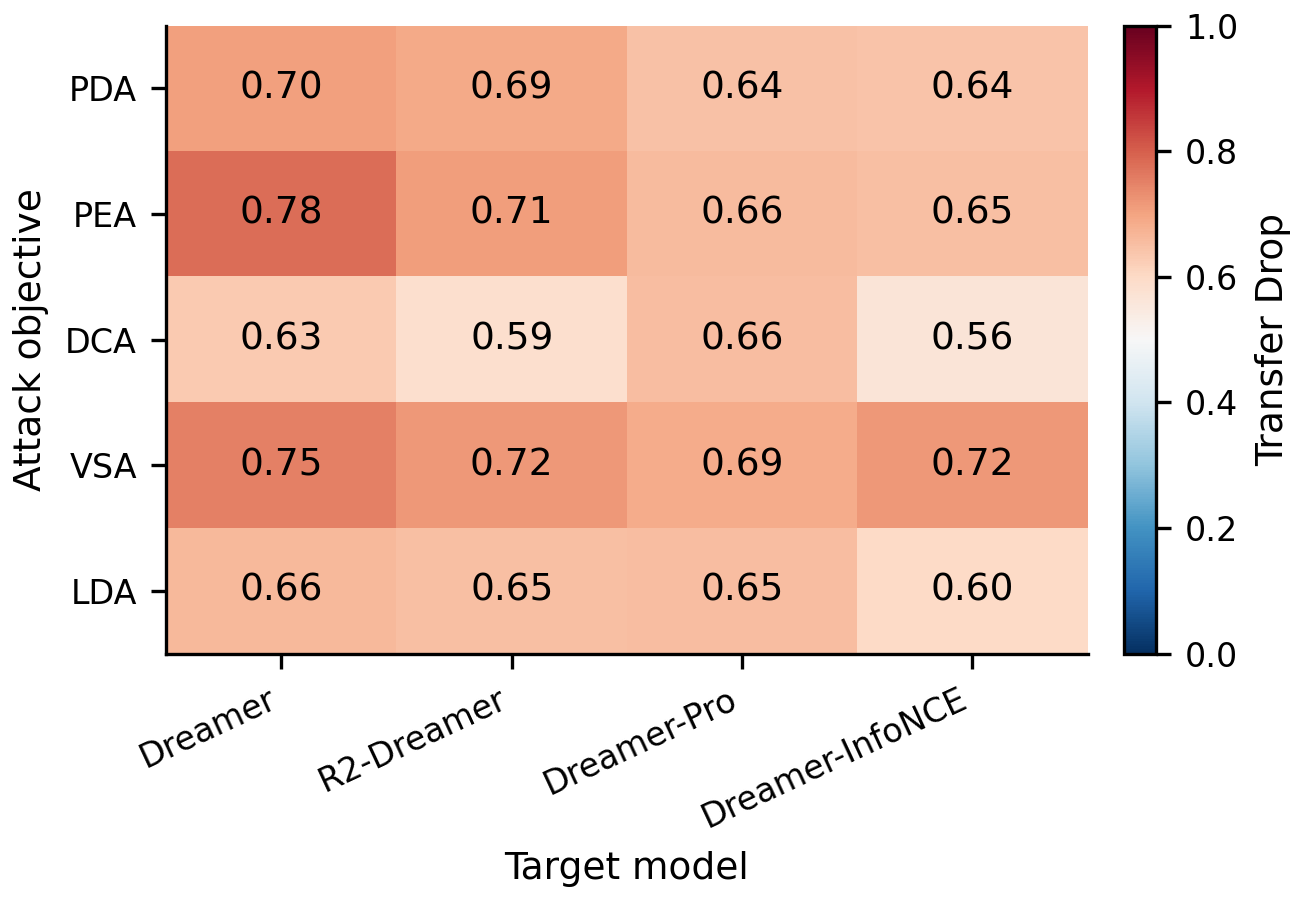}
}
\caption{
Multi-step transferability of adversarial objectives on DMC Reacher Easy.
Perturbations are generated on a source model with MI-FGSM and evaluated on different target models.
Higher Transfer Drop indicates stronger black-box degradation.
Panel (a) averages over attack objectives and perturbation budgets for each source-target pair; diagonal entries are omitted.
Panel (b) ranks the five attack objectives by average transfer strength.
Panel (c) separates the same objective scores by target model, showing that VSA and PEA transfer more strongly than DCA in this task.
}
\label{fig:reacher_transfer_attack}
\end{figure*}

The contrast between the representative tasks in Fig.~\ref{fig:defense_recovery_plate_slide_back} and Fig.~\ref{fig:defense_recovery_cartpole_balance} indicates that defense effectiveness depends on the visual and control demands of the task.
Plate Slide Back benefits more from preprocessing, likely because its undefended adversarial failures are partly driven by perturbations to object and goal appearance that can be smoothed while retaining enough coarse spatial structure for sliding control.
The task also has redundant visual cues: the plate, table, and goal geometry remain recognizable after JPEG compression or median filtering, so the defended observation can recover part of the clean policy evidence.
Cartpole Balance is different.
In the undefended setting, adversarial perturbations can quickly damage the visual estimate of pole angle and cart position, and the control loop has little tolerance for delay or smoothing error.
Preprocessing may remove some adversarial noise, but it can also blur the same thin pole and cart features that the agent needs for stabilization.
This task dependence suggests that input defenses should not be evaluated only through average recovery.
A defense can help in manipulation tasks with redundant visual cues while failing in balance or precision-control tasks where the agent depends on subtle image features.
Overall, the defense study reinforces the main benchmark conclusion: adversarial robustness for world models cannot be solved by generic image preprocessing alone and must be evaluated with adaptive, task-aware attacks.

\subsection{Transferability of Adversarial Objectives}

The preceding results evaluate attacks in a white-box setting.
We further test whether perturbations generated on one Dreamer-family agent transfer to other agents on DMC Reacher Easy.
For each source model, we generate multi-step perturbations with MI-FGSM \cite{dong2018boosting} using the five ARB4WM objectives and evaluate the same perturbed observations on held-out target models.
We report Transfer Drop, defined as $1-\mathrm{nAUC}$, where nAUC is computed over $\epsilon\in\{0,0.02,0.04,0.08,0.12,0.16\}$ and normalized by the DMC score range of 1000.

Fig.~\ref{fig:reacher_transfer_attack} shows that transferability is not uniform across source models or objectives.
Perturbations generated from Dreamer-InfoNCE transfer most strongly to the other agents, with average Transfer Drop around 0.76--0.77 across the three target models.
Dreamer as a source produces weaker transferred attacks, with average Transfer Drop around 0.55--0.56.
Across objectives, VSA and PEA have the largest average transfer effect, while DCA is the weakest in this setting.
This suggests that attacks targeting value prediction and policy entropy capture failure directions that are less tied to a single source model than RSSM consistency perturbations on DMC Reacher Easy.

\subsection{Policy Saliency under Adversarial Objectives}

To complement the return-based robustness evaluation, we further analyze how adversarial objectives affect the spatial evidence used by the policy.
We use an occlusion-based saliency method on DMC Reacher Easy.
Starting from the same rollout, we select the 20th frame and compute saliency maps for four Dreamer-family agents under clean observations and multi-step iterative attacks with $\epsilon=0.06$ and 10 optimization steps.
The saliency score at each image location is defined by the change in the policy action mean after occluding a local image patch.
This analysis does not measure trajectory-level performance directly; instead, it visualizes where each policy becomes sensitive under different attack objectives.

\begin{figure}[pos=tbp]
\centering
\includegraphics[width=\linewidth]{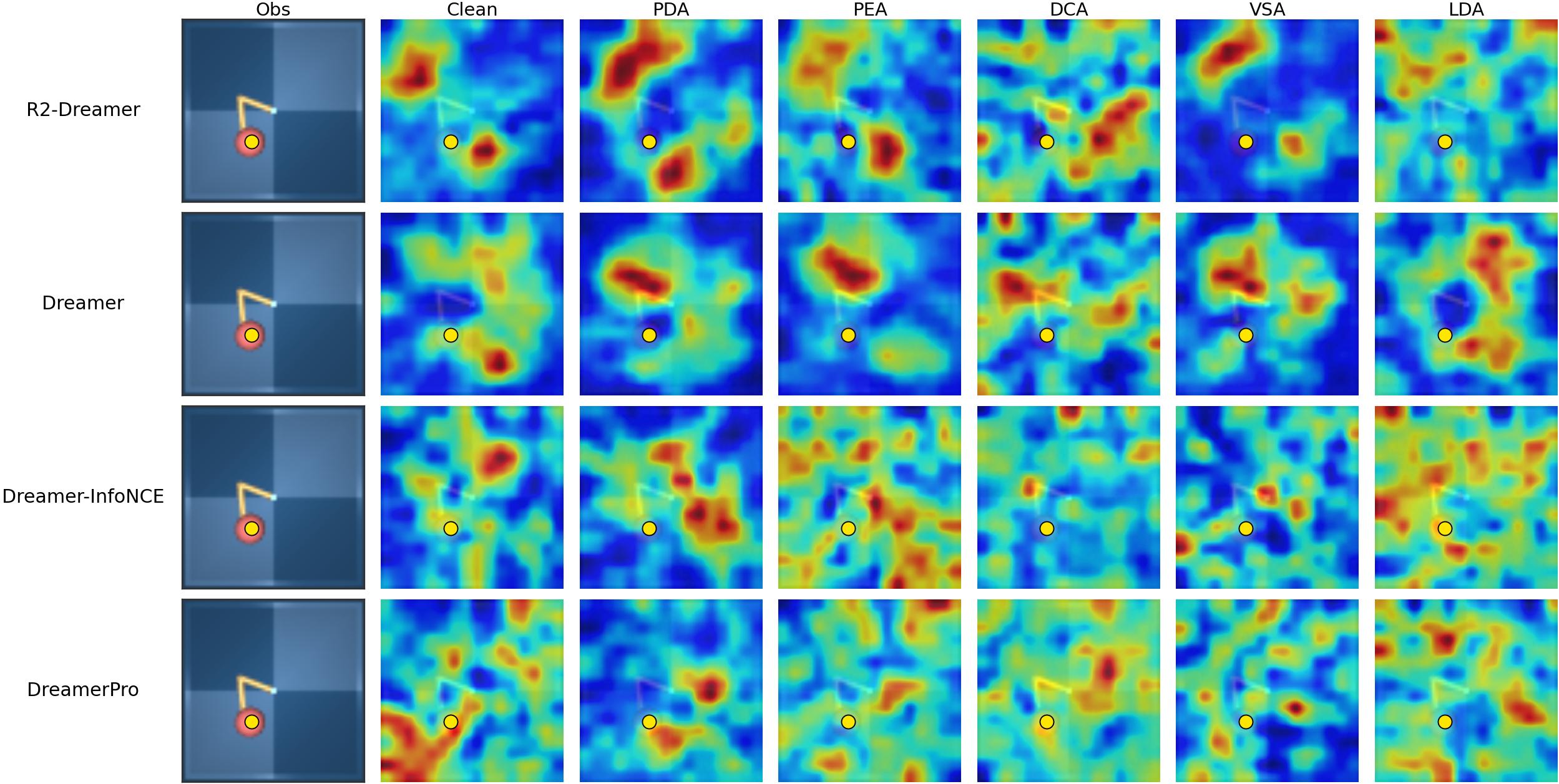}
\caption{
Policy saliency maps on DMC Reacher Easy at the 20th frame of a shared rollout.
Rows correspond to Dreamer-family agents, and columns compare clean observations with adversarial observations generated by PDA, PEA, DCA, VSA, and LDA.
The yellow dot marks the target position.
The visualization shows that different attack objectives alter the spatial sensitivity of the policy in distinct ways, even when the underlying environment frame appears visually similar.
}
\label{fig:policy_saliency_frame20}
\end{figure}

\begin{table*}[t]
\centering
\caption{
Policy saliency metrics on DMC Reacher Easy at frame 20.
Target Ratio measures the fraction of saliency mass inside the target region; higher is better.
Center Distance measures the distance between the saliency centroid and the target; lower is better.
Entropy measures saliency dispersion; lower values indicate more concentrated saliency.
Model rows report clean metrics and the average over five adversarial objectives.
Attack rows report the average over four models.
}
\label{tab:policy_saliency_metrics}
\scriptsize
\setlength{\tabcolsep}{3pt}
\renewcommand{\arraystretch}{0.92}
\resizebox{\textwidth}{!}{
\begin{tabular}{llccccccccc}
\toprule
\textbf{Group} & \textbf{Name}
& \textbf{Clean Target Ratio} & \textbf{Attack Target Ratio} & \textbf{$\Delta$ Target Ratio}
& \textbf{Clean Center Dist.} & \textbf{Attack Center Dist.} & \textbf{$\Delta$ Center Dist.}
& \textbf{Clean Entropy} & \textbf{Attack Entropy} & \textbf{$\Delta$ Entropy} \\
\midrule
\multirow{4}{*}{Model}
& R2-Dreamer~\cite{morihira2026r2}       & 0.0301 & 0.0181 & -0.0119 & 12.2265 & 14.1417 &  1.9152 & 0.9799 & 0.9813 &  0.0015 \\
& Dreamer~\cite{hafner2020dreamer}          & 0.0342 & 0.0209 & -0.0133 & 15.3066 & 14.2989 & -1.0077 & 0.9794 & 0.9780 & -0.0013 \\
& Dreamer-InfoNCE~\cite{oord2018representation}  & 0.0347 & 0.0246 & -0.0101 & 14.6282 & 13.3585 & -1.2697 & 0.9818 & 0.9867 &  0.0048 \\
& Dreamer-Pro~\cite{deng2022dreamerpro}      & 0.0370 & 0.0283 & -0.0087 & 10.8590 & 13.7272 &  2.8683 & 0.9859 & 0.9886 &  0.0026 \\
\midrule
\multirow{5}{*}{Attack}
& PDA & -- & 0.0238 & -0.0102 & -- & 13.5156 & 0.2606 & -- & 0.9776 & -0.0042 \\
& PEA & -- & 0.0226 & -0.0114 & -- & 13.6084 & 0.3533 & -- & 0.9858 &  0.0041 \\
& DCA & -- & 0.0219 & -0.0121 & -- & 14.1949 & 0.9398 & -- & 0.9878 &  0.0061 \\
& VSA & -- & 0.0224 & -0.0116 & -- & 14.0942 & 0.8391 & -- & 0.9790 & -0.0027 \\
& LDA & -- & 0.0242 & -0.0098 & -- & 13.9948 & 0.7397 & -- & 0.9880 &  0.0062 \\
\bottomrule
\end{tabular}
}
\end{table*}

Fig.~\ref{fig:policy_saliency_frame20} shows that adversarial objectives can change the spatial sensitivity of the policy even when the visual frame itself remains almost unchanged.
This distinction is important.
The attacks are bounded pixel perturbations, so the arm configuration and target location in the displayed observation can look visually identical across columns.
However, the saliency maps are computed from the policy response after local occlusion, and therefore reveal how the internal decision function uses different image regions under each attack objective.
The figure shows that the attacks do not merely add visually imperceptible noise; they also change which parts of the observation become causally relevant to the policy output.

The model-wise patterns suggest that representation-learning choices affect not only return robustness but also the spatial organization of policy evidence.
For R2-Dreamer \cite{morihira2026r2}, the clean saliency is not perfectly concentrated on the target, but several attacked maps still preserve visible sensitivity around the target-arm region.
This is consistent with the main benchmark results, where R2-Dreamer obtains the strongest average robustness across MetaWorld and DMC.
Dreamer \cite{hafner2020dreamer} exhibits more objective-dependent behavior: some attacks preserve localized sensitivity, whereas others move salient regions toward background structures.
Dreamer-InfoNCE \cite{oord2018representation} and Dreamer-Pro \cite{deng2022dreamerpro} show comparatively diffuse maps under several objectives, indicating that adversarial perturbations can make their policy response depend on broader visual regions rather than a compact task-relevant area.
These differences suggest that nominal representation objectives can shape the geometry of policy sensitivity, even when the four agents share the same Dreamer-style RSSM decision pipeline.

Table~\ref{tab:policy_saliency_metrics} provides a quantitative summary of these observations.
Across all four models, the average Target Ratio decreases under attack, indicating that adversarial perturbations reduce the fraction of saliency mass assigned to the target region.
This decrease is modest in absolute value because the target occupies only a small part of the $64\times64$ image, but the trend is consistent across models.
The Center Distance metric provides a complementary view: R2-Dreamer and Dreamer-Pro show positive average shifts under attack, meaning that the saliency centroid moves farther from the target, while Dreamer and Dreamer-InfoNCE show negative average shifts on this particular frame.
This does not necessarily mean that Dreamer and Dreamer-InfoNCE are more robust; rather, it indicates that attacks can also concentrate saliency on localized non-target regions or move the centroid closer while still disrupting the policy-relevant evidence.
Therefore, Target Ratio, Center Distance, and Entropy should be interpreted jointly rather than as independent robustness scores.

The attack-wise summary further shows that different objectives perturb the policy in different ways.
DCA produces the largest average increase in Center Distance, suggesting that attacking RSSM posterior-prior consistency tends to move policy sensitivity away from the target more strongly than policy-level objectives.
VSA also produces a large Center Distance increase, which is consistent with the return-based observation that value-oriented attacks can be damaging: corrupting value estimates can make the policy sensitive to regions that are not directly tied to target localization.
LDA yields the smallest Target Ratio drop but one of the largest Entropy increases, suggesting a different failure mode in which saliency becomes more spatially diffuse rather than simply displaced.
PDA and PEA cause smaller centroid shifts on average, but they still reduce Target Ratio, showing that direct policy attacks can weaken target-centered evidence without necessarily producing the largest spatial displacement.

Overall, this saliency analysis provides a diagnostic bridge between pixel-level perturbations and trajectory-level robustness degradation.
It supports the central claim of ARB4WM that adversarial vulnerability in world models is not limited to the final policy head.
Attacks targeting policy distributions, value prediction, latent representations, and RSSM dynamics can induce different spatial failure modes, which are not fully captured by return curves alone.
At the same time, because the analysis is based on one representative frame from DMC Reacher Easy, we treat it as complementary evidence rather than a standalone robustness metric.
The return-based nAUC results remain the primary robustness measure, while saliency helps explain how different attack objectives alter the visual evidence used by the agent.

\subsection{Robustness Testing System Interface}

\begin{figure*}[pos=tbp]
\centering
\includegraphics[width=\linewidth]{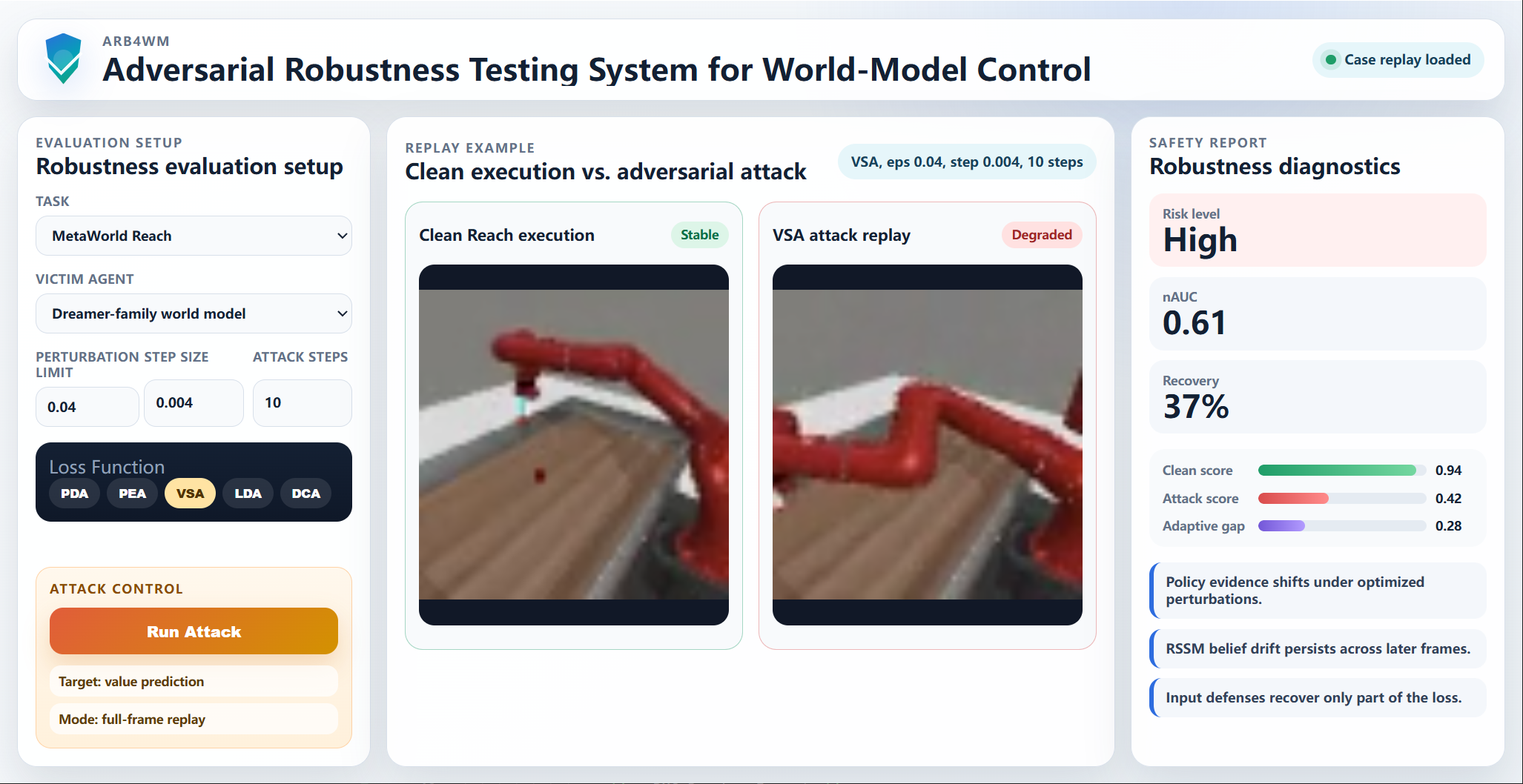}
\caption{
Robustness testing system interface for ARB4WM.
The system integrates task and agent selection, adversarial-attack parameter configuration, loss-function selection, clean-versus-attacked rollout replay, and safety-report indicators.
The displayed example uses a MetaWorld reach task with a value-suppression attack, showing how the framework connects adversarial visual perturbations with closed-loop behavior degradation and diagnostic robustness metrics.
}
\label{fig:testing_system_interface}
\end{figure*}

Fig.~\ref{fig:testing_system_interface} shows the system interface used to organize the robustness evaluation and present the selected attack case.
The left panel configures the task, victim world-model agent, perturbation budget, step size, attack iterations, and component-level loss function.
The center panel compares clean and attacked rollouts for the selected task, allowing the user to inspect how visually small perturbations change closed-loop behavior.
The right panel summarizes the corresponding safety report, including robustness score, recovery estimate, adaptive gap, and diagnostic findings.
The interface is designed as a reporting layer for ARB4WM: it does not replace the underlying attack implementation, but presents the attack configuration, replay evidence, and robustness indicators in a form suitable for model comparison and engineering safety review.

\section{Model-wise Robustness Ranking and Mechanistic Interpretation}
\label{sec:robustness_ranking_interpretation}

In this section, we summarize model-wise robustness rankings and interpret the observed differences through the shared encoder--RSSM--head inference structure.

\begin{figure}[pos=htbp]
\centering
\includegraphics[width=\linewidth]{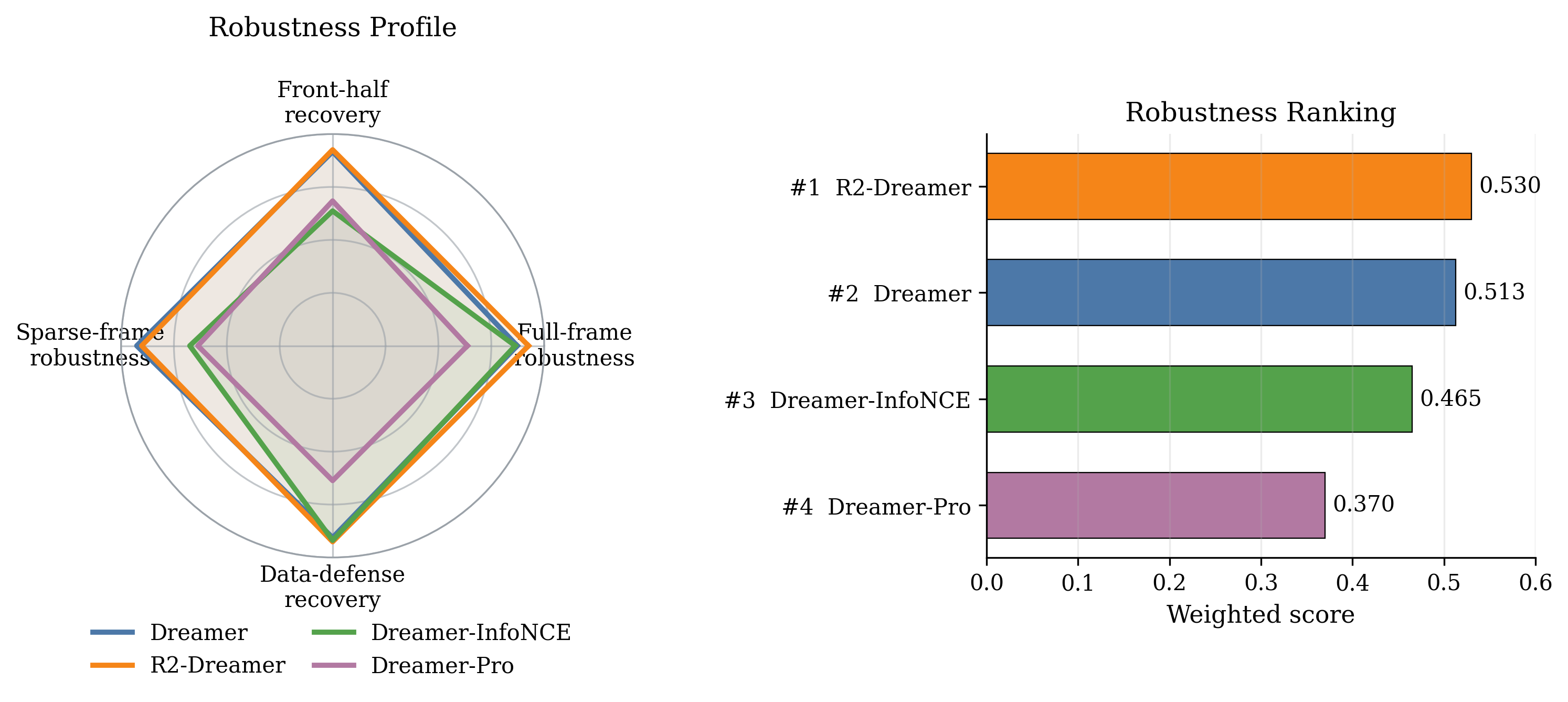}
\caption{
Model-wise robustness overview with full-frame robustness computed over all 20 selected DMC and MetaWorld tasks.
The radar plots summarize four complementary robustness dimensions: full-frame attack nAUC over the full benchmark, first-half attack nAUC as a recovery proxy after perturbations stop, sparse-frame robustness with attacks every two frames, and input-defense recovery.
The first-half, sparse-frame, and input-defense dimensions use the four DMC tasks from the temporal and defense analyses.
For visualization and ranking, each dimension is min--max normalized across models and then mapped to the observed full-frame nAUC range, so the displayed scores remain on an nAUC-like scale rather than an exaggerated $[0,1]$ scale.
The weighted score assigns the largest weight to full-frame robustness ($0.55$) and equal weights to first-half recovery, sparse-frame robustness, and input-defense recovery ($0.15$ each).
}
\label{fig:robustness_overview_ranking}
\end{figure}

Fig.~\ref{fig:robustness_overview_ranking} summarizes model-wise robustness after the detailed attack, temporal, defense, and saliency analyses.
This ranking is not intended to replace the per-objective tables.
Instead, it provides a compact view of how each model behaves under four complementary conditions that stress different parts of the same inference pipeline.
R2-Dreamer and Dreamer form the leading group in this overview.
R2-Dreamer obtains the highest weighted score because the 20-task full-frame average and input-defense recovery outweigh Dreamer's slightly stronger sparse-frame nAUC on the four-task temporal subset.
Dreamer-InfoNCE and Dreamer-Pro trail the leading pair, consistent with their weaker DMC robustness in Table~\ref{tab:main_nauc}.

The four agents share the same inference structure in our implementation:
\begin{equation}
\label{eq_shared_inference}
e_t = E_\theta(o_t), \quad
s_t = F_\phi(s_{t-1}, a_{t-1}, e_t), \quad
u_t = H_\psi(s_t),
\end{equation}
where $E_\theta$ is the visual encoder, $F_\phi$ is the RSSM update, and $H_\psi$ denotes the actor, critic, or attack-relevant prediction head.
The architectural differences among the four agents are introduced mainly through the representation objective used during training, not through a different adversarial inference path.
Therefore, the robustness ranking should be interpreted as a consequence of how the training objective shapes the local geometry of $E_\theta$ and the RSSM features consumed by $H_\psi$.

For a small visual perturbation $\delta_t$, a first-order expansion gives
\begin{equation}
\label{eq_single_step_jacobian}
\Delta u_t
\approx
J_H(s_t) J_F(s_{t-1}, a_{t-1}, e_t) J_E(o_t) \delta_t,
\end{equation}
where $J_E$, $J_F$, and $J_H$ are the corresponding Jacobians.
For recurrent world models, the perturbation also propagates through future belief states:
\begin{equation}
\label{eq_recurrent_jacobian}
\Delta u_t
\approx
\sum_{k \le t}
J_H
\left(
\prod_{j=k+1}^{t} A_j
\right)
B_k J_E(o_k)\delta_k,
\end{equation}
where $A_j=\partial s_j/\partial s_{j-1}$ and $B_k=\partial s_k/\partial e_k$.
Equations~\eqref{eq_single_step_jacobian} and~\eqref{eq_recurrent_jacobian} show why full-frame attacks receive the largest ranking weight: they repeatedly inject perturbations into every term of the recurrent sum, so the attacker can affect both the current action and the latent memory used by later decisions.
They also show why models with identical inference graphs can have different robustness: the learned training objective changes the norms, correlations, and conditioning of $J_E$, $B_k$, and the feature directions used by $H_\psi$.

The standard Dreamer variant adds an image reconstruction objective through the decoder, encouraging the encoder to retain pixel information while the RSSM KL, reward, continuation, actor, and value losses keep the latent state control-relevant.
This may help preserve task-relevant visual cues under intermittent perturbations, consistent with Dreamer's strong sparse-frame nAUC in Fig.~\ref{fig:robustness_overview_ranking}.
Because reconstruction does not explicitly decorrelate latent dimensions or constrain the input Jacobian, this advantage should still be interpreted empirically.

R2-Dreamer adds a Barlow-Twins-style redundancy-reduction loss between projected RSSM features and detached encoder embeddings:
\begin{equation}
\label{eq_barlow_loss}
\mathcal L_{\mathrm{R2}}
=
\sum_i (C_{ii}-1)^2
\;+\;
\lambda \sum_{i\ne j} C_{ij}^2,
\end{equation}
where $C$ is the cross-correlation matrix between normalized projected features and normalized embeddings.
The diagonal term preserves feature--embedding alignment, while the off-diagonal term penalizes redundant feature dimensions.
By discouraging multiple latent coordinates from carrying the same perturbation-sensitive factor, this objective can improve conditioning under the linearized model in Eq.~\eqref{eq_recurrent_jacobian}.
This provides a plausible explanation for R2-Dreamer's stronger first-half and defense-recovery behavior: after perturbations stop or are partially removed by preprocessing, a less redundant latent basis can re-anchor the belief state more effectively.

Dreamer-InfoNCE uses the same projector branch but replaces the redundancy-reduction loss with an instance-discrimination objective:
\begin{equation}
\label{eq_infonce_loss}
\mathcal L_{\mathrm{InfoNCE}}
=
-
\log
\frac{\exp(q_i^\top k_i)}
{\sum_j \exp(q_i^\top k_j)} ,
\end{equation}
where $q_i$ is a projected RSSM feature and $k_j$ is a detached encoder embedding from the batch.
This objective encourages features to separate positive samples from many negatives.
Such separation can improve discriminability, but may also create sharper feature gradients near negative-sample boundaries.
This helps explain Dreamer-InfoNCE's intermediate ranking: its contrastive geometry can learn informative features without necessarily suppressing adversarially exploitable directions in the control-relevant latent state.

Dreamer-Pro introduces prototype-based representation learning with normalized observation and feature projections, low-temperature prototype logits, Sinkhorn assignments, EMA targets, and translation augmentation:
\begin{equation}
\label{eq_proto_logits}
p(c \mid o_t)
\propto
\exp
\left(
\frac{\langle g(E_\theta(o_t)), p_c\rangle}{\tau}
\right).
\end{equation}
Prototype structure can improve semantic grouping, but it also creates assignment boundaries in the projected feature space.
With a small temperature $\tau$, small changes in $E_\theta(o_t)$ can produce large logit changes near a boundary, altering the feature direction passed to the RSSM and policy/value heads.
This mechanism is consistent with Dreamer-Pro's low ranking and negative input-defense recovery, where preprocessing may shift observations across prototype boundaries or distort prototype-aligned cues.

This interpretation does not claim that one auxiliary representation objective is universally more robust than another.
It instead explains the observed ranking through the shared inference graph and the training-induced geometry of each representation.
Full-frame robustness depends strongly on recurrent amplification in Eq.~\eqref{eq_recurrent_jacobian}, and the full benchmark average in Fig.~\ref{fig:robustness_overview_ranking} favors R2-Dreamer.
Recovery-oriented settings depend more on how quickly the RSSM belief returns to a stable region after perturbations stop or are transformed, which also favors the redundancy-reduced geometry learned by R2-Dreamer.
The main conclusion is that adversarial robustness depends not only on clean representation quality, but also on the local geometry of the encoder--RSSM--head composition induced by the training loss.

\section{Engineering Implications}
\label{sec:engineering_implications}

In this section, we discuss what the benchmark results imply for safety assessment and pre-deployment testing of world-model controllers in engineered continuous-control systems.

The results have three implications for safety assessment of world-model controllers in engineered continuous-control systems.
First, robustness should be evaluated as a pipeline property.
The experiments show that value suppression, latent drift, and dynamics-consistency attacks can degrade control performance even when the attack is not designed to directly change the final action distribution.
For an industrial controller, this means that testing only the policy output may miss failures in the critic, recurrent belief state, or learned transition model that later influence closed-loop behavior.

Second, temporal exposure matters for risk assessment.
Full-frame attacks produce the strongest degradation, but first-half and sparse attacks reveal whether early corruption persists in the recurrent latent state and whether clean observations can re-anchor the belief after a disturbance.
This distinction is relevant to industrial sensing systems, where corrupted frames may occur intermittently due to occlusion, communication faults, sensor interference, or malicious manipulation.
A controller that recovers quickly from sparse corruption presents a different safety profile from one whose latent state remains contaminated after the disturbance disappears.

Third, input-level defenses should be treated as partial mitigations rather than certification mechanisms.
JPEG compression and median filtering recover part of the attack-induced loss in some tasks, but adaptive attacks reduce this apparent benefit and task-level results vary substantially.
This suggests that safety evaluation should include adaptive-defense testing and task-specific analysis instead of relying on average recovery alone.
In practice, ARB4WM can be used as a pre-deployment screening tool: candidate world-model controllers can be compared under the same attack objectives, exposure protocols, and recovery metrics before more expensive testing in high-fidelity digital twins or physical equipment.

\section{Conclusion}

This paper presents ARB4WM, a safety-oriented robustness testing framework for world-model agents in industrial continuous control.
We evaluate four Dreamer-family agents across MetaWorld and DMC under five white-box test objectives, two gradient-based perturbation optimizers, and multiple temporal exposure protocols.
The experimental results show that small visual perturbations can substantially degrade the performance of world-model agents, especially under multi-step iterative attacks.
They also show that failures targeting value estimation, latent representations, and RSSM dynamics can be as important as policy-space failures, indicating that safety assessment for world models should consider the full latent decision-making pipeline.

Our results further suggest that architecture and representation design affect adversarial robustness differently across benchmarks and control regimes.
R2-Dreamer \cite{morihira2026r2} achieves the strongest average robustness in the main MetaWorld and DMC evaluations, while Dreamer-Pro \cite{deng2022dreamerpro} shows weaker robustness despite its nominal policy optimization improvements.
These findings highlight the need for robustness-aware world-model design and provide a testing foundation for future work on secure visual model-based control in engineering systems.

\clearpage
\bibliographystyle{unsrtnat}
\bibliography{references}

@article{morihira2026r2,
  title={R2-Dreamer: Redundancy-reduced world models without decoders or augmentation},
  author={Morihira, Naoki and Nahar, Amal and Bharadwaj, Kartik and Kato, Yasuhiro and Hayashi, Akinobu and Harada, Tatsuya},
  journal={arXiv preprint arXiv:2603.18202},
  year={2026}
}

@article{guo2026wmattack,
  title={WMAttack: Automated Attack Search for Adversarial Evaluation of World-Model Agents},
  author={Guo, Zhixiang and Liang, Siyuan and Fu, Shi and Guo, Cheng and Balogh, Andr{\'a}s and Jelasity, M{\'a}rk and Tao, Dacheng},
  journal={arXiv preprint arXiv:2605.23220},
  year={2026}
}

@article{zhang2026hallucination,
  title={Adversarial Attacks Against World Models: Hallucination-Driven Policy Failure},
  author={Zhang, Junjian and Tan, Hao and Li, Ruonan and Li, Aiping and Gu, Zhaoquan},
  journal={Applied Sciences},
  volume={16},
  number={11},
  pages={5484},
  year={2026},
  doi={10.3390/app16115484}
}

@article{oord2018representation,
  title={Representation learning with contrastive predictive coding},
  author={Oord, Aaron van den and Li, Yazhe and Vinyals, Oriol},
  journal={arXiv preprint arXiv:1807.03748},
  year={2018}
}

@inproceedings{deng2022dreamerpro,
  title={Dreamerpro: Reconstruction-free model-based reinforcement learning with prototypical representations},
  author={Deng, Fei and Jang, Ingook and Ahn, Sungjin},
  booktitle={International conference on machine learning},
  pages={4956--4975},
  year={2022},
  organization={PMLR}
}

@article{hafner2025mastering,
  title={Mastering diverse control tasks through world models},
  author={Hafner, Danijar and Pasukonis, Jurgis and Ba, Jimmy and Lillicrap, Timothy},
  journal={Nature},
  pages={1--7},
  year={2025},
  publisher={Nature Publishing Group UK London}
}

@inproceedings{andrychowicz2021matters,
  title={What matters for on-policy deep actor-critic methods? a large-scale study},
  author={Andrychowicz, Marcin and Raichuk, Anton and Sta{\'n}czyk, Piotr and Orsini, Manu and Girgin, Sertan and Marinier, Rapha{\"e}l and Hussenot, Leonard and Geist, Matthieu and Pietquin, Olivier and Michalski, Marcin and others},
  booktitle={International conference on learning representations},
  year={2021}
}

@article{jones2025adversarial,
  title={Adversarial attacks on robotic vision language action models},
  author={Jones, Eliot Krzysztof and Robey, Alexander and Zou, Andy and Ravichandran, Zachary and Pappas, George J and Hassani, Hamed and Fredrikson, Matt and Kolter, J Zico},
  journal={arXiv preprint arXiv:2506.03350},
  year={2025}
}

@article{ha2018world,
  title={World models},
  author={Ha, David and Schmidhuber, J{\"u}rgen},
  journal={arXiv preprint arXiv:1803.10122},
  year={2018}
}

@article{lecun2022path,
  title={A path towards autonomous machine intelligence version 0.9. 2, 2022-06-27},
  author={LeCun, Yann},
  journal={Open Review},
  volume={62},
  number={1},
  pages={1--62},
  year={2022}
}

@article{hafner2020mastering,
  title={Mastering atari with discrete world models},
  author={Hafner, Danijar and Lillicrap, Timothy and Norouzi, Mohammad and Ba, Jimmy},
  journal={arXiv preprint arXiv:2010.02193},
  year={2020}
}

@inproceedings{hafner2019planet,
  title={Learning Latent Dynamics for Planning from Pixels},
  author={Hafner, Danijar and Lillicrap, Timothy and Fischer, Ian and Villegas, Ruben and Ha, David and Lee, Honglak and Davidson, James},
  booktitle={Proceedings of the 36th International Conference on Machine Learning},
  pages={2555--2565},
  year={2019},
  volume={97},
  series={Proceedings of Machine Learning Research},
  publisher={PMLR}
}

@inproceedings{hafner2020dreamer,
  title={Dream to Control: Learning Behaviors by Latent Imagination},
  author={Hafner, Danijar and Lillicrap, Timothy and Ba, Jimmy and Norouzi, Mohammad},
  booktitle={International Conference on Learning Representations},
  year={2020}
}

@article{tassa2018dmcontrol,
  title={DeepMind Control Suite},
  author={Tassa, Yuval and Doron, Yotam and Muldal, Alistair and Erez, Tom and Li, Yazhe and de Las Casas, Diego and Budden, David and Abdolmaleki, Abbas and Merel, Josh and Lefrancq, Andrew and Lillicrap, Timothy and Riedmiller, Martin},
  journal={arXiv preprint arXiv:1801.00690},
  year={2018}
}

@inproceedings{yu2020metaworld,
  title={Meta-World: A Benchmark and Evaluation for Multi-Task and Meta Reinforcement Learning},
  author={Yu, Tianhe and Quillen, Deirdre and He, Zhanpeng and Julian, Ryan and Hausman, Karol and Finn, Chelsea and Levine, Sergey},
  booktitle={Proceedings of the Conference on Robot Learning},
  pages={1094--1100},
  year={2020},
  volume={100},
  series={Proceedings of Machine Learning Research},
  publisher={PMLR}
}

@article{gu2024advancing,
  title={Advancing humanoid locomotion: Mastering challenging terrains with denoising world model learning},
  author={Gu, Xinyang and Wang, Yen-Jen and Zhu, Xiang and Shi, Chengming and Guo, Yanjiang and Liu, Yichen and Chen, Jianyu},
  journal={arXiv preprint arXiv:2408.14472},
  year={2024}
}

@inproceedings{wang2024drivedreamer,
  title={DriveDreamer: Towards Real-World-Drive World Models for Autonomous Driving},
  author={Wang, Xiaofeng and Zhu, Zheng and Huang, Guan and Chen, Xinze and Zhu, Jiagang and Lu, Jiwen},
  booktitle={European Conference on Computer Vision},
  pages={55--72},
  year={2024},
  organization={Springer}
}

@inproceedings{zheng2024occworld,
  title={Occworld: Learning a 3d occupancy world model for autonomous driving},
  author={Zheng, Wenzhao and Chen, Weiliang and Huang, Yuanhui and Zhang, Borui and Duan, Yueqi and Lu, Jiwen},
  booktitle={European conference on computer vision},
  pages={55--72},
  year={2024},
  organization={Springer}
}

@article{szegedy2013intriguing,
  title={Intriguing properties of neural networks},
  author={Szegedy, Christian and Zaremba, Wojciech and Sutskever, Ilya and Bruna, Joan and Erhan, Dumitru and Goodfellow, Ian and Fergus, Rob},
  journal={arXiv preprint arXiv:1312.6199},
  year={2013}
}

@article{FGSM,
  title={Explaining and harnessing adversarial examples},
  author={Goodfellow, Ian J and Shlens, Jonathon and Szegedy, Christian},
  journal={arXiv preprint arXiv:1412.6572},
  year={2014}
}

@article{PGD,
  title={Towards deep learning models resistant to adversarial attacks},
  author={Madry, Aleksander and Makelov, Aleksandar and Schmidt, Ludwig and Tsipras, Dimitris and Vladu, Adrian},
  journal={arXiv preprint arXiv:1706.06083},
  year={2017}
}

@inproceedings{dong2018boosting,
  title={Boosting adversarial attacks with momentum},
  author={Dong, Yinpeng and Liao, Fangzhou and Pang, Tianyu and Su, Hang and Zhu, Jun and Hu, Xiaolin and Li, Jianguo},
  booktitle={Proceedings of the IEEE Conference on Computer Vision and Pattern Recognition},
  pages={9185--9193},
  year={2018}
}

@article{yuan2019adversarial,
  title={Adversarial examples: Attacks and defenses for deep learning},
  author={Yuan, Xiaoyong and He, Pan and Zhu, Qile and Li, Xiaolin},
  journal={IEEE transactions on neural networks and learning systems},
  volume={30},
  number={9},
  pages={2805--2824},
  year={2019},
  publisher={IEEE}
}

@inproceedings{cohen2019certified,
  title={Certified adversarial robustness via randomized smoothing},
  author={Cohen, Jeremy and Rosenfeld, Elan and Kolter, Zico},
  booktitle={international conference on machine learning},
  pages={1310--1320},
  year={2019},
  organization={PMLR}
}

@article{brown2017adversarial,
  title={Adversarial patch},
  author={Brown, Tom B and Man{\'e}, Dandelion and Roy, Aurko and Abadi, Mart{\'\i}n and Gilmer, Justin},
  journal={arXiv preprint arXiv:1712.09665},
  year={2017}
}

@inproceedings{eykholt2018robust,
  title={Robust physical-world attacks on deep learning visual classification},
  author={Eykholt, Kevin and Evtimov, Ivan and Fernandes, Earlence and Li, Bo and Rahmati, Amir and Xiao, Chaowei and Prakash, Atul and Kohno, Tadayoshi and Song, Dawn},
  booktitle={Proceedings of the IEEE conference on computer vision and pattern recognition},
  pages={1625--1634},
  year={2018}
}

@inproceedings{papernot2017practical,
  title={Practical black-box attacks against machine learning},
  author={Papernot, Nicolas and McDaniel, Patrick and Goodfellow, Ian and Jha, Somesh and Celik, Z Berkay and Swami, Ananthram},
  booktitle={Proceedings of the 2017 ACM on Asia conference on computer and communications security},
  pages={506--519},
  year={2017}
}

@inproceedings{chen2020hopskipjumpattack,
  title={Hopskipjumpattack: A query-efficient decision-based attack},
  author={Chen, Jianbo and Jordan, Michael I and Wainwright, Martin J},
  booktitle={2020 ieee symposium on security and privacy (sp)},
  pages={1277--1294},
  year={2020},
  organization={IEEE}
}

@article{kos2017delving,
  title={Delving into adversarial attacks on deep policies},
  author={Kos, Jernej and Song, Dawn},
  journal={arXiv preprint arXiv:1705.06452},
  year={2017}
}

@article{huang2017adversarial,
  title={Adversarial attacks on neural network policies},
  author={Huang, Sandy and Papernot, Nicolas and Goodfellow, Ian and Duan, Yan and Abbeel, Pieter},
  journal={arXiv preprint arXiv:1702.02284},
  year={2017}
}

@inproceedings{lin2017tactics,
  title={Tactics of Adversarial Attack on Deep Reinforcement Learning Agents},
  author={Lin, Yen-Chen and Hong, Zhang-Wei and Liao, Yuan-Hong and Shih, Meng-Li and Liu, Ming-Yu and Sun, Min},
  booktitle={Proceedings of the Twenty-Sixth International Joint Conference on Artificial Intelligence},
  pages={3756--3762},
  year={2017},
  doi={10.24963/ijcai.2017/525}
}

@inproceedings{pinto2017robust,
  title={Robust Adversarial Reinforcement Learning},
  author={Pinto, Lerrel and Davidson, James and Sukthankar, Rahul and Gupta, Abhinav},
  booktitle={Proceedings of the 34th International Conference on Machine Learning},
  pages={2817--2826},
  year={2017},
  volume={70},
  series={Proceedings of Machine Learning Research},
  publisher={PMLR}
}

@inproceedings{zhang2020robust,
  title={Robust Deep Reinforcement Learning against Adversarial Perturbations on State Observations},
  author={Zhang, Huan and Chen, Hongge and Xiao, Chaowei and Li, Bo and Liu, Mingyan and Boning, Duane and Hsieh, Cho-Jui},
  booktitle={Advances in Neural Information Processing Systems},
  volume={33},
  pages={21024--21037},
  year={2020}
}

@inproceedings{cao2019adversarial,
  title={Adversarial sensor attack on lidar-based perception in autonomous driving},
  author={Cao, Yulong and Xiao, Chaowei and Cyr, Benjamin and Zhou, Yimeng and Park, Won and Rampazzi, Sara and Chen, Qi Alfred and Fu, Kevin and Mao, Z Morley},
  booktitle={Proceedings of the 2019 ACM SIGSAC conference on computer and communications security},
  pages={2267--2281},
  year={2019}
}

@inproceedings{zeng2024world,
  title={World Models: The Safety Perspective},
  author={Zeng, Zifan and Zhang, Chongzhe and Liu, Feng and Sifakis, Joseph and Zhang, Qunli and Liu, Shiming and Wang, Peng},
  booktitle={2024 IEEE 35th International Symposium on Software Reliability Engineering Workshops (ISSREW)},
  pages={369--376},
  year={2024},
  organization={IEEE}
}

@inproceedings{stutz2019disentangling,
  title={Disentangling adversarial robustness and generalization},
  author={Stutz, David and Hein, Matthias and Schiele, Bernt},
  booktitle={Proceedings of the IEEE/CVF conference on computer vision and pattern recognition},
  pages={6976--6987},
  year={2019}
}

@inproceedings{andriushchenko2020square,
  title={Square Attack: A Query-Efficient Black-Box Adversarial Attack via Random Search},
  author={Andriushchenko, Maksym and Croce, Francesco and Flammarion, Nicolas and Hein, Matthias},
  booktitle={European Conference on Computer Vision},
  pages={484--501},
  year={2020},
  organization={Springer}
}

@inproceedings{croce2020autoattack,
  title={Reliable Evaluation of Adversarial Robustness with an Ensemble of Diverse Parameter-Free Attacks},
  author={Croce, Francesco and Hein, Matthias},
  booktitle={International Conference on Machine Learning},
  pages={2206--2216},
  year={2020},
  organization={PMLR}
}

@inproceedings{croce2021robustbench,
  title={RobustBench: A Standardized Adversarial Robustness Benchmark},
  author={Croce, Francesco and Andriushchenko, Maksym and Sehwag, Vikash and Debenedetti, Edoardo and Flammarion, Nicolas and Chiang, Mung and Mittal, Prateek and Hein, Matthias},
  booktitle={Advances in Neural Information Processing Systems Datasets and Benchmarks Track},
  year={2021}
}

@inproceedings{behzadan2017robustness,
  title={On the Robustness of Deep Reinforcement Learning to Adversarial Attacks},
  author={Behzadan, Vahid and Munir, Arslan},
  booktitle={Machine Learning and Data Mining in Pattern Recognition},
  pages={262--274},
  year={2017},
  organization={Springer}
}

@inproceedings{gleave2020adversarial,
  title={Adversarial Policies: Attacking Deep Reinforcement Learning},
  author={Gleave, Adam and Dennis, Michael and Wild, Cody and Kant, Neel and Levine, Sergey and Russell, Stuart},
  booktitle={International Conference on Learning Representations},
  year={2020}
}

@inproceedings{micheli2023iris,
  title={Transformers are Sample-Efficient World Models},
  author={Micheli, Vincent and Alonso, Eloi and Fleuret, Francois},
  booktitle={International Conference on Learning Representations},
  year={2023}
}

@inproceedings{hansen2024tdmpc2,
  title={TD-MPC2: Scalable, Robust World Models for Continuous Control},
  author={Hansen, Nicklas and Su, Hao and Wang, Xiaolong},
  booktitle={International Conference on Learning Representations},
  year={2024}
}

@inproceedings{yang2024simulators,
  title={Learning Interactive Real-World Simulators},
  author={Yang, Sherry and Du, Yilun and Ghasemipour, Kamyar and Tompson, Jonathan and Kaelbling, Leslie and Schuurmans, Dale and Abbeel, Pieter},
  booktitle={International Conference on Learning Representations},
  year={2024}
}

@inproceedings{huang2024safedreamer,
  title={SafeDreamer: Safe Reinforcement Learning with World Models},
  author={Huang, Weidong and Ji, Jiaming and Xia, Chunhe and Zhang, Borong and Yang, Yaodong},
  booktitle={International Conference on Learning Representations},
  year={2024}
}

@article{ye2024robustmbrl,
  title={Towards Robust Model-Based Reinforcement Learning against Adversarial Corruption},
  author={Ye, Chenlu and He, Jiafan and Gu, Quanquan and Zhang, Tong},
  journal={Proceedings of the 41st International Conference on Machine Learning},
  volume={235},
  pages={56982--57017},
  year={2024}
}

@article{sun2024latentdynamic,
  title={Learning Latent Dynamic Robust Representations for World Models},
  author={Sun, Ruixiang and Zang, Hongyu and Li, Xin and Islam, Riashat},
  journal={arXiv preprint arXiv:2405.06263},
  year={2024}
}

@article{zollicoffer2025surprise,
  title={World Model Robustness via Surprise Recognition},
  author={Zollicoffer, Geigh and Chopra, Tanush and Yan, Mingkuan and Ma, Xiaoxu and Eaton, Kenneth and Riedl, Mark},
  journal={arXiv preprint arXiv:2512.01119},
  year={2025}
}

@article{as2024safe,
  title={Safe Exploration Using Bayesian World Models and Log-Barrier Optimization},
  author={As, Yarden and Sukhija, Bhavya and Krause, Andreas},
  journal={arXiv preprint arXiv:2405.05890},
  year={2024}
}

@article{upadhyay2026worldbench,
  title={WorldBench: Disambiguating Physics for Diagnostic Evaluation of World Models},
  author={Upadhyay, Rishi and Zhang, Howard and Solomon, Jim and Agrawal, Ayush and Boreddy, Pranay and Satya Narayana, Shruti and Ba, Yunhao and Wong, Alex and de Melo, Celso M. and Kadambi, Achuta},
  journal={arXiv preprint arXiv:2601.21282},
  year={2026}
}

\clearpage
\appendix
\section{Task-Level Robustness Results}
\label{app:task_level_robustness}

This appendix reports task-level robustness curves for all 20 benchmark tasks and four Dreamer-family agents.
For each task, the figure shows the performance of each model under Random, PDA, PEA, DCA, VSA, and LDA perturbations.
Solid lines denote single-step attacks, while dashed lines denote multi-step iterative attacks.

\subsection{Per-Task Robustness Curves}

The following figures provide the task-level results behind the aggregate nAUC values reported in the main paper.
Each figure corresponds to one task and contains four panels, one for each victim model.
Within a panel, colors denote attack objectives and line style denotes the optimizer.
The Random curve provides a visual-noise baseline, while the five white-box objectives show how different components of the world-model pipeline fail as the perturbation budget increases.
These figures reveal several effects that are averaged out in the main tables: some tasks exhibit abrupt performance collapse after a small perturbation threshold, some models retain high performance under random noise but fail quickly under optimized attacks, and the relative strength of PDA, PEA, DCA, VSA, and LDA varies across tasks and agents.
They are included to make the benchmark results auditable at the level of individual environments rather than only through cross-task means.

% Auto-generated by util/plot_appendix_task_robustness.py
% Paste into the appendix as needed.

\begin{figure*}[t]
\centering
\includegraphics[width=\textwidth]{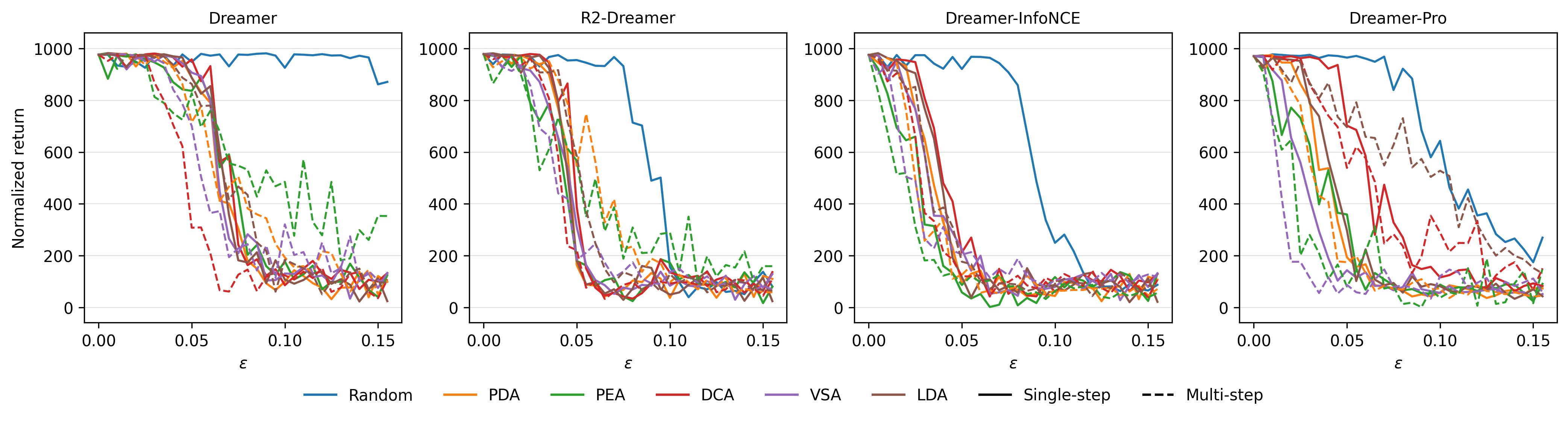}
\caption{Task-level robustness curves on DMC Reacher Easy. Each column shows one victim model. Solid lines denote single-step attacks, and dashed lines denote multi-step attacks.}
\label{fig:appendix_dmc-reacher-easy}
\end{figure*}

\begin{figure*}[t]
\centering
\includegraphics[width=\textwidth]{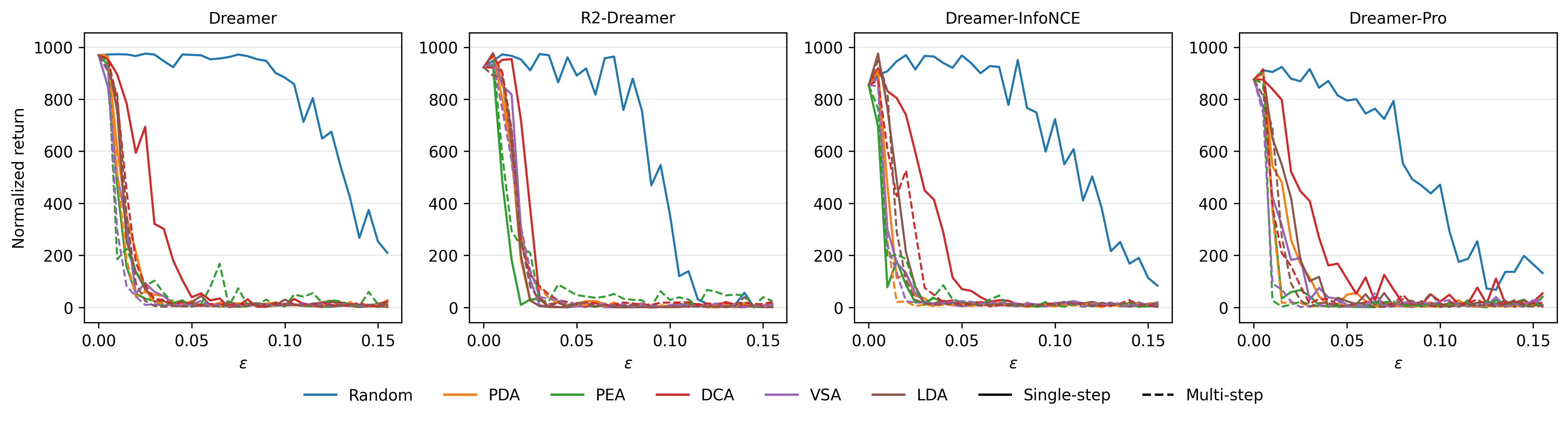}
\caption{Task-level robustness curves on DMC Reacher Hard. Each column shows one victim model. Solid lines denote single-step attacks, and dashed lines denote multi-step attacks.}
\label{fig:appendix_dmc-reacher-hard}
\end{figure*}

\begin{figure*}[t]
\centering
\includegraphics[width=\textwidth]{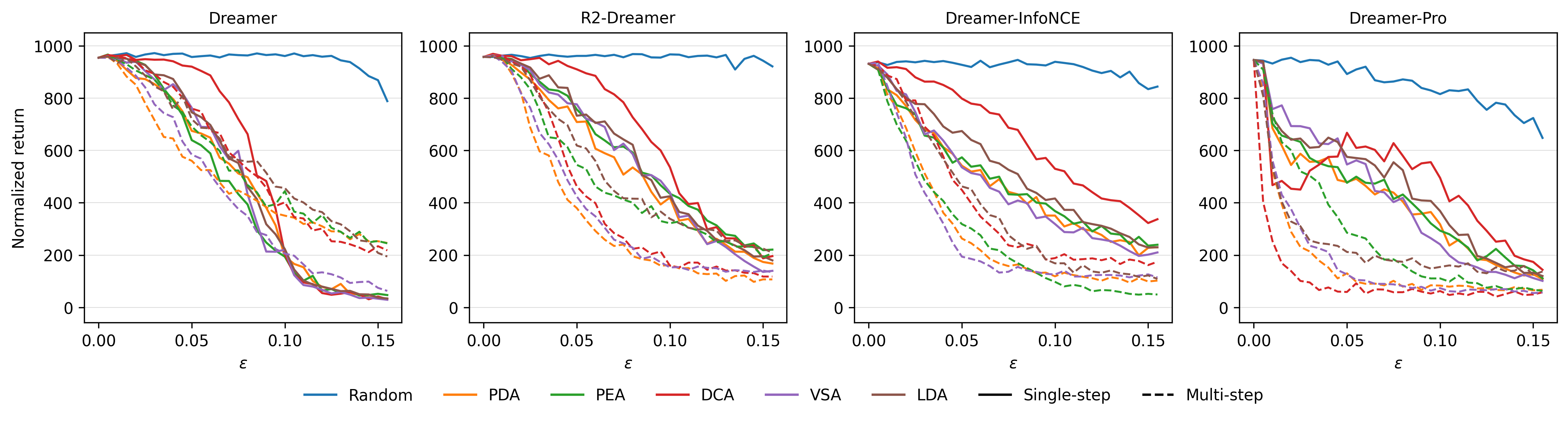}
\caption{Task-level robustness curves on DMC Walker Walk. Each column shows one victim model. Solid lines denote single-step attacks, and dashed lines denote multi-step attacks.}
\label{fig:appendix_dmc-walker-walk}
\end{figure*}

\begin{figure*}[t]
\centering
\includegraphics[width=\textwidth]{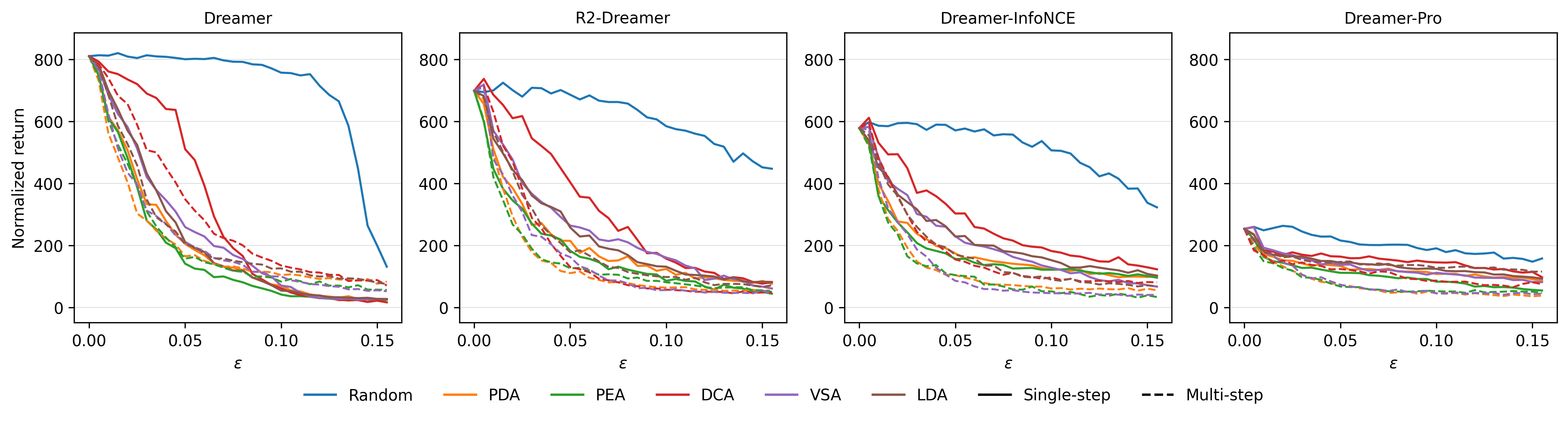}
\caption{Task-level robustness curves on DMC Walker Run. Each column shows one victim model. Solid lines denote single-step attacks, and dashed lines denote multi-step attacks.}
\label{fig:appendix_dmc-walker-run}
\end{figure*}

\begin{figure*}[t]
\centering
\includegraphics[width=\textwidth]{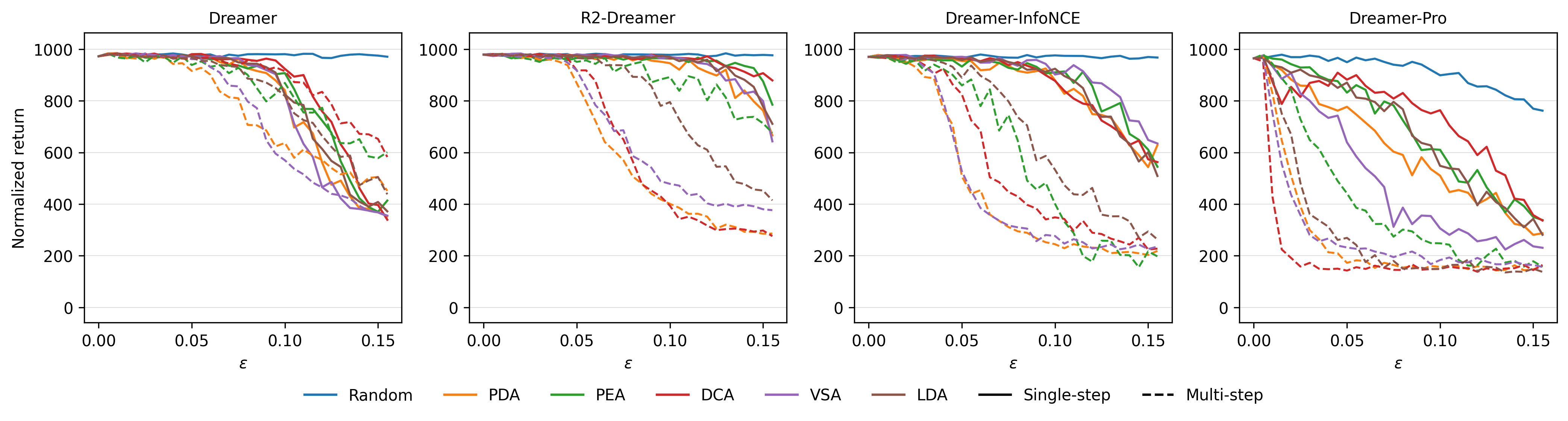}
\caption{Task-level robustness curves on DMC Walker Stand. Each column shows one victim model. Solid lines denote single-step attacks, and dashed lines denote multi-step attacks.}
\label{fig:appendix_dmc-walker-stand}
\end{figure*}

\begin{figure*}[t]
\centering
\includegraphics[width=\textwidth]{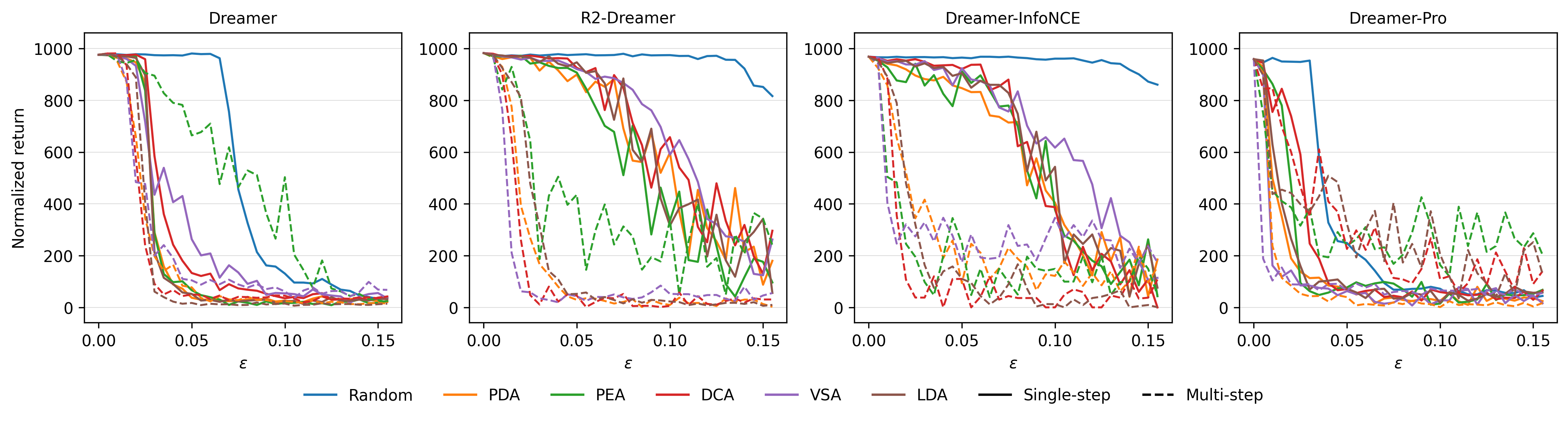}
\caption{Task-level robustness curves on DMC Ball In Cup Catch. Each column shows one victim model. Solid lines denote single-step attacks, and dashed lines denote multi-step attacks.}
\label{fig:appendix_dmc-ball-in-cup-catch}
\end{figure*}

\begin{figure*}[t]
\centering
\includegraphics[width=\textwidth]{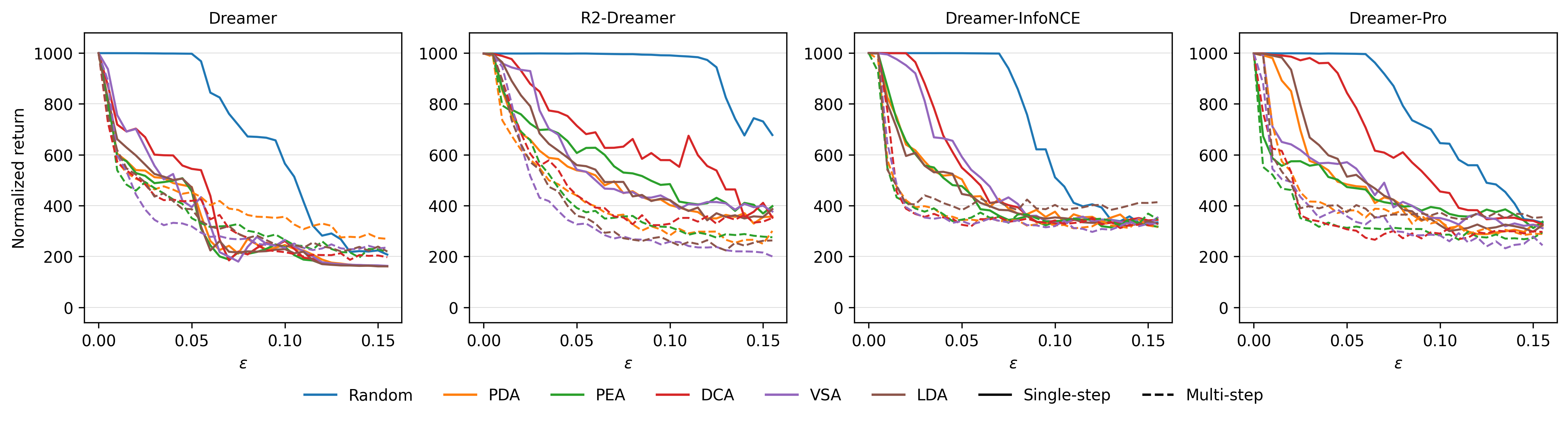}
\caption{Task-level robustness curves on DMC Cartpole Balance. Each column shows one victim model. Solid lines denote single-step attacks, and dashed lines denote multi-step attacks.}
\label{fig:appendix_dmc-cartpole-balance}
\end{figure*}

\begin{figure*}[t]
\centering
\includegraphics[width=\textwidth]{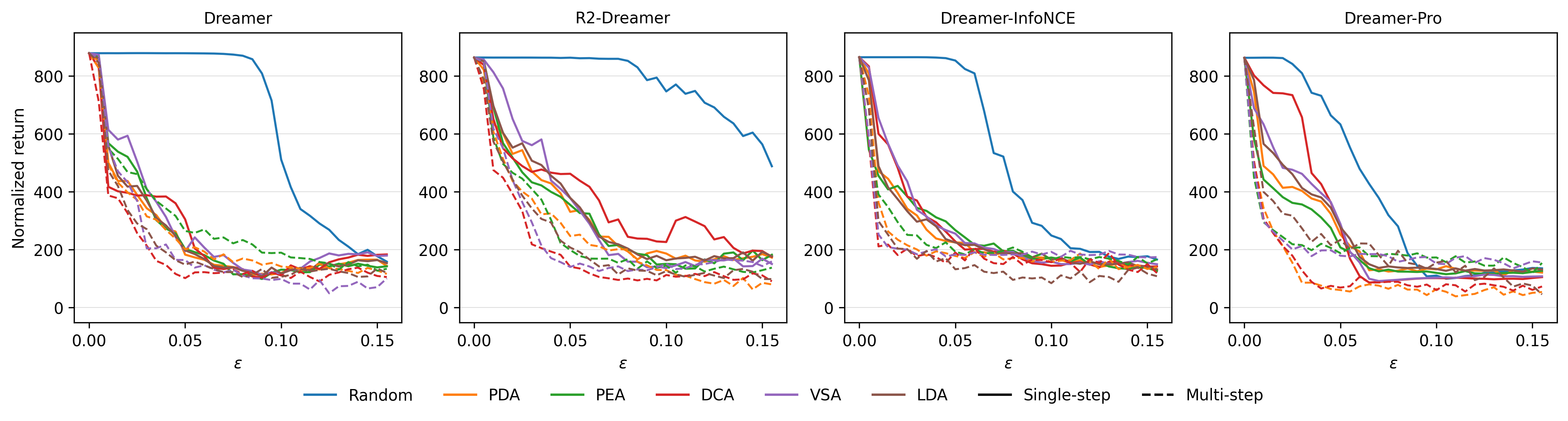}
\caption{Task-level robustness curves on DMC Cartpole Swingup. Each column shows one victim model. Solid lines denote single-step attacks, and dashed lines denote multi-step attacks.}
\label{fig:appendix_dmc-cartpole-swingup}
\end{figure*}

\begin{figure*}[t]
\centering
\includegraphics[width=\textwidth]{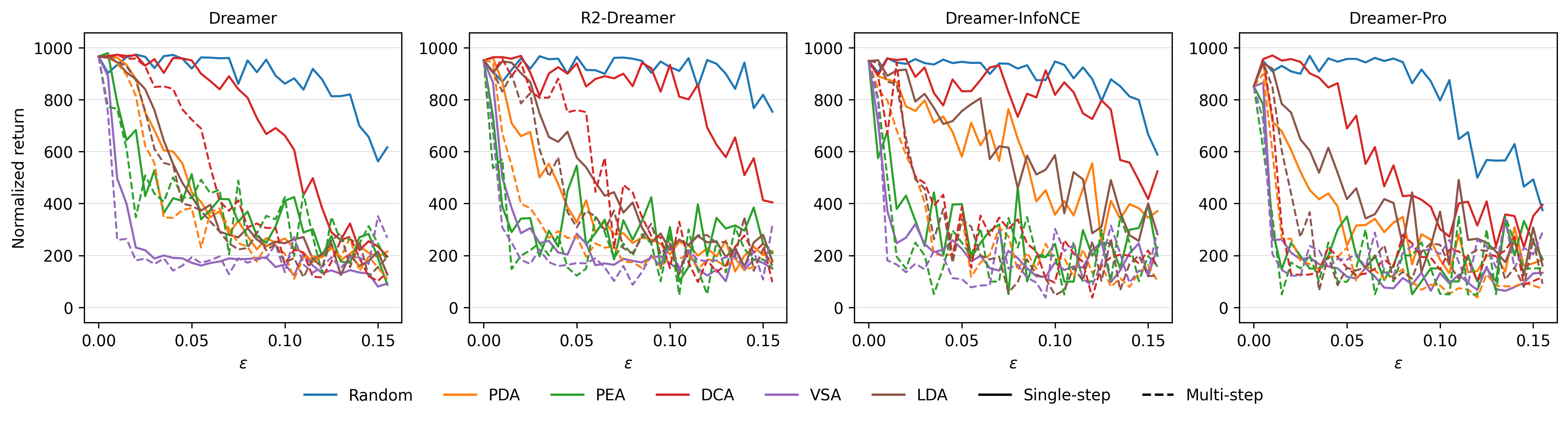}
\caption{Task-level robustness curves on DMC Finger Turn Easy. Each column shows one victim model. Solid lines denote single-step attacks, and dashed lines denote multi-step attacks.}
\label{fig:appendix_dmc-finger-turn-easy}
\end{figure*}

\begin{figure*}[t]
\centering
\includegraphics[width=\textwidth]{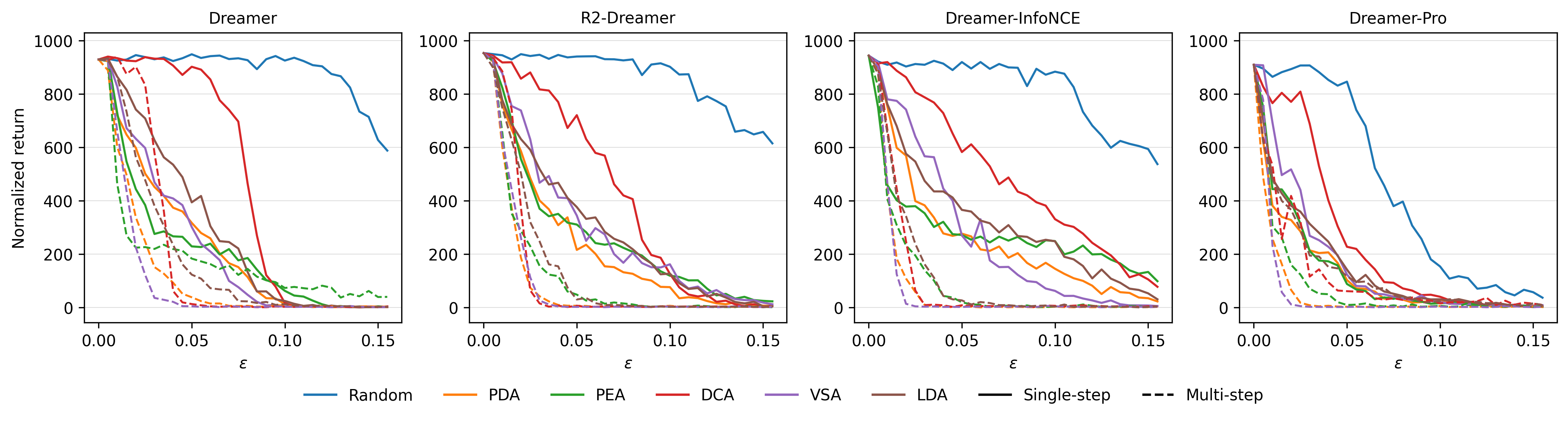}
\caption{Task-level robustness curves on DMC Hopper Stand. Each column shows one victim model. Solid lines denote single-step attacks, and dashed lines denote multi-step attacks.}
\label{fig:appendix_dmc-hopper-stand}
\end{figure*}

\begin{figure*}[t]
\centering
\includegraphics[width=\textwidth]{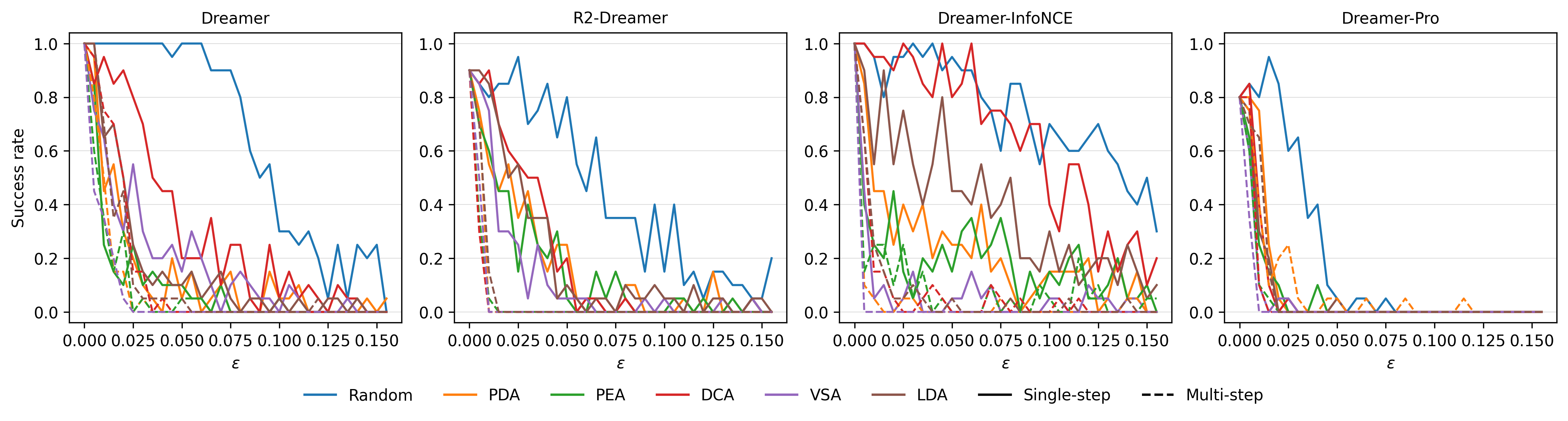}
\caption{Task-level robustness curves on MetaWorld Coffee Pull. Each column shows one victim model. Solid lines denote single-step attacks, and dashed lines denote multi-step attacks.}
\label{fig:appendix_metaworld-coffee-pull}
\end{figure*}

\begin{figure*}[t]
\centering
\includegraphics[width=\textwidth]{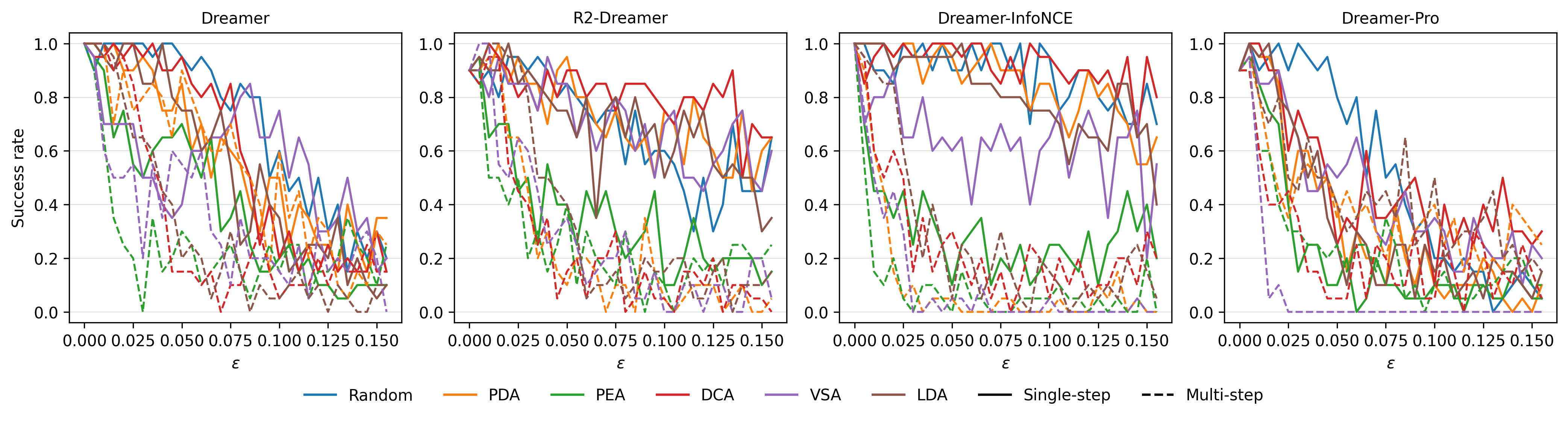}
\caption{Task-level robustness curves on MetaWorld Reach. Each column shows one victim model. Solid lines denote single-step attacks, and dashed lines denote multi-step attacks.}
\label{fig:appendix_metaworld-reach}
\end{figure*}

\begin{figure*}[t]
\centering
\includegraphics[width=\textwidth]{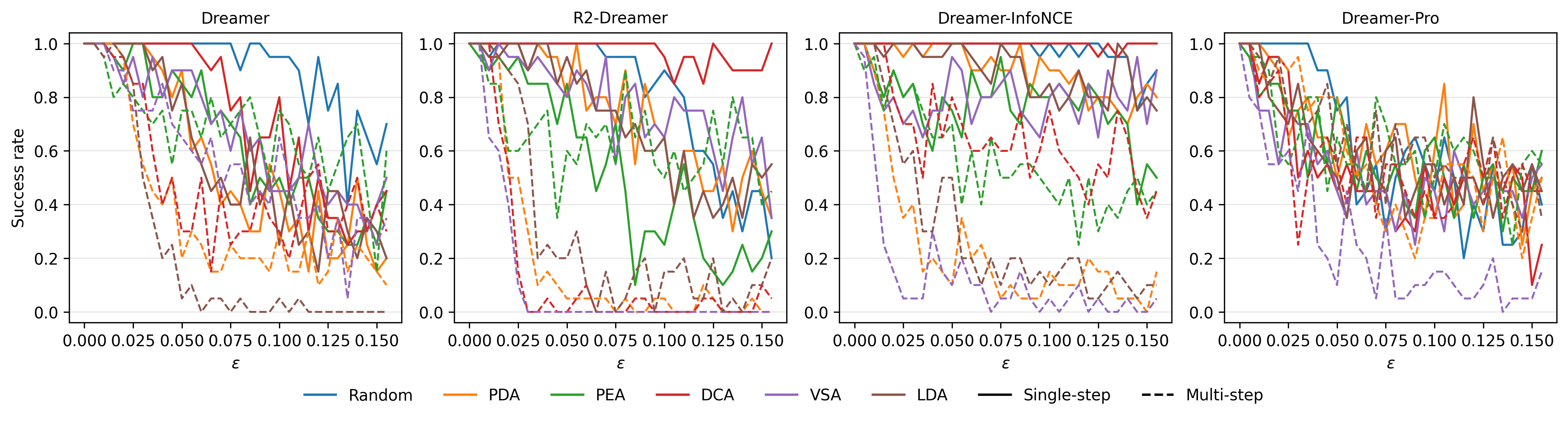}
\caption{Task-level robustness curves on MetaWorld Reach Wall. Each column shows one victim model. Solid lines denote single-step attacks, and dashed lines denote multi-step attacks.}
\label{fig:appendix_metaworld-reach-wall}
\end{figure*}

\begin{figure*}[t]
\centering
\includegraphics[width=\textwidth]{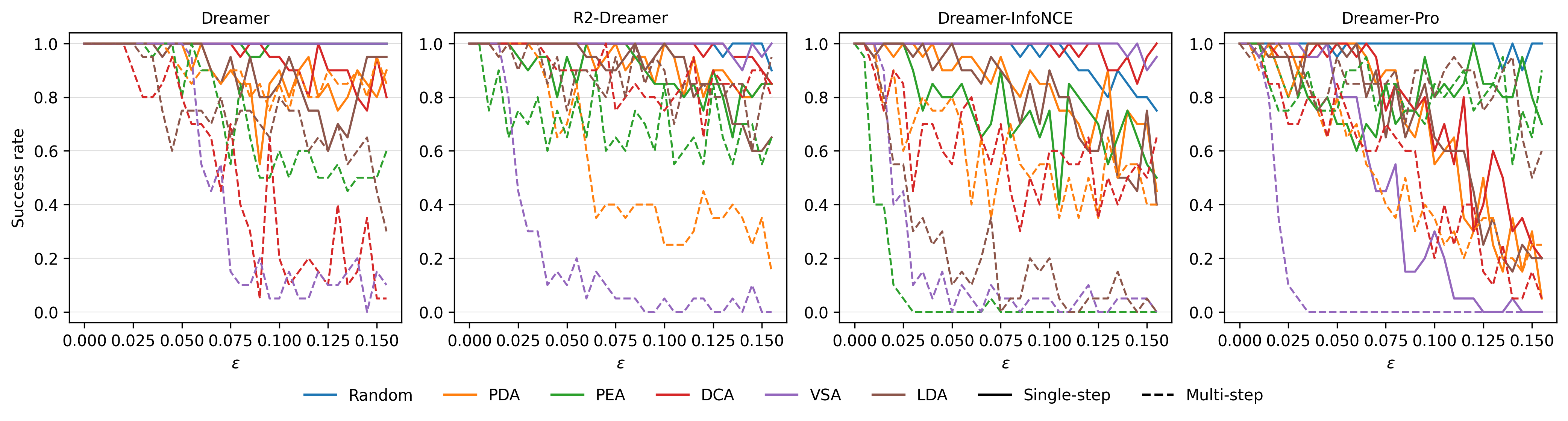}
\caption{Task-level robustness curves on MetaWorld Window Close. Each column shows one victim model. Solid lines denote single-step attacks, and dashed lines denote multi-step attacks.}
\label{fig:appendix_metaworld-window-close}
\end{figure*}

\begin{figure*}[t]
\centering
\includegraphics[width=\textwidth]{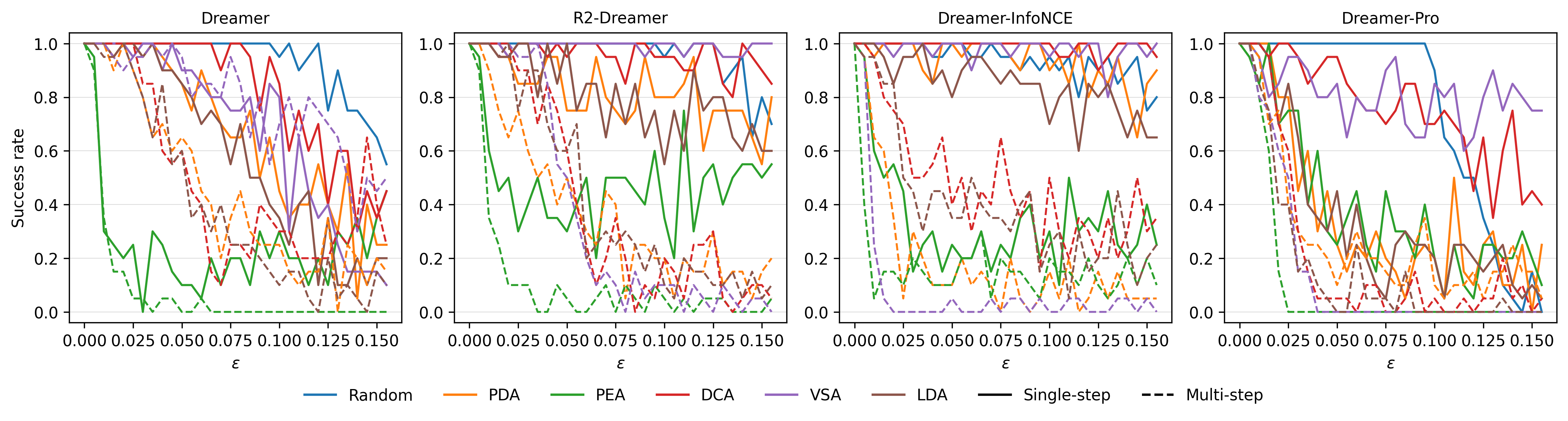}
\caption{Task-level robustness curves on MetaWorld Window Open. Each column shows one victim model. Solid lines denote single-step attacks, and dashed lines denote multi-step attacks.}
\label{fig:appendix_metaworld-window-open}
\end{figure*}

\begin{figure*}[t]
\centering
\includegraphics[width=\textwidth]{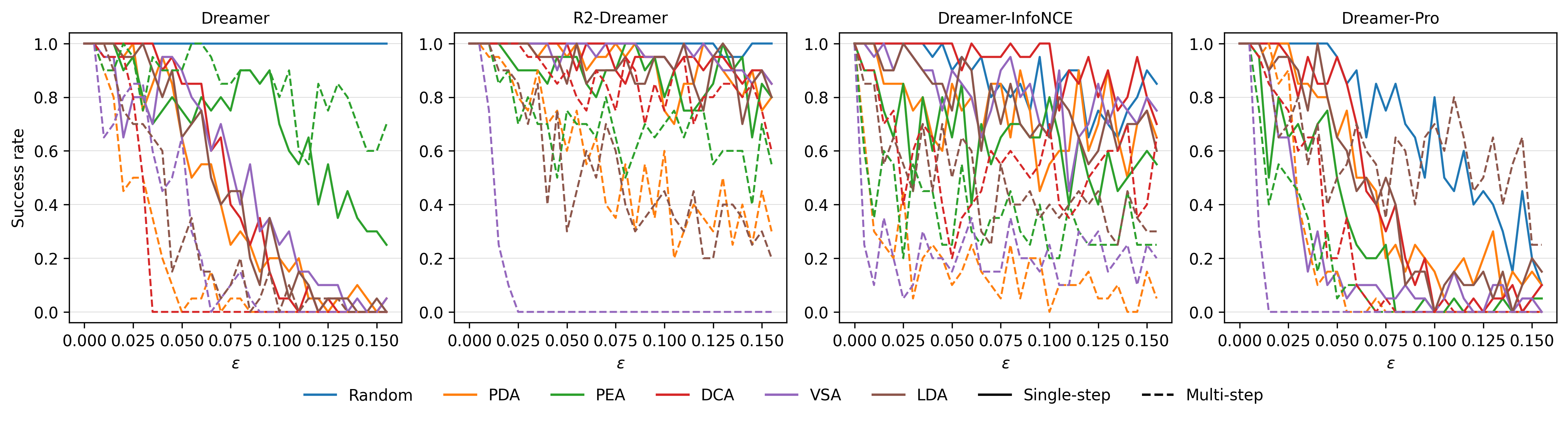}
\caption{Task-level robustness curves on MetaWorld Button Press. Each column shows one victim model. Solid lines denote single-step attacks, and dashed lines denote multi-step attacks.}
\label{fig:appendix_metaworld-button-press}
\end{figure*}

\begin{figure*}[t]
\centering
\includegraphics[width=\textwidth]{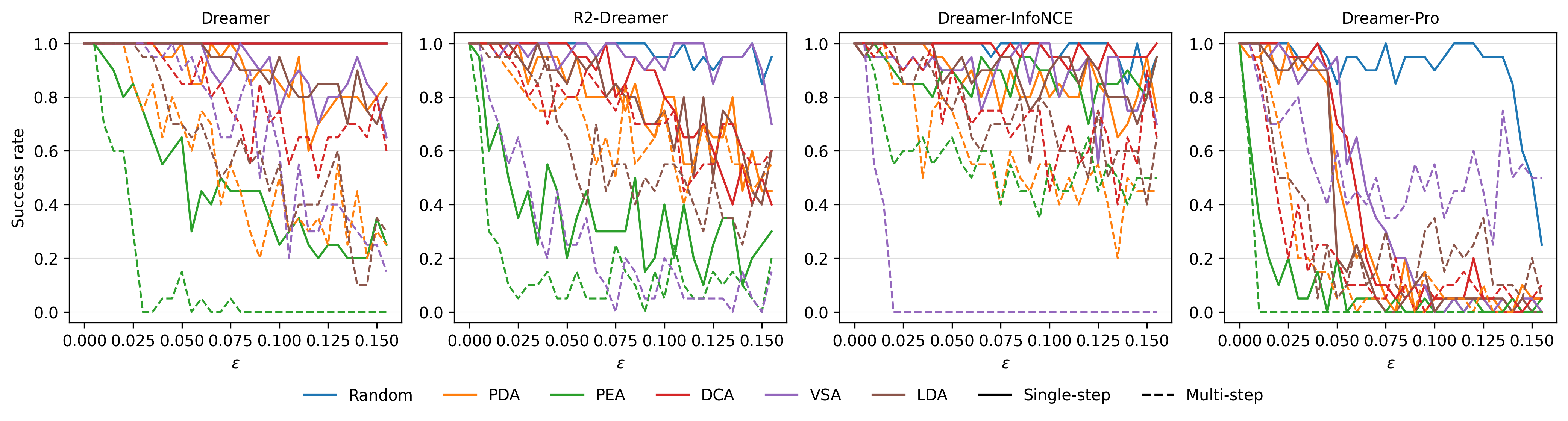}
\caption{Task-level robustness curves on MetaWorld Handle Press. Each column shows one victim model. Solid lines denote single-step attacks, and dashed lines denote multi-step attacks.}
\label{fig:appendix_metaworld-handle-press}
\end{figure*}

\begin{figure*}[t]
\centering
\includegraphics[width=\textwidth]{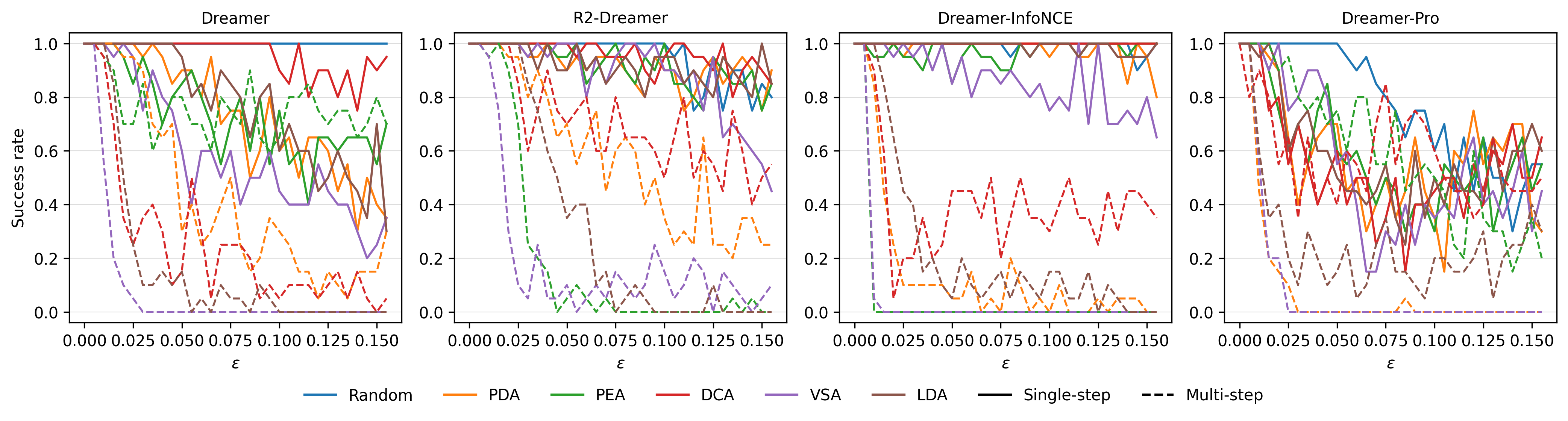}
\caption{Task-level robustness curves on MetaWorld Plate Slide Back. Each column shows one victim model. Solid lines denote single-step attacks, and dashed lines denote multi-step attacks.}
\label{fig:appendix_metaworld-plate-slide-back}
\end{figure*}

\begin{figure*}[t]
\centering
\includegraphics[width=\textwidth]{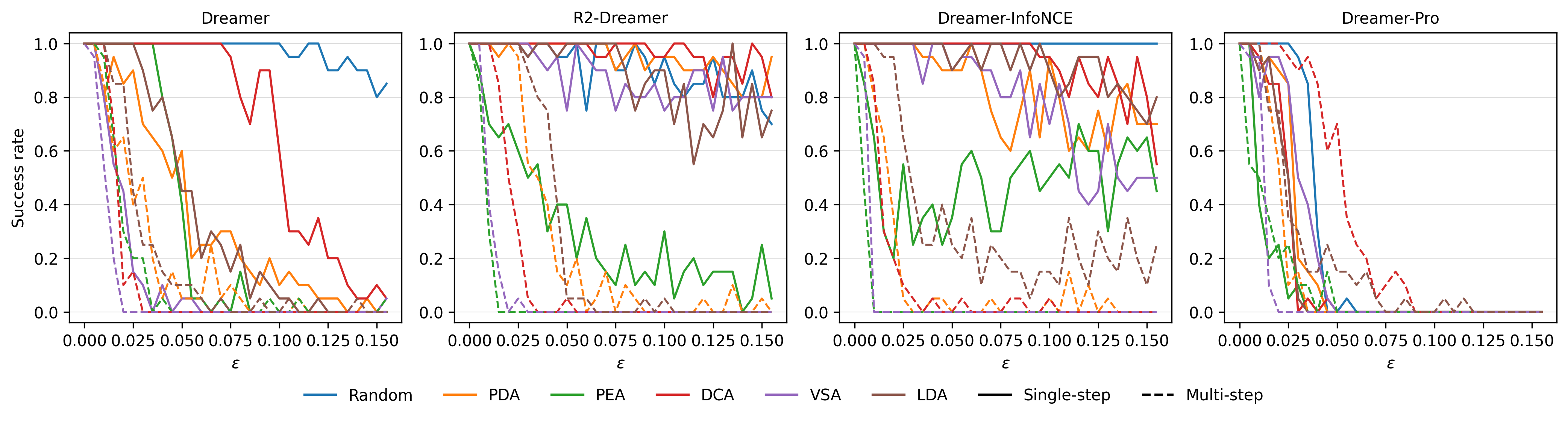}
\caption{Task-level robustness curves on MetaWorld Plate Slide Back Side. Each column shows one victim model. Solid lines denote single-step attacks, and dashed lines denote multi-step attacks.}
\label{fig:appendix_metaworld-plate-slide-back-side}
\end{figure*}

\begin{figure*}[t]
\centering
\includegraphics[width=\textwidth]{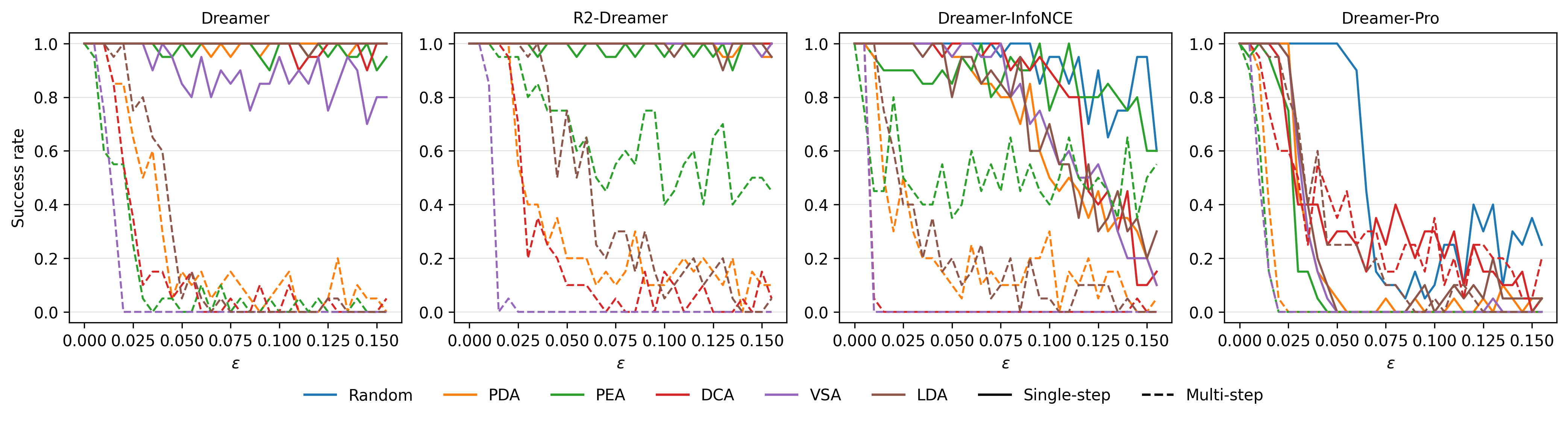}
\caption{Task-level robustness curves on MetaWorld Plate Slide Side. Each column shows one victim model. Solid lines denote single-step attacks, and dashed lines denote multi-step attacks.}
\label{fig:appendix_metaworld-plate-slide-side}
\end{figure*}

\end{document}